\documentclass[10pt,journal,compsoc]{IEEEtran}
\usepackage{times}
\usepackage{epsfig}
\usepackage{graphicx}
\usepackage{amsmath}
\usepackage{amssymb}
\usepackage{enumitem}
\usepackage{bm}
\usepackage{multirow}
\usepackage[pagebackref=true,breaklinks=true,letterpaper=true,colorlinks,bookmarks=false]{hyperref}
\newcommand{\eg}{\text{e.g.}}
\newcommand{\ie}{\text{i.e.}}

\ifCLASSOPTIONcompsoc
  \usepackage[nocompress]{cite}
\else
  \usepackage{cite}
\fi

\ifCLASSINFOpdf
\else
\fi

\newcommand{\re}[1]{{\color{black}{#1}}}

\begin{document}
\setlength{\abovedisplayskip}{2pt}
\setlength{\belowdisplayskip}{2pt}
\title{Deconfounded Image Captioning: A Causal Retrospect}

\author{Xu~Yang,%~\IEEEmembership{Member,~IEEE,}
        ~Hanwang~Zhang,%~\IEEEmembership{Fellow,~OSA,}
        ~and~Jianfei~Cai%~\IEEEmembership{Life~Fellow,~IEEE}% <-this % stops a space
\IEEEcompsocitemizethanks{\IEEEcompsocthanksitem Xu Yang is currently an Associate Professor at School of Computer Science and Engineering of Southeast University, China.\protect\\
E-mail: xuyangseu@ieee.org

\IEEEcompsocthanksitem Hanwang Zhang is currently an Assistant Professor at Nanyang Technological University, Singapore.\protect\\
E-mail: hanwangzhang@ntu.edu.sg

\IEEEcompsocthanksitem Jianfei Cai is currently a Professor at the Data Science \& AI Department at the Faculty of IT, Monash University, Australia.\protect\\ 
E-mail: Jianfei.Cai@monash.edu

}}% <-this % stops an unwanted space

\markboth{IEEE TRANSACTIONS ON PATTERN ANALYSIS AND MACHINE INTELLIGENCE}%
{Shell \MakeLowercase{\textit{et al.}}: Bare Demo of IEEEtran.cls for Computer Society Journals}

\IEEEtitleabstractindextext{%
\begin{abstract}
Dataset bias in vision-language tasks is becoming one of the main problems which hinders the progress of our community. Existing solutions lack a principled analysis about why modern image captioners easily collapse into dataset bias. In this paper, we present a novel perspective: Deconfounded Image Captioning (DIC), to find out the answer of this question, then retrospect modern neural image captioners, and finally propose a DIC framework: DICv1.0 to alleviate the negative effects brought by dataset bias. DIC is based on causal inference, whose two principles: the backdoor and front-door adjustments, help us review previous studies and design new effective models. In particular, we showcase that DICv1.0 can strengthen two prevailing captioning models and can achieve a single-model 131.1 CIDEr-D and 128.4 c40 CIDEr-D on Karpathy split and online split of the challenging MS COCO dataset, respectively. Interestingly, DICv1.0 is a natural derivation from our causal retrospect, which opens promising directions for image captioning.
\end{abstract}

% Note that keywords are not normally used for peerreview papers.
}

% make the title area
\maketitle

\IEEEdisplaynontitleabstractindextext

\IEEEpeerreviewmaketitle

%#################################Introduction#############################
\IEEEraisesectionheading{\section{Introduction}\label{sec:introduction}}
\IEEEPARstart{R}{ecently}, our vision-language community has paid more and more attention to dataset bias. On one hand, it shows that we have already achieved impressive or even super-human performances on many benchmark leader-boards. On the other hand, it also shows that we do not actually believe in these ``super-human'' systems due to the obvious dataset bias, \eg, image captioners are likely to imply gender discrimination~\cite{hendricks2018women} (\eg, generating ``a man'' instead of ``a woman'' even through the image actually shows ``a woman is snowboarding'') and hallucinate what they do not see~\cite{rohrbach2018object} (\eg, generating ``sitting on the bench'' when only seeing people talking on the phone); VQA agents usually answer ``Yes'' when being asked ``Are the apples red?'' even without a look at the image~\cite{goyal2017making}. 

The ideal solution %first attempt 
to eliminate dataset bias is to collect a perfect and totally balanced dataset. However, since humans are living in a world filled with various biases, \eg, long-tailed concept distributions~\cite{reed2001pareto,liu2019large}, reporting bias~\cite{gordon2013reporting,misra2016seeing}, and language bias~\cite{bolukbasi2016man,caliskan2017semantics}, we are easily trapped in the loop: ``making a dataset''--``it's biased''--``making a new one''~\footnote{by Alexei Efros@CVPR2019 Computer Vision After 5 Years workshop ( \url{https://futurecv.github.io/schedule.html})}. Promisingly, by observing that our biological vision-language system works well even though the around real word is biased, we are prompted to evolve from diagnosis to treatment, by changing from collecting new datasets~\cite{johnson2017clevr,goyal2017making,agrawal2018don,sharma2018conceptual} to designing unbiased models~\cite{hendricks2018women,cadene2019rubi} for confronting the bias. However, recent studies still lack a principled analysis about why a vision-language learning system easily collapses into dataset bias. In this paper, we exploit the concepts of \textbf{confounder} to analysis this problem, and propose a principled solution to alleviate the negative effects brought by dataset bias based on the two principles of causal inference: \textbf{the backdoor and front-door adjustments}~\cite{pearl2000causality,pearl2016causal,PearlMackenzie18}. In particular, we use Image Captioning (IC) as the case study because it has the longest history\footnote{Image captioning~\cite{yao2010i2t} has been proposed in 2010, while the other two important vision language tasks, Visual Question Answering and Visual Grounding, were proposed in 2015~\cite{antol2015vqa} and 2014~\cite{kazemzadeh2014referitgame}, respectively.} and the simplest cross-modal objective among all the vision-language tasks~\cite{mao2016generation,vinyals2015show,antol2015vqa,das2017visual,hudson2019gqa}. 

\begin{figure}[t]
\centering
\includegraphics[width=1\linewidth,trim = 5mm 5mm 5mm 5mm,clip]{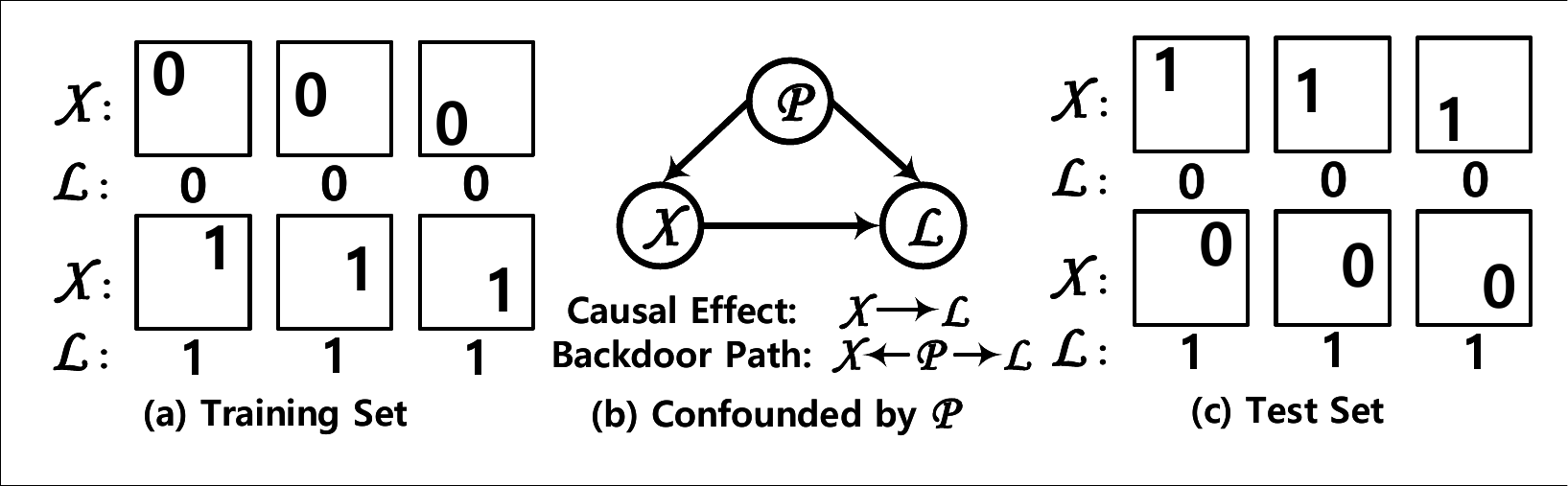}
   \caption{ The toy example about how the confounder disturbs the learning of the true causal effect, where $\mathcal{X}$, $\mathcal{L}$, and $\mathcal{P}$ denote image features, label, and position, respectively. (a) A training dataset constructed with a bias that ``0'' is mostly on the left and ``1'' is mostly on the right.
   (b) If we train a digit classifier by the training set in (a), it may not capture the true causal effect $\mathcal{X} \rightarrow \mathcal{L}$, while exploiting the ``position knowledge'' to make predictions (\eg, when recognizing pixels on the left, predicting the label as ``0'') through the backdoor path $\mathcal{X} \leftarrow \mathcal{P} \rightarrow \mathcal{L}$ induced by the confounder $\mathcal{P}$. (c) As a result, if we feed this classifier with the images where ``1'' appears on the left and ``0'' on the right, the incorrect predictions will be made.
   }
\label{fig:fig_digit}
% \vspace{-0.2in}
\end{figure}
\begin{figure*}[t]
\centering
\includegraphics[width=1\linewidth,trim = 5mm 5mm 5mm 5mm,clip]{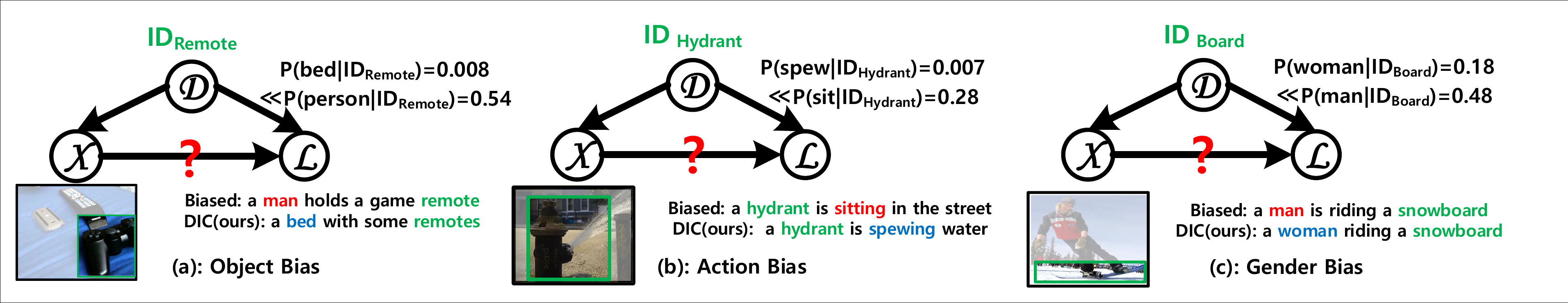}
   \caption{Three examples show how the confounder $\mathcal{D}$ induces the spurious correlation. The probability denotes the percentage of co-occurrence of two words in the training set, \eg, $P(\text{person}|ID_{\text{Remote}})$ means that ``person'' and ``remote'' contributes the 54\% occurrences of ``remote''. Here we use $ID$ to signify the appearing visual concept. The top caption is generated by Up-Down~\cite{anderson2018bottom} and the bottom one is generated by our UD-DICv1.0. Red and blue denote the wrong and right words, respectively, and green denotes the concept which may cause the wrong word.
   }
\label{fig:fig_demo}
% \vspace{-0.2in}
\end{figure*}
In the classic causal inference framework~\cite{pearl2000causality,pearl2016causal,PearlMackenzie18}, if a variable is a common cause of the other two variables, such variable is called the confounder. One toy example about confounder in machine learning is given in Fig.~\ref{fig:fig_digit}, where the training dataset is constructed with a bias, i.e., ``0'' mostly appears on the left and ``1'' mostly appears on the right. Such biased dataset entails two causal effects: the position $\mathcal{P}$ affects the image features $\mathcal{X}$ and the digit label $\mathcal{L}$. For example, the left part of an image is likely to contain pixels configuring ``0'' and if the left part of an image contains black pixels, the label is likely to be ``0''. In the causal inference, we can draw two causal links to describe these causal effects: $\mathcal{P} \rightarrow \mathcal{X}$ and $\mathcal{P} \rightarrow \mathcal{L}$, where the tail is the cause of the head. If we want to build a digit classifier to learn the true causal effect $\mathcal{X} \rightarrow \mathcal{L}$ while employing the biased dataset to train this classifier (Fig.~\ref{fig:fig_digit}(a)), this classifier may simply recognize which part of the image contains more black pixels, \ie, through $\mathcal{X} \leftarrow \mathcal{P}$\footnote{It is noteworthy that the link direction only indexes the causal direction and the information can also pass from the head to the tail, \eg, recognizing pixels from $\mathcal{X}$ to $\mathcal{P}$.}, and then use this position knowledge to classify the label, \ie, through $\mathcal{P} \rightarrow \mathcal{L}$. In this way, this classifier learns the \textbf{spurious correlation} between $\mathcal{X}$ and $\mathcal{L}$ through \textbf{the backdoor path} $\mathcal{X} \leftarrow \mathcal{P} \rightarrow \mathcal{L}$ induced by the confounder $\mathcal{P}$ as in Fig.~\ref{fig:fig_digit}(b). As a result, this classifier does not really exploit the digit shapes to predict the labels. During testing, if we provide some images where ``1'' appears on left or ``0'' appears on right as in Fig.~\ref{fig:fig_digit}(c), this classifier will make incorrect predictions. This is how the confounder disturbs the learning of the true causal effect.

Extending this toy example to the IC case, with a similar basic causal graph shown in Fig.~\ref{fig:fig_demo}, we denote $\mathcal{D}$ as a confounder due to dataset biases, $\mathcal{X}$ as image features to be learned and $\mathcal{L}$ as the resultant caption. In particular, we can treat the set of some frequently appearing concepts $\mathcal{D}$ in the training dataset as the confounder. On one hand, the more images of these concepts that a network is trained with, the more the learned representation $\mathcal{X}$ is affected. Thus, we have the causal effect $\mathcal{D} \rightarrow \mathcal{X}$. On the other hand, these concepts affect the frequency of some related words in the captions $\mathcal{L}$, \eg, the most preferred concept $ID_{\text{Person}}$ in MS COCO largely increases the frequency of the words ``woman'', ``man'', and some human related actions ``sit'', ``stand''. Thus, these concepts affect the captions and we have $\mathcal{D} \rightarrow \mathcal{L}$. When we use the dataset to train a captioner, although we hope to learn the true causal effect $\mathcal{X} \rightarrow \mathcal{L}$, the captioner is likely to learn some spurious correlation between $\mathcal{X}$ and $\mathcal{L}$ due to the confounder $\mathcal{D}$, \ie, over-exploiting the co-occurrence between the recognized concepts and some words to generate the biased captions. For example, as in Fig.~\ref{fig:fig_demo} (c), after recognizing the concept $ID_{\text{Board}}$ through $\mathcal{X} \leftarrow \mathcal{D}$, the captioner will exploit the higher co-occurrence between $ID_{\text{Board}}$ with ``man'' through $\mathcal{D} \rightarrow \mathcal{L}$ to generate a wrong word.

So far, we have shown that $\mathcal{D}$ is a confounder and in Section~\ref{sub_sec:backdoor} we will further show that if the \textbf{observational probability} $P(\mathcal{L}|\mathcal{X})$ is used as the training target, the captioner is likely to learn the spurious correlation. The rest is how to deconfound it ---  \emph{Deconfounded Image Captioning} (\textbf{DIC}) --- which is our goal in this paper. A principled solution is to apply a new \textbf{interventional probability} as the training target: $P(\mathcal{L}|do(\mathcal{X}))$, which is fundamentally different from $P(\mathcal{L}|\mathcal{X})$ due to the \textbf{do-operator} $do(\cdot)$ ~\cite{pearl2016causal,pearl2000causality}. Intuitively, the do-operator cuts off the link $\mathcal{X}\leftarrow \mathcal{D}$ and then the backdoor path: $\mathcal{X} \leftarrow \mathcal{D} \rightarrow \mathcal{L}$ is blocked and the spurious correlation is eliminated. As a result, $\mathcal{X}$ and $\mathcal{L}$ are deconfounded and the captioner can learn the true causal effect $\mathcal{X}\rightarrow\mathcal{L}$. In Section~\ref{sub_sec:backdoor} and~\ref{sub_sec:frontdoor}, we will introduce two fundamental techniques to calculate $P(\mathcal{L}|do(\mathcal{X}))$, which are \textbf{the backdoor and front-door adjustments}~\cite{pearl2016causal,PearlMackenzie18}, respectively. 

In addition to the contribution of introducing causal inference to analyze the dataset bias problem in IC, our technical contribution of this paper is to propose a novel DIC framework called \textbf{DICv1.0} (see Section~\ref{sec:dic}) based on both the backdoor and front-door adjustments. We apply our DIC in two prevailing models: Up-Down~\cite{anderson2018bottom} and AoANet~\cite{huang2019attention}, and help both of them boost the CIDEr-D scores from 126.4 to 129.5 and from 128.7 to
131.1, respectively, where the latter one is submitted to the MS COCO Caption test server and achieves a 128.4 CIDEr c40. Furthermore, another contribution is that we retrospect the major captioners in the 6-year-old IC community from the causal view (see Section~\ref{sec:cau_retro}). Interestingly, we will show how this causal retrospect enlightens us to develop our DICv1.0 as well as pointing us to a new future.

%#################################Preliminaries#############################
\section{Preliminaries: Deconfounding}

We hope to train an image captioner to learn the true causal effect $\mathcal{X}\rightarrow \mathcal{L}$: the captioner should reason the caption $\mathcal{L}$ from the image features $\mathcal{X}$ instead of exploiting the spurious correlations induced by the confounder $\mathcal{D}$. In this section, we review two main deconfounding techniques which help us achieve this goal: the backdoor and front-door adjustments~\cite{pearl2016causal,pearl2000causality}.

\subsection{The Backdoor Adjustment}
\label{sub_sec:backdoor}
We first use the digit classifier example to illustrate that the observational probability $P(\mathcal{L}|\mathcal{X})$ often encourages the classifier to learn spurious correlation. 
By Bayes' theorem, we have
\begin{equation}
\label{equ:equ_bayes_digt}
    P(\mathcal{L}|\mathcal{X})=\sum\nolimits_p{P(\mathcal{L}|\mathcal{X},p)P(p|\mathcal{X})},
\end{equation}
where $p$ denotes the position of the black pixels, which is either ``left'' or ``right''. Now assuming that $\mathcal{X}_0$ denotes the feature of an image containing ``0'' with label $\mathcal{L}_0$ and, by Eq.~\eqref{equ:equ_bayes_digt}, we obtain:
\begin{equation}
\small
\label{equ:equ_bayes_digt2}
\begin{aligned}
 &P(\mathcal{L}_0|\mathcal{X}_0) \\
 = &P(\mathcal{L}_0|\mathcal{X}_0,left)P(left|\mathcal{X}_0) + P(\mathcal{L}_0|\mathcal{X}_0,right)P(right|\mathcal{X}_0).
\end{aligned}
\end{equation}
Due to the position bias in the dataset, we have $P(left|\mathcal{X}_0)\gg P(right|\mathcal{X}_0)$ and thus
\begin{equation}
\small
\label{equ:equ_bayes_digt3}
 P(\mathcal{L}_0|\mathcal{X}_0) \approx P(\mathcal{L}_0|\mathcal{X}_0,left) P(left|\mathcal{X}_0) \approx P(\mathcal{L}_0|left) P(left|\mathcal{X}_0).
\end{equation}
We have the second approximation in Eq.~\eqref{equ:equ_bayes_digt3} because $\mathcal{X}_0$ is learnable and the connection between ``left'' and $\mathcal{L}_0$ is strong. Maximizing $P(\mathcal{L}_0|\mathcal{X}_0)$ over the training set are likely to encourage $\mathcal{X}_0$ to converge to the position feature, since recognizing the position is easier than recognizing the digit shape, which is also referred as short-cut learning~\cite{geirhos2020shortcut}. In this way, the classifier learns the spurious correlation induced by the backdoor path $\mathcal{X} \leftarrow \mathcal{P} \rightarrow \mathcal{L}$.

The backdoor adjustment calculates the interventional distribution $P(\mathcal{L}|do(\mathcal{X}))$ by revising the observational distribution in Eq.~\eqref{equ:equ_bayes_digt} as:
\begin{equation}
\label{equ:equ_do_digt}
    P(\mathcal{L}|do(\mathcal{X}))=\sum\nolimits_p{P(\mathcal{L}|\mathcal{X},p)P(p)}.
\end{equation} 
By comparing Eq.~\eqref{equ:equ_bayes_digt} and Eq.~\eqref{equ:equ_do_digt}, we can find that $P(p|\mathcal{X})$ becomes $P(p)$, which means that the position is not affected by $\mathcal{X}$ while being intervened to be ``left'' or ``right'' regardless of $\mathcal{X}$. Assuming $P(p=left)=P(p=right)=0.5$, for the input $\mathcal{X}_0$, Eq.~\eqref{equ:equ_do_digt} becomes
\begin{equation}
\label{equ:equ_do_digt_1}
    P(\mathcal{L}_0|do(\mathcal{X}_0))=\frac{1}{2} \sum\nolimits_p{P(\mathcal{L}_0|\mathcal{X}_0,p)}.
\end{equation}
Although the probability of having $(\mathcal{X}_0,p=right)$ in the training set (say 10\% or less) is much smaller than that of $(\mathcal{X}_0,p=left)$, it will not be ignored in Eq.~\eqref{equ:equ_do_digt_1} and maximizing Eq.~\eqref{equ:equ_do_digt_1} will encourage $\mathcal{X}_0$ to learn something that is different from the position feature and can lead to $\mathcal{L}_0$.

In the IC case, almost all the modern captioners are trained by the observational distribution $P(\mathcal{L}|\mathcal{X})$, where $\mathcal{X}$ are usually based the features extracted from visual backbones~\cite{ren2015faster}. Various spurious correlations have been observed in the existing captioners as shown in Fig.~\ref{fig:fig_demo}. Treating the frequently appearing concepts as the confounder $\mathcal{D}$, 
we can decompose $P(\mathcal{L}|\mathcal{X})$ as:
\begin{equation}
\label{equ:equ_bayes}
    P(\mathcal{L}|\mathcal{X})=\sum\nolimits_d{P(\mathcal{L}|\mathcal{X},d)P(d|\mathcal{X})},
\end{equation}
where $d$ is a concept ID in $\mathcal{D}$, \eg, $ID_{\text{Apple}}$ or $ID_{\text{Banana}}$. Since a captioning dataset usually contains more samples about certain concepts than the others, \eg, MS COCO contains 80 frequently appearing objects~\cite{lin2014microsoft,chen2015microsoft} and thus these objects are the more preferred concepts than the others, especially the attribute or relation concepts like $ID_{\text{Green}}$, $ID_{\text{Small}}$, $ID_{\text{Sit}}$, or $ID_{\text{Ride}}$.

Then during training, it is much easier for the visual encoders to recognize the objects and neglect the objects' attributes or relations. After recognizing these objects, the captioner might only learn to predict the attributes or relations by their co-occurrences with those preferred concepts instead of really reasoning from the images. For example, if a dataset contains much more ``red apple'' than ``green apple'', after recognizing $ID_{\text{Apple}}$, which means that $P(ID_{\text{Apple}}|\mathcal{X})\approx1$, Eq.~\eqref{equ:equ_bayes} could degrade to $P(\mathcal{L}|ID_{\text{Apple}})$ to make the captioner build strong connection between the colour ``red'' with the concept $ID_{\text{Apple}}$ without seeing the real colour. Fig.~\ref{fig:fig_demo} shows three examples, where $\mathcal{X}\leftarrow\mathcal{D}$ in the causal graph denotes the recognizing of some preferred concepts and $\mathcal{D}\rightarrow\mathcal{L}$ denotes the learning or exploiting of the co-occurrence.

To deconfound the image captioning, we can also use the backdoor adjustment as Eq.~\eqref{equ:equ_do_digt} to split $\mathcal{D}$ into different concepts and calculate the average causal effects at these concepts:
\begin{equation}
\label{equ:equ_do}
    P(\mathcal{L}|do(\mathcal{X}))=\sum\nolimits_d{P(\mathcal{L}|\mathcal{X},d)P(d)}.
\end{equation} 
Then if we train our captioner by maximizing $P(\mathcal{L}|do(\mathcal{X}))$, the captioner will learn to maximize $P(\mathcal{L}|\mathcal{X},d)$ at every concept instead of just one $ID_{\text{Apple}}$, \eg, learning $\mathcal{L}=green$ from $ID_{\text{Banana}}$ as well. In this way, the captioner is encouraged to build more connections between $\mathcal{X}$ and $\mathcal{L}$, \ie, reasoning the caption from the image.

\subsection{The Front-door Adjustment}
\label{sub_sec:frontdoor}
\begin{figure}[t]
\centering
\includegraphics[width=1\linewidth,trim = 5mm 5mm 5mm 5mm,clip]{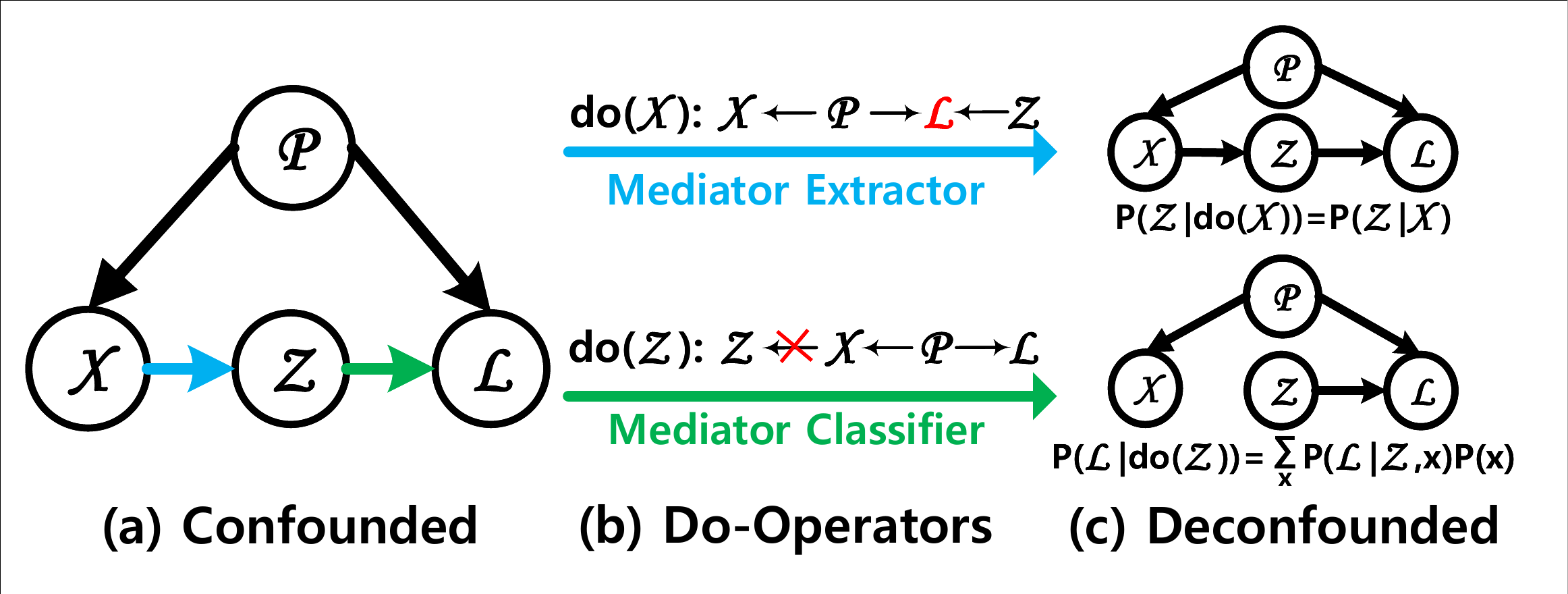}
\caption{The toy example of using the front-door adjustment to solve the biased 0-vs-1 problem. The front-door graph requires an additional mediator $\mathcal{Z}$ to divide the classification process into two parts: a mediator extractor $\mathcal{X}\rightarrow\mathcal{Z}$ and a mediator classifier $\mathcal{Z}\rightarrow\mathcal{L}$. To deconfound this front-door based digit classifier, we should chain together the causal effects of two parts: $P(\mathcal{Z}|do(\mathcal{X}))$ and $P(\mathcal{L}|do(\mathcal{Z}))$ by blocking two backdoor paths: $\mathcal{X} \leftarrow \mathcal{P} \rightarrow \mathcal{L} \leftarrow \mathcal{Z}$ and $\mathcal{Z}\leftarrow\mathcal{X}\leftarrow\mathcal{P}\rightarrow\mathcal{L}$, respectively. The first one is already blocked due to the collider at $\mathcal{L}$ and the second one can be blocked by cutting off the link $\mathcal{X}\rightarrow\mathcal{Z}$.}
\label{fig:fig_frontdoor}
% \vspace{-0.2in}
\end{figure}
In Eq.~\eqref{equ:equ_do}, we have seen that the backdoor adjustment requires us to split $\mathcal{D}$ into different levels. In other words, we need to know what the confounder is in advance in order to apply the backdoor adjustment. However, in real situations, dataset biases could be very complex and it is hard to know and disentangle different types of confounders. For example, in IC, it is an oversimplification to split $\mathcal{D}$ into individual concepts since dataset biases are not only brought by each single concept independently but more likely by the combinations of different concepts, \eg, relation words are affected by both subject and object concepts, for which a sufficient split requires the combinational complexity. More seriously, since researchers usually throw away the pre-training dataset and only use the pre-trained model, they have no idea which concepts appeared more frequently. Thus, it is almost impossible for us to get a reasonable split of $\mathcal{D}$ to apply the backdoor adjustment.

Fortunately, the front-door adjustment~\cite{pearl2016causal} can also be used to calculate $P(\mathcal{L}|do(\mathcal{X}))$ when we cannot split $\mathcal{D}$. 
Here we still use the position-biased digit classification in Fig.~\ref{fig:fig_digit} as the toy example to demonstrate the basic ideas of the front-door adjustment, and then briefly discuss our IC case. As shown in Fig.~\ref{fig:fig_frontdoor}(a), to apply the front-door adjustment, an additional mediator $\mathcal{Z}$ should be inserted between $\mathcal{X}$ and $\mathcal{L}$ to construct a front-door path $\mathcal{X} \rightarrow \mathcal{Z} \rightarrow \mathcal{L}$ for helping transmit knowledge. 

Here we treat ``the shape'' as the mediator $\mathcal{Z}$ in this digit classifier and then this classifier is divided into two parts: a \textbf{shape extractor} $\mathcal{X} \rightarrow \mathcal{Z}$ and a \textbf{shape classifier} $\mathcal{Z} \rightarrow \mathcal{L}$. Then for the conditional probability $P(\mathcal{L}|\mathcal{X})$, we can decompose it by chaining together two parts' conditional probabilities:
\begin{equation}
\label{equ:front_bayes}
    P(\mathcal{L}|\mathcal{X})=\sum\nolimits_{\bm{z}}P(\mathcal{Z}=\bm{z}|\mathcal{X})P(\mathcal{L}|\mathcal{Z}=\bm{z}).
\end{equation} 
Similarly, we can decompose the interventional probability $P(\mathcal{L}|do(\mathcal{X}))$ as:
\begin{equation} \label{equ:front_do1}
    P(\mathcal{L}|do(\mathcal{X}))=\sum\nolimits_{\bm{z}}P(\mathcal{Z}=\bm{z}|do(\mathcal{X}))P(\mathcal{L}|do(\mathcal{Z}=\bm{z})).
\end{equation}
Next, we compare the conditional and interventional probabilities of these two parts to see whether they are the same or not and why the biases are induced.

\noindent\textbf{Shape Extractor $\mathcal{X} \rightarrow \mathcal{Z}$.} As shown in the blue part of Fig.~\ref{fig:fig_frontdoor}(b), for the causal link $\mathcal{X} \rightarrow \mathcal{Z}$, due to the collider at $\mathcal{L}$, the backdoor path between $\mathcal{X}$ and $\mathcal{Z}$: $\mathcal{X} \leftarrow \mathcal{P} \rightarrow \mathcal{L} \leftarrow \mathcal{Z}$ is already blocked. Thus the interventional probability is equal to the conditional one:
\begin{equation} \label{equ:front_do2}
    P(\mathcal{Z}=\bm{z}|do(\mathcal{X})) = P(\mathcal{Z}=\bm{z}|\mathcal{X}).
\end{equation}
The details about why the collider at $\mathcal{L}$ blocks this backdoor path is given in Section~A.2.3 of the supplementary material. For simplicity, we assume this extractor is perfect that it extracts the shape ``circle'' from digit ``0'' and ``line'' from ``1''. 
%and meantime, after observing the extracted shape is ``circle'', we know that the digit must be ``0''. 

\noindent\textbf{Shape Classifier $\mathcal{Z} \rightarrow \mathcal{L}$.} As shown in the green part of Fig.~\ref{fig:fig_frontdoor}(b), the backdoor path between $\mathcal{Z}$ and $\mathcal{L}$: $\mathcal{Z}\leftarrow\mathcal{X}\leftarrow\mathcal{P}\rightarrow\mathcal{L}$ is not naturally blocked, which means the shape classifier is affected by the confounder $\mathcal{P}$. To clearly see how $\mathcal{P}$ affects this classifier, we should  expand its conditional distribution $P(\mathcal{L}|\mathcal{Z})$ as:
\begin{equation} \label{equ:bayes_digit1}
\small
\begin{aligned}
    P(\mathcal{L}|\mathcal{Z}=z) &= \sum\nolimits_{\bm{x}}\sum\nolimits_{p}  P(\mathcal{L},\bm{x},p|z) \\
    &=\sum\nolimits_{\bm{x}}\sum\nolimits_{p}P(\mathcal{L}|\bm{x},p,z)P(\bm{x},p|z) \\
    &=\sum\nolimits_{\bm{x}}\sum\nolimits_{p}P(\mathcal{L}|\bm{x},p,z)P(p|\bm{x},z) P(\bm{x}|z) \\
    &=\sum\nolimits_{\bm{x}}\sum\nolimits_{p}P(\mathcal{L}|p,z)P(p|\bm{x}) P(\bm{x}|z).
\end{aligned}
\footnote{Given the observed $\mathcal{P}=p$ and $\mathcal{Z}=z$, $\mathcal{L}$ and $\mathcal{X}$ are conditional independent: $\mathcal{L} \perp\!\!\!\perp \mathcal{X} | \mathcal{P},\mathcal{Z}$, thus we have $P(\mathcal{L}|\bm{x},p,z)=P(\mathcal{L}|p,z)$. Also, since $\mathcal{Z} \perp\!\!\!\perp \mathcal{P} | \mathcal{X}$, we have $P(p|\bm{x},z)=P(p|\bm{x})$. Both of them establish the forth equation.}
\end{equation}
Now assuming the shape is $\mathcal{Z}=\text{circle}$ and the corresponding label is $\mathcal{L}=\mathcal{L}_0$ :
\begin{equation} \label{equ:bayes_digit2}
\small
    P(\mathcal{L}_0|\text{circle}) =\sum\nolimits_{\bm{x}}\sum\nolimits_p P(\mathcal{L}_0|p,\text{circle})P(p|\bm{x}) P(\bm{x}|\text{circle}).
\end{equation}
Assuming the perfect situation, $P(\mathcal{X}_0|\text{circle})=1$ and $P(\mathcal{X}_1|\text{circle})=0$, with the position bias, $P(\text{left}|\mathcal{X}_0)\gg P(\text{right}|\mathcal{X}_0)$, we have:
\begin{equation} \label{equ:bayes_digit3}
\small
\begin{aligned}
    & P(\mathcal{L}_0|\mathcal{Z}=\text{circle})  \\
    &\approx\sum\nolimits_{\bm{x}\in \mathcal{X}_0}P(\mathcal{L}_0|\text{left},\text{circle})P(\text{left}|\bm{x}) P(\bm{x}|\text{circle}) \\
    & \approx P(\mathcal{L}_0|\text{left},\text{circle}).
\end{aligned}
\end{equation}
For convenience, we use $\mathcal{X}_0$ to denote the feature space of images with ``0'' and $\bm{x}\in \mathcal{X}_0$ to denote the feature $\bm{x}$ from this space. With this shape extractor, the classifier will predict from both ``left'' and ``circle'', which is more robust than the one without this shape mediator (Eq.~\eqref{equ:equ_bayes_digt3}). However, we still run into a similar situation as Eq.~\eqref{equ:equ_bayes_digt3} that if the classifier is trained by maximizing Eq.~\eqref{equ:bayes_digit3}, it does not know whether ``circle'' or ``left'' causes the digit ``0''. As a result, this classifier may be polluted by the position bias as the backdoor case. Moreover, the shape extractor is also imperfect in practice. If the followed classifier learns the position bias, the ahead extractor could also be polluted.
%, \eg, the extractor may do not know which knowledge, shape or position, should be extracted if the followed classifier does not know which knowledge is legitimate. 
As a result, the whole system following $\mathcal{X} \rightarrow \mathcal{Z} \rightarrow \mathcal{L}$ might learn the position bias instead of the causal effect.

To address this, we can cut off $\mathcal{Z} \leftarrow \mathcal{X}$ to block the backdoor path  $\mathcal{Z}\leftarrow\mathcal{X}\leftarrow\mathcal{P}\rightarrow\mathcal{L}$, as shown in the green part of Fig.~\ref{fig:fig_frontdoor}(b). Then, similar to Eq.~\eqref{equ:equ_do}, we have:
\begin{equation} \label{equ:front_do3}
    P(\mathcal{L}|do(\mathcal{Z}=\bm{z}))=\sum_{\bm{x}}{P(\mathcal{L}|\bm{z},\bm{x})P(\bm{x})},
\end{equation}
where $\bm{x}$ represents all the possible representations in the feature space. In Eq.~\eqref{equ:front_do3}, since $\bm{x}$ keeps changing while the shape knowledge does not, the classifier will more easily learn the causal effect, \eg, ``circle'' causes ``0''. Once this classifier knows that ``circle'' is the true cause, the former extractor is also encouraged to extract ``circle'' from the images with ``0''. As a result, both extractor and classifier will learn the true causal effects.

To sum up, by applying Eq.~\eqref{equ:front_do2} and~\eqref{equ:front_do3} into Eq.~\eqref{equ:front_do1}, we have the front-door adjustment as: 
\begin{equation}
\label{equ:equ_frontdoor}
    P(\mathcal{L}|do(\mathcal{X}))= 
    \sum\nolimits_{\bm{z}}P(\bm{z}|\mathcal{X})\sum\nolimits_{\bm{x}}P(\mathcal{L}|\bm{z},\bm{x})P(\bm{x}).
\end{equation}
In our IC case, as we will see in Section~\ref{sec:cau_retro}, researchers have also built some front-door based captioners with different mediators, \eg, the prevailing visual attention mechanism~\cite{xu2015show} treats the features extracted from different positions as the mediator. 

%############################Retrospect#############################
\section{Related Work}
\subsection{A Causal Retrospect}
\label{sec:cau_retro}
\begin{figure*}[t]
\centering
\includegraphics[width=1\linewidth,trim = 5mm 5mm 5mm 5mm,clip]{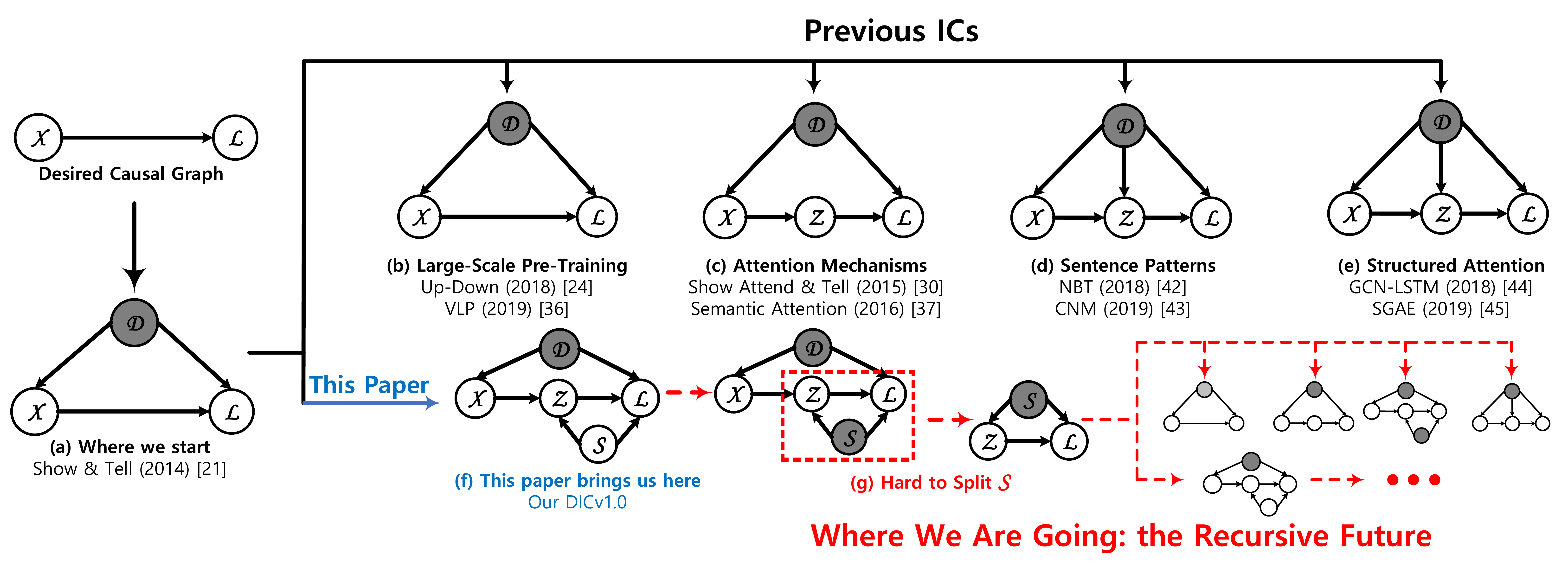}
   \caption{The causal retrospect of the major captioners. The past/present/future are colored by black/blue/red, respectively. We will look ahead the recursive future in Section~\ref{sec:epilogue}. In these causal graphs, the shadowed variables signify that it is hard to split them for applying the backdoor adjustment, \eg, $\mathcal{D}$ in (a) and $\mathcal{S}$ in (g).
   }
\label{fig:fig_retro}
% \vspace{-0.2in}
\end{figure*}
We follow Fig.~\ref{fig:fig_retro} to retrospect the milestone image captioners proposed in recent six years from the causal view, which provides a novel perspective of understanding their contributions. 

\noindent\textbf{Show \& Tell}~\cite{vinyals2015show} (Fig.~\ref{fig:fig_retro}(a)).
Compared with earlier template-based captioners like Baby Talk~\cite{kulkarni2013babytalk}, Show \& Tell is where we start to deploy deep neural network based encoder-decoder for captioning. Also, since Show \& Tell, almost all the following captioners are built on the pre-trained model to narrow the gap between vision and language, while also intensifying the ``preferred concepts'' problem and encouraging the model to learn the spurious correlation.
 
\noindent\textbf{Large-Scale Training} (Fig.~\ref{fig:fig_retro}(b)).
A straightforward way to deconfound is to turn the confounder $\mathcal{D}$ into a non-confounder, \ie, constructing an infinite and balanced dataset. In this way, the captioner can learn every possible visual-semantic pairs from the abundant training samples and as a result, the captioner will no longer collapse into biases. We can approximate this by scaling up the training data. Many studies fall into this large-scale training strategy, \eg, Up-Down~\cite{anderson2018bottom} exploited Visual Genome~\cite{krishna2017visual} with a larger label space, HIP~\cite{yao2019hierarchy} used more fine-grained segmentation annotations, and vision-language BERT frameworks~\cite{lu2019vilbert,zhou2019unified} trained their network by 3 million image-caption pairs from Conceptual Captions~\cite{sharma2018conceptual}. After training or pre-training with those large-scale datasets, these captioners' performances are boosted.
However, it is still almost impossible to collect an infinite and perfect dataset where no bias exists since even the real world is essentially unbalanced. Thus, large-scale training cannot completely deconfound the IC.

\noindent\textbf{Attention Mechanisms} (Fig.~\ref{fig:fig_retro}(c)).
SAT~\cite{xu2015show} was the first work which applied the attention mechanism in IC, and then this mechanism was used in almost every follow-up IC systems~\cite{you2016image,lu2017knowing,yao2017boosting,anderson2018bottom,zha2019context,huang2019attention,cornia2020meshed,zhou2020more}. By this mechanism, captioners first scan over all the visual concepts of an image and then select suitable ones conditioned on the language context, \eg, spatial attention~\cite{xu2015show,anderson2018bottom} and semantic attention~\cite{you2016image} selected the most informative visual features and semantic labels, to generate the captions. The attention mechanism is an approximation of both the backdoor and front-door adjustments. Compared with the backdoor adjustment (Eq.~\eqref{equ:equ_do}), where $P(\mathcal{L}|do(\mathcal{X}))$ averages over all the visual concepts of $\mathcal{D}$, while the attention mechanism only averages over the appearing visual concepts of the given image, which only construct a small subset of $\mathcal{D}$. Thus $\mathcal{D}$ is not totally deconfounded. From the perspective of the front-door adjustment, the attention mechanism treats the visual feature extracted from a specific position as the mediator $\mathcal{Z}$. Then the captioners generate the words by weighted averaging the visual features at the selected positions at different time steps.
However, it uses the observational conditional probability $P(\mathcal{L}|\mathcal{X})$ (Eq.~\eqref{equ:front_bayes}) instead of $P(\mathcal{L}|do(\mathcal{X}))$ (Eq.~\eqref{equ:equ_frontdoor}) to compute the word distribution, which does not deconfound the confounder. Thus these captioners still easily learn the dataset bias~\cite{rohrbach2018object}.
\re{Some researchers~\cite{zhou2020more} use image-text matching as a weakly training supervision for improving attention qualities, while such supervision comes from the confounder $\mathcal{D}$, this captioner is still confounded.}

\noindent\textbf{Sentence Patterns} (Fig.~\ref{fig:fig_retro}(d)).
Researchers also designed ICs which imitate humans to dynamically structure sentence patterns for captioning. In particular, they used diverse modules for different patterns, \eg, NBT~\cite{lu2018neural} designed two modules respectively for nouns and all the other words, and CNM~\cite{yang2019learning} applied four fine-grained modules for nouns, adjectives, relation related words, and function words. During captioning, sentence patterns are dynamically composed for selecting suitable modules to generate the corresponding words. At first glance, this framework looks like a front-door IC where the mediator $\mathcal{Z}$ is sentence pattern. However, if we look closer, it can be discovered that since both the sentence patterns and the captions are learnt from the language resource of the dataset, both $\mathcal{Z}$ and $\mathcal{L}$ are affected by the concepts $\mathcal{D}$ in the language resource, \eg, the more nouns appear, the more \textsc{object} module is trained. However, due to this causal link $\mathcal{D} \rightarrow \mathcal{Z}$, the front-door adjustment cannot be used and thus these captioners are still confounded.

\noindent\textbf{Structured Attention} (Fig.~\ref{fig:fig_retro}(e)).
When humans describe an image, they can build a semantic structure (\eg, scene graph) about this image and then turn this structure into the caption. Inspired by this, researchers~\cite{yao2018exploring,yang2019auto,chen2020say,yang2020auto} proposed to learn a scene graph from the image and to apply Graph Neural Network~\cite{micheli2009neural,atwood2016diffusion,niepert2016learning} to transform this scene graph into a series of embeddings that each embedding corresponds to a sub-structure, \eg, a triplet structure like ``subject-predicate-object''. Then the captioners dynamically attend to these sub-structure embeddings to generate the words. 
\re{Such a strategy is also widely applied in other vision-language tasks like VQA~\cite{teney2017graph,li2019relation}, Grounding~\cite{jing2020visual,liu2019learning}, and Visual Dialogue~\cite{yu2020learning}, \eg,~\cite{yu2020learning} exploits the scene graph containing both visual and semantic knowledge of the image to complete the dialogue.
}
These models also belong to the front-door causal graph where $\mathcal{Z}$ is the structure or the graph. However, such structures are usually affected by $\mathcal{D}$, \eg, the learning and predicting of the relation is also biased by some preferred concepts~\cite{lu2016visual}. Moreover, the applied scene graph parsers~\cite{zellers2018neural} are usually trained by VG~\cite{krishna2017visual}, while VG is also used to pre-train these captioners. Thus we also have the causal link $\mathcal{D} \rightarrow \mathcal{Z}$ and the front-door adjustment cannot be used. Consequently, it is hard to deconfound such captioners and they are still encouraged to learn the spurious correlation.

\subsection{Causal Inference in Computer Vision}
Recently, researchers have begun to exploit the techniques from Causality~\cite{pearl2000causality,rubin2005causal} to design novel methods for solving various learning tasks or develop new perspectives for rethinking certain learning problems. For example, disentangled causal mechanisms are exploited to design more robust disentanglement measurements~\cite{suter2019robustly}, generate counterfactual samples~\cite{abbasnejad2020counterfactual,kocaoglu2017causalgan}, and alleviate negative effects brought by different biases~\cite{veitch2020adapting,tang2020unbiased,wang2020visual,qi2020two,niu2021counterfactual,yang2021causal}. Specifically, both~\cite{wang2020visual} and~\cite{qi2020two} assume that the confounder is observable and then they apply the backdoor adjustment to deconfound. In contrast, the confounder is not observable in our case. Thus, our situation is more challenging and we exploit not only the backdoor but also the front-door adjustments for deconfounding. For~\cite{tang2020unbiased,niu2021counterfactual}, they apply other causality techniques: Natural Indirect Effect and Total Direct Effect for eliminating the mediator effect instead of deconfounding.

\re{
\subsection{Group Caption}
Group-based Image Captioning~\cite{chen2018groupcap,li2020context} also exploits the other image-caption samples for helping generate accurate
and discriminative captions, while they have three major differences compared with our research. First, their target is different from ours. They focus on group-based captioning, which requires the captioner to exploit the relevance and diversity knowledge from the other images, while our DIC focuses on one-line (single image) captioning. Second, due to the different aims, the problem settings are also different. Specifically, their methods need to divide the whole captioning dataset into different image groups, \eg, images in the same group should have similar scene graphs, while our DIC does not require such explicit image groups, but uses more implicit groups, \eg, the same group should contain similar visual concepts (in the backdoor adjustment) or similar visual features (in the front-door adjustment). Third, the technical details are different. Both~\cite{chen2018groupcap} and~\cite{li2020context} exploit the relevance and diversity knowledge of the images from the intra-group and the inter-group, respectively. However, our DIC does not explicitly distinguish the relevance and diversity knowledge, but only measures the average causal effect over each group.
}
%#############################DIC#####################################
\section{Deconfounded Image Captioning}
\label{sec:dic}
In this section, we discuss how to derive our DICv1.0 from the causal retrospect in Section~\ref{sec:cau_retro}. In Section~\ref{sub_sec:choose_z}, we first discuss how to choose the mediator $\mathcal{Z}$ from three candidates, \ie, spatial position, sentence pattern, and graph structure. Then we derive the interventional probability $P(\mathcal{L}|do(\mathcal{X}))$ and discuss the two major challenges of calculating this probability in a deep network. Both challenges are respectively addressed in Section~\ref{sub_sec:wngm_dict} and~\ref{sub_sec:dict}. Lastly, we detail how to implement our DICv1.0 into the prevailing encoder-decoder framework in Section~\ref{sub_sec:arch}.

\subsection{Choosing $\mathcal{Z}$} \label{sub_sec:choose_z}
\begin{figure*}[t]
\centering
\includegraphics[width=1\linewidth,trim = 5mm 5mm 5mm 5mm,clip]{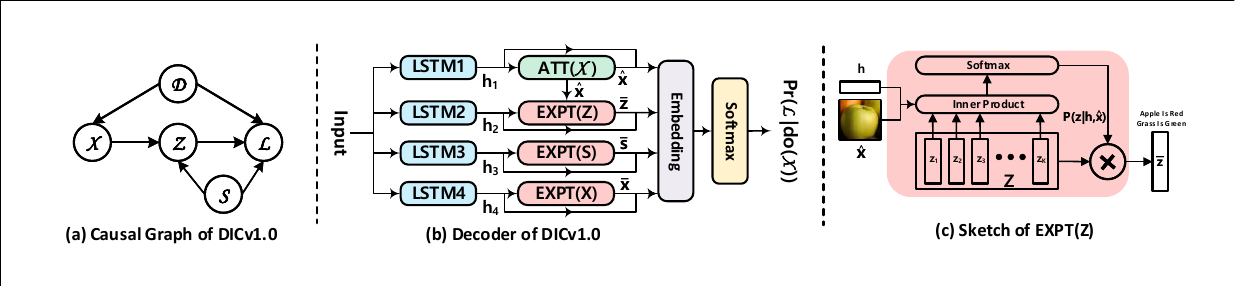}
   \caption{(a) The causal graph of our DICv1.0 captioner, where $\mathcal{Z}$ and $\mathcal{S}$ denote the commonsense structure and the caption vocabulary, respectively. There are two confounders in this model, which are $\mathcal{D}$ and  $\mathcal{S}$. (b) The sketch of our DICv1.0's decoder, where \textsc{att}/\textsc{expt} represent attention/expectation modules. (c) The sketch of $\textsc{expt}(\bm{Z})$ (Eq.~\eqref{equ:expect_mod}), where $\otimes$ denotes the matrix multiplication $\bar{z}=\bm{Z}\bm{p}$.
   }
\label{fig:fig_arch}
\vspace{-0.1in}
\end{figure*}
As discussed in Section~\ref{sub_sec:frontdoor}, since $\mathcal{D}$ is very complex and almost unavailable, the backdoor adjustment cannot be deployed, we have to deconfound the IC by the front-door adjustment. To achieve this, the first challenge is the selection of the mediator $\mathcal{Z}$. Based on the retrospect in Section~\ref{sec:cau_retro}, we know there are three candidates of $\mathcal{Z}$: spatial position (in attention mechanism), sentence pattern, and some kind of the structure (in structured attention). For spatial position (Fig.~\ref{fig:fig_retro}(c)), it only provides the visual concepts of the given image, which is a small subset of $\mathcal{D}$, and thus it is not a good candidate. For sentence pattern (Fig.~\ref{fig:fig_retro}(d)), the calculation of its probability is almost impossible since it is decided by the whole caption, which is not available during the caption generation, so sentence pattern is also not a good candidate. For structured attention (Fig.~\ref{fig:fig_retro}(e)), it is designed to be decided by the image only, while researchers~\cite{yao2018exploring,yang2019auto} learn the structure generator from the same dataset as the captioning, which adds the link $\mathcal{D} \rightarrow \mathcal{Z}$ into the causal graph and as the result, this IC is confounded. However, if we only associate the structure with $\mathcal{X}$, we have a causal graph which cuts off the link $\mathcal{D} \rightarrow \mathcal{Z}$ as Fig.~\ref{fig:fig_retro}(f) and then we can apply the front-door adjustment.

To achieve this, we sample a commonsense structure set from ConceptNet~\cite{liu2004conceptnet} and treat it as the mediator $\mathcal{Z}$, \eg, ``car-at location-road'' and ``apple-is-red''. Given such a commonsense structure ``subject-predicate-object'', we transform it into a semantic embedding $\bm{z}$:
\begin{equation}\label{equ:equ_rela}
    \bm{z}=\text{ReLU}(\text{FC}([\bm{e}_s, \bm{e}_p, \bm{e}_o])),
\end{equation}
where $\bm{e}_s, \bm{e}_p, \bm{e}_o$ denote the trainable word embeddings of the subject, predicate, and object, respectively, and FC denotes the fully connected layer. Compared with the original visual features, these semantic embeddings will naturally be more informative since they also belong to the semantic domain as the captions, which means that this commonsense structure set is a suitable candidate of the mediator. Intuitively, when a captioner uses this commonsense structure as the mediator, this captioner will first use the visual features to retrieve the related semantic embeddings ($\mathcal{X} \rightarrow \mathcal{Z}$) and then exploit the retrieved embeddings to generate the caption ($\mathcal{Z} \rightarrow \mathcal{L}$).

However, when we sample the related commonsense structures $\mathcal{Z}$, we use the words from the caption vocabulary $\mathcal{S}$ as the keys (see Section~\ref{subsec:dataset}), \eg, ``person'' is used to collect the triplets ``person-related to-human'' or ``person-capable of-run''. Therefore, the keywords affect the collections of the triplets and thus we have $\mathcal{S} \rightarrow \mathcal{Z}$. On the other hand, since the vocabulary $\mathcal{S}$ provides the words to compose the caption $\mathcal{L}$, the causal link $\mathcal{S} \rightarrow \mathcal{L}$ is naturally built. Consequently, $\mathcal{S}$ becomes the common cause of both $\mathcal{Z}$ and $\mathcal{L}$, which means that $\mathcal{S}$ is an additional confounder between them. Then, if a keyword, \eg, ``person'', is frequently used to sample the triplets, the words which are closely related to this keyword, \eg, ``man'', or ``woman'', will also appear more frequently in the captions. Thus, the additional biases are induced. 

Now we have the complete causal graph in Fig.~\ref{fig:fig_arch}(a) where two confounders exist: $\mathcal{D}$ and $\mathcal{S}$. To calculate the causal effect $P(\mathcal{L}|do(\mathcal{X}))$, both two confounders should be deconfounded: $\mathcal{D}$ can be deconfounded by the front-door adjustment (see Section~\ref{sub_sec:frontdoor}) and $\mathcal{S}$ can be deconfounded by the backdoor adjustment (Section~\ref{sub_sec:backdoor}) since $\mathcal{S}$ is an observable vocabulary.

This is our \textbf{DICv1.0} which deconfounds image captioning. Briefly, we provide the formula of $P(\mathcal{L}|do(\mathcal{X}))$ here and the detail derivations are given in Section~B.1 of the supplementary material:
\begin{equation}
\begin{aligned}
\label{equ:equ_decon_nic}
    &P(\mathcal{L}|do(\mathcal{X})) \\
    = &\sum\nolimits_{\bm{s}}P(\bm{s})\sum\nolimits_{\bm{x}}P(\bm{x})\sum\nolimits_{\bm{z}}P(\bm{z}|\mathcal{X})[P(\mathcal{L}|\bm{s},\bm{x},\bm{z})] \\
    = &\mathbb{E}_{\bm{s}}\mathbb{E}_{\bm{x}}\mathbb{E}_{[\bm{z}|\mathcal{X}]}[P(\mathcal{L}|\bm{s},\bm{x},\bm{z})].
\end{aligned}
\end{equation}
It can be found that it is the expected value of $P(\mathcal{L}|\bm{s},\bm{x},\bm{z})$ according to three variables $\bm{s}$, $\bm{x}$, and $\bm{z}$, which denote the word embeddings of the keywords, the visual features, and the semantic embeddings, respectively. Notice that $\mathcal{X}$ denotes the current image which we want to generate the caption from and $\bm{x}$ denotes the potential visual feature of the whole representation space.

To implement our DICv1.0 into the encoder-decoder framework, we parameterize $P(\mathcal{L}|do(\mathcal{X}))$ as a network whose last layer is a Softmax layer that implements $P(\mathcal{L}|\bm{s},\bm{x},\bm{z})$ as:
\begin{equation}
\label{equ:equ_network}
P(\mathcal{L}|\bm{s},\bm{x},\bm{z}) = \text{Softmax}[g(\bm{s},\bm{x},\bm{z})],
\end{equation}
where $g(\cdot)$ is an embedding layer before the Softmax layer, \eg, it can be an FC layer. 
However, this brings one challenge that in order to compute $P(\mathcal{L}|do(\mathcal{X}))$ in Eq.~\eqref{equ:equ_decon_nic}, we need a huge number of outputs sampled from this network to get the expectations of $P(\mathcal{L}|\bm{s},\bm{x},\bm{z})$. To solve this challenge, we propose a two-step approximation which allows us to forward the network only once to get an estimation of $P(\mathcal{L}|do(\mathcal{X}))$. The first step is called Normalized Weighted Geometric Mean (NWGM) approximation~\cite{xu2015show,srivastava2014dropout,baldi2014dropout} which absorbs the expectations into the network (see Section~\ref{sub_sec:wngm_dict}). The second step is to sample finite $\bm{s}$, $\bm{x}$, and $\bm{z}$ from the respective representation spaces of $\mathcal{S}$, $\mathcal{X}$, and $\mathcal{Z}$ for estimating the expectations in Eq.~\eqref{equ:equ_decon_nic} (see Section~\ref{sub_sec:dict}). 

\subsection{Normalized Weighted Geometric Mean}
\label{sub_sec:wngm_dict}
By NWGM approximation~\cite{xu2015show}, the expectation of a Softmax unit is approximated as the Softmax of the expectation: 
\begin{equation}
\label{equ:equ_nwgm1}
\begin{aligned}
    &P(\mathcal{L}|do(\mathcal{X})) \\ 
    &= \mathbb{E}_{\bm{s}}\mathbb{E}_{\bm{x}}\mathbb{E}_{[\bm{z}|\mathcal{X}]}\{\text{Softmax}[g(\bm{s},\bm{x},\bm{z})]\} \\
    & \approx \text{Softmax}\{ \mathbb{E}_{\bm{s}}\mathbb{E}_{\bm{x}}\mathbb{E}_{[\bm{z}|\mathcal{X}]}[g(\bm{s},\bm{x},\bm{z})]\}.
\end{aligned}
\end{equation}
Furthermore, if $g(\cdot)$ is an FC layer, we have:
\begin{equation}
\label{equ:equ_nwgm2}
P(\mathcal{L}|do(\mathcal{X})) \approx \text{Softmax}\{ g(\mathbb{E}_{\bm{s}}[\bm{s}],\mathbb{E}_{\bm{x}}[\bm{x}],\mathbb{E}_{[\bm{z}|\mathcal{X}]}[\bm{z}])\} .
\end{equation}
Here we can put the expectations into the FC layer $g(\cdot)$ because the linear projection of the expectation of one variable equals to the expectation of the linear projection of that variable. More details about the derivations of Eq.~\eqref{equ:equ_nwgm1} and~\eqref{equ:equ_nwgm2} by NWGM approximation are given in Section~B.2 of the supplementary material.

\subsection{EXPT Modules for Expectations}
\label{sub_sec:dict}
Before this section, we formulate all the probability equations at the sentence level, \ie, $\mathcal{L}$ denotes a whole caption. When we compute the probabilities at word level, for convenience, we still use $\mathcal{L}$ to denote the word and write the word distribution as $P(\mathcal{L}|do(\mathcal{X}),\bm{h})$, where $\bm{h}$ is a context vector which accumulates the knowledge of the previously generated words. By modifying Eq.~\eqref{equ:equ_decon_nic} (see Section~B.1 and B.2 of the supplementary material), the word distribution becomes:
\begin{equation}
\begin{aligned}
\label{equ:equ_decon_nic_hid1}
     &P(\mathcal{L}|do(\mathcal{X}),\bm{h}) \\
    = &\sum\nolimits_{\bm{s}}P(\bm{s}|\bm{h})\sum\nolimits_{\bm{x}}P(\bm{x}|\bm{h})\sum\nolimits_{\bm{z}}P(\bm{z}|\mathcal{X},\bm{h})[P(\mathcal{L}|\bm{s},\bm{x},\bm{z},\bm{h})] \\
    = &\mathbb{E}_{[\bm{s}|\bm{h}]}\mathbb{E}_{[\bm{x}|\bm{h}]}\mathbb{E}_{[\bm{z}|\mathcal{X},\bm{h}]}[P(\mathcal{L}|\bm{s},\bm{x},\bm{z},\bm{h})].
\end{aligned}
\end{equation}
Similar to Eq.~\eqref{equ:equ_nwgm1} and~\eqref{equ:equ_nwgm2}, we can put the expectations into the Softmax by NWGM approximation:
\begin{equation}
\begin{aligned}
\label{equ:equ_decon_nic_hid2}
    &P(\mathcal{L}|do(\mathcal{X}),\bm{h}) \\
    &\approx \text{Softmax}\{ g(\mathbb{E}_{[\bm{s}|\bm{h}]}[\bm{s}],\mathbb{E}_{[\bm{x}|\bm{h}]}[\bm{x}],\mathbb{E}_{[\bm{z}|\mathcal{X},\bm{h}]}[\bm{z}])\}.
\end{aligned}
\end{equation}
Then we should calculate: $\mathbb{E}_{[\bm{s}|\bm{h}]}[\bm{s}]$, $\mathbb{E}_{[\bm{x}|\bm{h}]}[\bm{x}]$, and $\mathbb{E}_{[\bm{z}|\mathcal{X},\bm{h}]}[\bm{z}]$.
Here we use $\mathbb{E}_{[\bm{x}|\bm{h}]}[\bm{x}]$ as the example to show how to design an \textbf{\textsc{expt}} module for estimating its value. Mathematically, we have:
\begin{equation}
\label{equ:exp1}
\mathbb{E}_{[\bm{x}|\bm{h}]}[\bm{x}] = \sum\nolimits_{\bm{x}} P(\bm{x}|\bm{h}) \bm{x}.
\end{equation}
However, this sum is intractable since both the representation space of image feature $\mathcal{X}$ and the conditional distribution $P(\bm{x}|\bm{h})$ are so complex that we can not get a closed form solution. To estimate it, we apply two approximations to design our EXPT module, which are \textbf{learning the bases of the representation space $\mathcal{X}$} and \textbf{calculating the conditional distribution $P(\bm{x}|\bm{h})$ by the key-query operation}. We apply a dictionary learning algorithm~\cite{mairal2009online} to learn a basis dictionary $\bm{X}=\{\bm{x}_1,\bm{x}_2,...,\bm{x}_K\}$ from the representations of all the training images, where each element $\bm{x}_k$ is a basis vector. After learning this basis dictionary, we use it to approximate the whole representation space:
\begin{equation}
    \mathbb{E}_{[\bm{x}|\bm{h}]}[\bm{x}] = \sum\nolimits_k P(\bm{x}_k|\bm{h}) \bm{x}_k.
\end{equation}

For the conditional distribution $P(\bm{x}|\bm{h})$, we deploy the efficient query-key operation to approximate it, where $\bm{h}$ acts as the query and the elements in $\bm{X}$ as the key. This query-key operation is a routine operation used in the attention mechanism~\cite{xu2015show} and the memory network~\cite{miller2016key}. To sum up, we have $\textsc{expt}(\bm{X},\bm{h})$ to estimate $\mathbb{E}_{[\bm{x}|\bm{h}]}[\bm{x}]$:
\begin{equation} \label{equ:expect_mod}
\small
\begin{aligned}
 \textbf{Input:} \quad  &\bm{X}=\{\bm{x}_1,\bm{x}_2,...,\bm{x}_K\}, {}\bm{h} \\
 \textbf{Probability:} \quad  &
 \begin{aligned} &q_k={\bm{x}}_k^T\bm{h},\\
  &\bm{p}=\{p_1,...,p_K\}=\text{Softmax}(\{q_1,...,q_K\})
  \end{aligned}
  \\
 \textbf{Output:} \quad  &\mathbb{E}_{[\bm{x}|\bm{h}]}[\bm{x}] \approx \sum\nolimits_k{p_k{\bm{x}}_k}.
\end{aligned}
\end{equation}
Since all the operations of \textsc{expt} module are differentiable, $\bm{X}$ can also be updated during the end-to-end training. Similarly, as in Fig.~\ref{fig:fig_arch}(c), we use $\textsc{expt}(\bm{Z},\text{Cat}(\hat{\bm{x}},\bm{h}))$ to estimate $\mathbb{E}_{[\bm{z}|\mathcal{X},\bm{h}]}[\bm{z}]$, where $\bm{Z}$ is the semantic embedding dictionary that each element $\bm{z}$ of it is computed by Eq.~\eqref{equ:equ_rela}; $\text{Cat}(\cdot)$ denotes the concatenation operation; and $\hat{\bm{x}}$ is the attended feature of the image $\mathcal{X}$ at the current time step (see Section~\ref{sub_sec:arch}). When we collect the commonsense structure set, we use the nouns, verbs, and adjectives of the caption vocabulary as the keywords to search the related structures from ConceptNet. Meanwhile, these keywords' trainable embeddings construct the dictionary $\bm{S}$, which is used in $\textsc{expt}(\bm{S},\bm{h})$ to estimate $\mathbb{E}_{[\bm{s}|\bm{h}]}[\bm{s}]$. The numbers of elements of $\bm{X}$, $\bm{Z}$, and $\bm{S}$ are set to 10,000, 9,590, and 1,342.

\subsection{Network Architectures}
\label{sub_sec:arch}
We incorporate our DICv1.0 into two models: Up-Down~\cite{anderson2018bottom} and AoANet~\cite{huang2019attention} and name them as \textbf{UD-DICv1.0} and \textbf{AoA-DICv1.0}, respectively. In both models, the visual encoder is a ResNet-101 Faster R-CNN~\cite{ren2015faster} pre-trained on Visual Genome~\cite{krishna2017visual} as in Up-Down~\cite{anderson2018bottom}. 
The decoders of the two models have similar architecture:
\begin{equation} \label{equ:top_down}
\small
\begin{aligned} 
\textbf{Input:} \quad &\bm{u}^t=\text{Cat}(\tilde{\bm{x}}, \bm{e}^{t-1}, \bm{o}^{t-1}) \\
 \textbf{LSTM:} \quad  &\bm{h}_*^t=\text{LSTM}_*(\bm{u}^t;\bm{h}_*^{t-1}), \\
 \textbf{ATT:} \quad &\hat{\bm{x}}^t = \text{ATT}(\mathcal{X}, \bm{h}_1^t), 
 \\
 \textbf{EXPT:} \quad  &
 \begin{aligned} \bar{\bm{z}}^t &= \text{EXPT}(\bm{Z}, \text{Cat}(\hat{\bm{x}}^t,\bm{h}_2^t)),\\
 \bar{\bm{s}}^t &= \text{EXPT}(\bm{S}, \bm{h}_3^t) \\
 \bar{\bm{x}}^t &= \text{EXPT}(\bm{X}, \bm{h}_4^t)
  \end{aligned}
 \\
 \textbf{Output:} \quad  & P^t=\text{Softmax}(g(\hat{\bm{x}}^t,\bar{\bm{z}}^t,\bar{\bm{s}}^t,\bar{\bm{x}}^t,\bm{h}_*^t)), \\
\end{aligned}
\end{equation}
which is sketched in Fig.~\ref{fig:fig_arch}(b). The input $\bm{u}^t$ of this decoder concatenates three terms: $\tilde{\bm{x}}$ is the mean pooling of the visual features of the image $\mathcal{X}$, $\bm{e}^{t-1}$ is the previous generated word's embedding, and $\bm{o}^{t-1}$ is the previous embedding layer's output. The subsequent four LSTM layers generate four index vectors $\bm{h}_*$\footnote{For convenience, we omit all the superscript $t$, which represents the $t$-th time step, in the following formulas.}, which are used in the following ATT and EXPT modules. $\text{ATT}(\mathcal{X})$ represents an attention module which computes the attended vector $\hat{\bm{x}}$ from the visual feature $\mathcal{X}$. This $\hat{\bm{x}}$ is input into the \textsc{expt} module ($\textsc{expt}(\bm{Z},[\hat{\bm{x}},\bm{h}_2]$) to estimate $\mathbb{E}_{[\bm{z}|\mathcal{X},\bm{h}]}[\bm{z}]$ and into the embedding layer ($g(\cdot)$) since it is also a sample of the visual features. $\bar{\bm{z}}$, $\bar{\bm{s}}$, and $\bar{\bm{x}}$ denote the estimated vectors of $\mathbb{E}_{[\bm{z}|\mathcal{X},\bm{h}]}[\bm{z}]$, $\mathbb{E}_{[\bm{s}|\bm{h}]}[\bm{s}]$, and $\mathbb{E}_{[\bm{x}|\bm{h}]}[\bm{x}]$ by the corresponding \textsc{expt} modules, respectively. All of them with $\bm{h}_*$ are input into the embedding layer for computing the word distributions $P$.

The key differences of UD-DICv1.0 and AoA-DICv1.0 are that they use distinctive attention networks (ATT) and embedding layers $g(\cdot)$. In UD-DICv1.0, ATT is the Top-Down Attention~\cite{anderson2018bottom}:
\begin{equation} \label{equ:attend}
\small
\begin{aligned}
 \textbf{Input:} \quad  &\mathcal{X}=\{\bm{x}_1,\bm{x}_2,...,\bm{x}_N\}, \bm{h}_1,\\
 \textbf{Attention Weights:}  \quad  &
 \begin{aligned} &a_n=\bm{\omega}_{a}^T\tanh({\bm{W}_x\bm{x}_n+\bm{W}_h\bm{h}_1}),\\
  &\bm{\alpha}^t= \{\alpha_1, \alpha_2,..., \alpha_N\}=\text{softmax}(\bm{a}), 
  \end{aligned}
  \\
 \textbf{Output:} \quad  &\hat{\bm{x}}=\sum_{n=1}^N\alpha_n\bm{x}_n,, \\
\end{aligned}
\end{equation}
where $\bm{W}_x$, $\bm{W}_h$ are trainable matrices and $\bm{\omega}_{a}$ is a trainable vector. Here $g(\cdot)$ is an LSTM layer. In AoA-DICv1.0, ATT is the Attention on Attention~\cite{huang2019attention}:
\begin{equation} \label{equ:aoa_attend}
\small
\begin{aligned}
 \textbf{Input:} \quad  &\mathcal{X}=\{\bm{x}_1,\bm{x}_2,...,\bm{x}_N\}, \bm{h}_1,\\
 \textbf{MH-ATT:}  \quad  &
 \begin{aligned}
\textbf{head}_k= & \text{Softmax}( \frac{\bm{h}_1\bm{W}_k^1(\mathcal{X}\bm{W}_k^2)^T}{\sqrt{d_k}} )\mathcal{X}\bm{W}_k^3, \\
\mathcal{M} = & \text{Cat}(\textbf{head}_1,...,\textbf{head}_8)\bm{W}_{C}, \\
\hat{\bm{v}}= & \text{LeakyReLU}(\text{MLP}(\mathcal{M})),
\end{aligned}
  \\
 \textbf{Info-Vector:} \quad  &\bm{v}^i=\bm{W}_h^i\bm{h}_1+\bm{W}_{\mathcal{X}}^i\hat{\bm{v}}+\bm{b}^i, \\
 \textbf{ATT-Gate:} \quad  &
 \bm{v}^g=\sigma(\bm{W}_h^g\bm{h}_1+\bm{W}_{\mathcal{X}}^g\hat{\bm{v}}+\bm{b}^g),
 \\
 \textbf{Output:} \quad  &\hat{\bm{x}}=\bm{v}^i .*\bm{v}^g,
\end{aligned}
\end{equation}
where $\bm{W}_k^1,\bm{W}_k^2,\bm{W}_k^3,\bm{W}_C,\bm{W}_h^i,\bm{W}_{\mathcal{X}}^i,\bm{b}^i,\bm{W}_h^g,\bm{W}_{\mathcal{X}}^g,\bm{b}^g$ are all trainable parameters and $.*$ in the last row denotes element-wise product. In this case, $g(\cdot)$ is a GLU~\cite{dauphin2017language} layer. 

Notice that although the embedding layer is not an FC layer, we still observe less bias and better performances compared with the original models (see Section~\ref{subsub_sec:ablation_architecture}). More details of the network architectures of UD-DICv1.0 and AoA-DICv1.0, like the size of each layer in these modules, are given in Section~C of the supplementary material.

\subsection{Training Objectives}
When training both models, we minimize the cross-entropy loss by using $P(\mathcal{L}^*|do(\mathcal{X}))$ as the target in the first 35 epochs:
\begin{equation}
\small
     L_{XE} = -\log P(\mathcal{L}^*|do(\mathcal{X})),
\label{equ:equ_celoss}
\end{equation}
where $\mathcal{L}^*$ denotes the ground-truth caption. After that, we minimize the RL-based loss~\cite{rennie2017self} to train both models for another 65 epochs:
\begin{equation}
\small
    L_{RL} = -\mathbb{E}_{{\mathcal{L}^s} \sim P(\mathcal{L}|do(\mathcal{X})}[r(\mathcal{L}^s;\mathcal{L}^*)],
    \label{equ:equ_rlloss}
\end{equation}
where $r$ is a sentence-level metric between the sampled sentence $\mathcal{L}^s$ and the ground-truth $\mathcal{L}^{*}$, \eg, the CIDEr-D metric~\cite{vedantam2015cider}. We use Adam optimizer~\cite{kingma2014adam} to train both models. The learning rate for both models is initialized to $5e^{-4}$ and is decayed by $0.8$ for every $5$ epochs. Importantly, the learning rate of all the dictionaries in \textsc{expt} modules, \eg, $\bm{X}$, $\bm{Z}$, and $\bm{S}$, are set 10 times smaller than the other layers, \eg, $5e^{-5}$. The batch sizes for UD-DICv1.0 and AoA-DICv1.0 are set to 100 and 10, respectively. When inference, we use beam search with a beam size of 5. 

%####################################Experiments#######################
\section{Experiments}
\subsection{Datasets}
\label{subsec:dataset}
\noindent\textbf{MS COCO~\cite{chen2015microsoft}.} We validated our models on MS COCO IC dataset. In particular, our models were tested on two different splits: Karpathy split~\cite{karpathy2015deep} and the official online test split, which divide the whole dataset into $113,287/5,000/5,000$ and $82,783/40,504/40,775$ images for training/validation/test, respectively. We followed previous researches to pre-process our captions~\cite{anderson2018bottom,yang2019auto}. At last we trimmed each caption to a maximum of $16$ words and had a vocabulary of $10,369$ words by removing the words which appear less than $5$ times.

\noindent\textbf{ConceptNet~\cite{liu2004conceptnet}.}
ConceptNet has abundant commonsense structures which have the form of ``subject-predicate-object'' and each of them is assigned with an importance weight. We used the nouns, verbs, and adjectives which appear more than 20 times in MS COCO captioning dataset as the keywords to collect the related structures and all these 1,342 keywords were used to construct the dictionary $\bm{S}$. For each commonsense structure in ConceptNet, if its importance weight is larger than 2.5 and if both the subject and object of this structure are our keywords, we collected it. Finally, we had a semantic structure set with 9,590 structures and they were used to build the dictionary $\bm{Z}$. 

\subsection{Ablation Studies}
\label{sub_sec:ablation}
\subsubsection{Deconfounding Techniques}
\label{subsub_sec:ablation_architecture}
We used Up-Down as the backbone, whose architecture is the same as Fig.~\ref{fig:fig_arch} (b) but without the bottom three parallel LSTM layers and all the \textsc{expt} modules, and gradually incorporated new components into this backbone to design various ablation studies for validating the importance of different deconfounding techniques: the backdoor adjustment (Section~\ref{sub_sec:backdoor}), the front-door adjustment (Section~\ref{sub_sec:frontdoor}), and our DICv1.0 (Section~\ref{sub_sec:arch}).

\noindent\textbf{Comparing Methods.}

\noindent\textbf{UD:} We re-implemented Up-Down~\cite{anderson2018bottom} as our baseline, where only $\textsc{att}(\mathcal{X})$ exists in the decoder. 

\noindent\textbf{UD-BD:} Compared with UD, we followed the backdoor adjustment (Eq.~\eqref{equ:equ_do}) and estimated $\mathbb{E}_{[d|\bm{h}]}[d]$ by $\textsc{expt}(\bm{D},\bm{h})$ (Eq.~\eqref{equ:expect_mod}). We grouped the word embeddings of 80 visual concepts in MS COCO as the dictionary $\bm{D}$ since they are the most frequently appearing objects in the images and captions, which are likely to induce the spurious correlation. This baseline was designed to confirm the utility of the backdoor adjustment.

\noindent\textbf{UD-BD/$\bm{S}$ \& UD-BD/CC:} To test the effects of using more elements to split the confounder $\mathcal{D}$, we deploy UD+EXPT$(\bm{S},\bm{h})$ and UD+EXPT$(\bm{D}_C,\bm{h})$, respectively. $\bm{S}$ is the vocabulary with 1,342 keywords and $\bm{D}_C$ contains $20,932$ triplets sampled from ConceptNet~\cite{liu2004conceptnet}, which is equal to the size of dictionaries used in UD-DICV1.0 (10,000/9,590/1,342 elements in $\bm{X}$/$\bm{Z}$/$\bm{S}$).

\noindent\textbf{UD-FD/Cor:}
Compared with UD, we added $\textsc{expt}(\bm{Z},\text{Cat}(\hat{\bm{x}},\bm{h}))$ into the decoder. We can treat this as training a front-door IC by the observational correlation $P(\mathcal{L}|\mathcal{X})$ (Eq.~\eqref{equ:front_bayes}) and thus the spurious correlation still exists.

\noindent\textbf{UD-FD:} Compared with UD-FD/Cor, we added $\textsc{expt}(\bm{X},\bm{h})$ into the decoder. 
This can be treated as training a front-door IC by the NWGM approximation of $P(\mathcal{L}|do(\mathcal{X}))$ (Eq.~\eqref{equ:equ_frontdoor}) while neglecting the confounder $\mathcal{S}$. This baseline was used to confirm the utility of the front-door adjustment. We also expanded $\mathcal{Z}$ with additional 5,000 commonsense triplets to validate that whether more parameters bring consistent improvements as UD-DICv1.0.

\noindent\textbf{UD-DICv1.0:} Compared with UD-FD, we added $\textsc{expt}(\bm{S},\bm{h})$ into the decoder to get the integral framework of our DICv1.0 as in Fig.~\ref{fig:fig_arch}. This equals to train a front-door IC by using the NWGM approximation of Eq.~\eqref{equ:equ_decon_nic}, which deconfounds both two confounders: $\mathcal{D}$ and $\mathcal{S}$.

\noindent\textbf{Analyses of the Similarities.}
\begin{table*}[t]
\begin{center}
\caption{The performances of various ablation studies on MS COCO Karpathy split. The metrics: B@4, M, R, C, S, CHs, CHi, A@Gen, A@Attr, A@Act, and A@Quan denote BLEU@4, METEOR, ROUGE-L, CIDEr-D, SPICE, CHAIRs, CHAIRi, the accuracy of gender, attribute, action, and quantifier words. The symbols $\uparrow$ and $\downarrow$ mean the higher the better and the lower the better, respectively.}
\vspace{-0.1in}
\label{table:tab_kap_baseline}
\scalebox{1}{
\begin{tabular}{l c c c c c c c c c c c}
		\hline
		   Models   & B@4$\uparrow$ & M$\uparrow$ & R$\uparrow$ &  C$\uparrow$ & S$\uparrow$ & CHs$\downarrow$ & CHi$\downarrow$ & A@Gen$\uparrow$ & A@Attr$\uparrow$ & A@Act$\uparrow$ & A@Quan$\uparrow$\\ \hline
           UD           & $37.2$& $27.5$ & $57.3$ & $125.3$ & $21.1$ & $13.7$ & $8.9$ & $0.81$ & $0.41$ & $0.52$ &  $0.46$\\ 
           UD-BD       &$38.2$& $28.2$ & $58.0$ & $126.9$ & $21.3$ & $11.2$ & $7.6$ & $0.87$ & $0.50$ & $0.56$ &  $0.51$\\ 
           UD-BD/$\bm{S}$ &$38.4$& $28.3$ & $58.2$ & $127.1$ & $21.3$ & $11.0$ & $7.5$ & $0.87$ & $0.51$ & $0.57$ & $0.51$\\
           UD-BD/CC & $38.5$& $28.5$ & $58.6$ & $127.5$ & $21.6$ & $10.9$ & $7.4$ & $0.88$ & $0.52$ & $0.57$ &  $0.53$\\ 
           UD-FD/Cor  &$38.0$& $28.1$ & $58.0$ & $126.5$ & $21.1$ & $12.3$ & $8.3$ & $0.83$ & $0.46$ & $0.54$ &  $0.50$\\ 
           UD-FD       &$38.5$& $28.4$ & $58.7$ & $127.6$ & $21.8$ & $10.5$ & $7.0$ & $0.89$ & $0.55$ & $0.58$ &  $0.55$\\ 
           UD-FD/Ex    &$38.8$& $28.6$ & $58.9$ & $128.0$ & $21.9$ & $10.3$ & $6.8$ & $0.89$ & $0.56$ & $0.58$ &$0.56$\\ 
           UD-DICv1.0    & $\bm{39.0}$ & $\bm{28.8}$ & $\bm{58.8}$ & $\bm{128.8}$ & $\bm{22.0}$ & $\bm{10.1}$ & $\bm{6.5}$ & $\bm{0.91}$ & $\bm{0.58}$ & $\bm{0.60}$ & $\bm{0.57}$\\ \hline 
\end{tabular}
}
\end{center}
\vspace{-0.2in}
\end{table*}
\begin{table}[t]
\begin{center}
\re{
\caption{The results (\%) of human evaluations on bias and attention. In each row, ``Left'' and ``Right'' denote that which method is considered to be better, respectively, and ``Comparative'' denotes that two methods are considered to be similar.}
\label{table:tab_human_evaluation}
\scalebox{0.95}{
\begin{tabular}{c | c c c c }
		\hline
		    Terms & Models & Left & Right &  Comparative \\ \hline
           \multirow{3}{*}{Bias} & UD vs. UD-BD & 21 & 52 & 27 \\ 
           & UD vs. UD-DICv1.0 & 12 & 70 & 18   \\ 
           & UD-BD vs. UD-DICv1.0 & 14 & 61 & 25\\ \hline
           Attention & UD vs. UD-DICv1.0 & 17 & 68 & 15\\
           \hline
\end{tabular}
}
}
\end{center}
\vspace{-0.2in}
\end{table}
We followed previous studies to use the following five metrics: CIDEr-D~\cite{vedantam2015cider}, BLEU~\cite{papineni2002bleu}, METEOR\cite{banerjee2005meteor}, ROUGE~\cite{lin2004rouge}, and SPICE~\cite{anderson2016spice} to measure the similarities between the generated captions with the ground-truth captions. The similarity metrics of our UD-DICv1.0 and the baselines are reported in Table~\ref{table:tab_kap_baseline}. Since the ground-truth captions are annotated by us humans and we are usually good at grasping the causal effect, similarity metrics are naturally suitable to evaluate whether a captioner grasps such causal effect. 

Compared with the original UD, UD-DICv1.0 improves CIDEr-D from 125.3 to 128.8, indicating that UD-DICv1.0 generates the most similar captions with the ground-truth. By comparing UD-BD with UD, we observe that UD-BD achieves a higher CIDEr-D score, which confirms the utility of the backdoor adjustment even in this baseline we only use 80 visual concepts as the split of $\mathcal{D}$, which is a simplification of the real situation.
By comparing UD-BD, UD-BD/$\bm{S}$, and UD-BD/CC, we can find that the model with a larger dictionary has better similarity scores, \eg, UD-BD/CC achieves the highest CIDEr-D, 127.5, among them.

By sequentially comparing UD-FD/Cor, UD-FD, and UD-DICv1.0, we can observe that the performances are also sequentially improved, which confirms the effectiveness of deconfounding $\mathcal{D}$ (UD-FD vs. UD-FD/Cor) and $\mathcal{S}$ (UD-DICv1.0 vs. UD-FD). Also, UD-DICv1.0 achieves better performances than UD-FD/EX that has the same dictionary size with UD-DICv1.0, suggesting that deconfounding $\mathcal{S}$ is more effective than using more parameters in the dictionary. To sum up, it is safe to conclude that incorporating a mediator like the commonsense knowledge and exploiting additional resource like ConceptNet is not enough for generating high-quality captions unless we discover the hidden confounders and use the backdoor and front-door adjustments to deconfound them.

\begin{figure}[t]
\centering
\includegraphics[width=1\linewidth,trim = 5mm 5mm 5mm 5mm,clip]{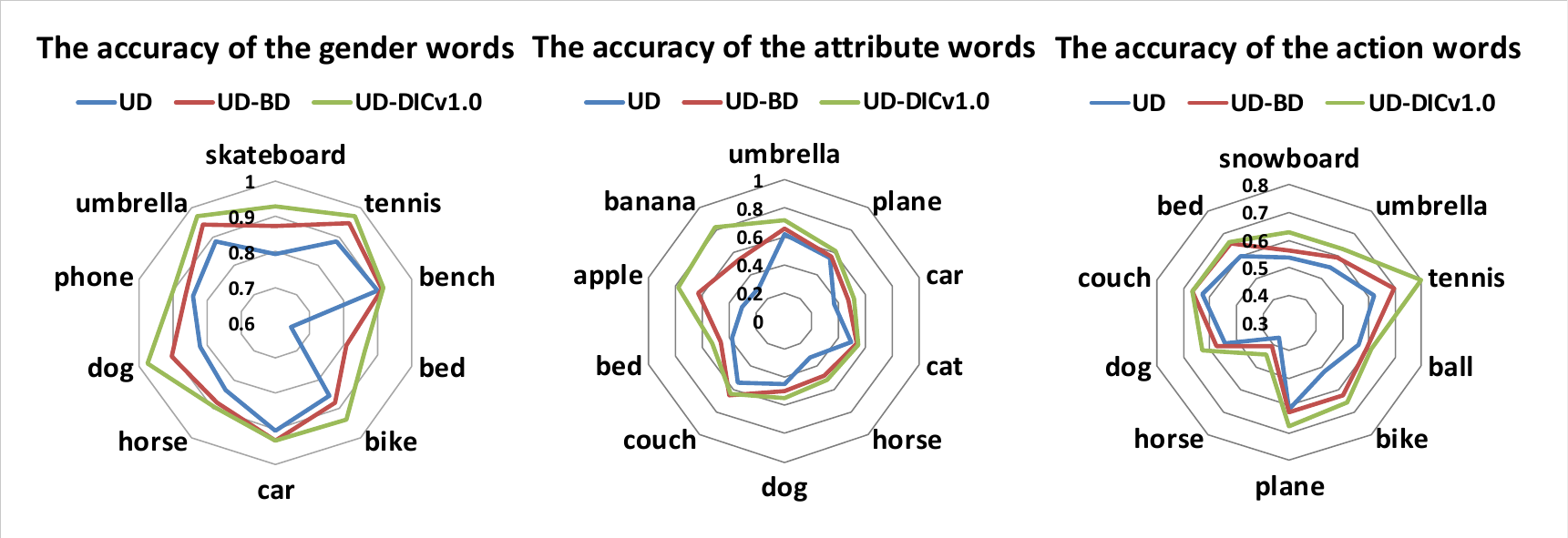}
  \caption{The accuracy of the gender, attribute, and action words when some specific visual concepts appear.
  }
\label{fig:fig_radar}
\vspace{-0.2in}
\end{figure}

\begin{figure*}[t]
\centering
\includegraphics[width=1\linewidth,trim = 5mm 5mm 5mm 5mm,clip]{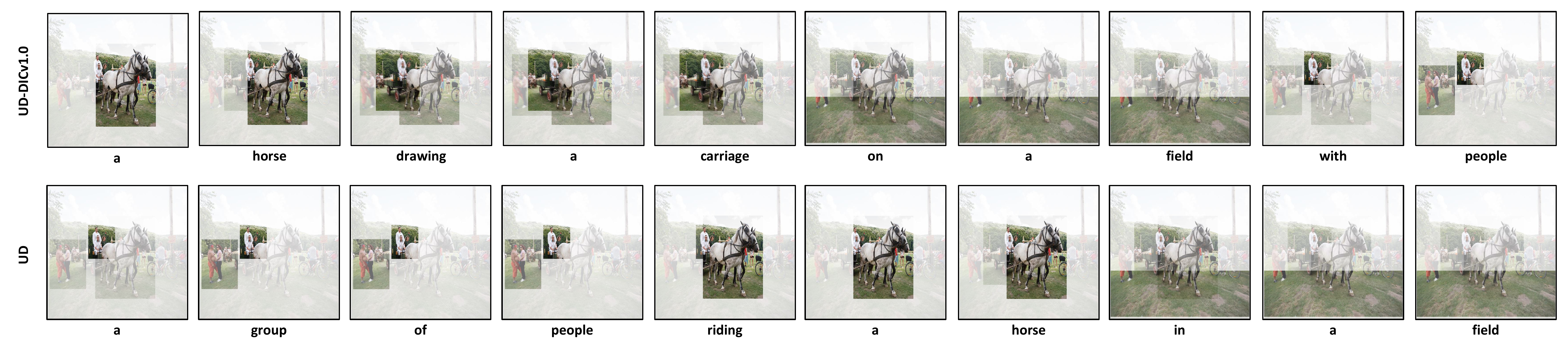}
  \caption{The attention comparisons between UD-DICv1.0 and UD. In the top row, UD-DICv1.0 can generate the right action “draw” by attending to the carriage and the horse, while in the bottom row, UD generates the wrong action ``ride'' by attending to the people and the horse.
  }
\label{fig:fig_att_response}
\vspace{-0.2in}
\end{figure*}

\begin{figure*}[t]
\centering
\includegraphics[width=1\linewidth,trim = 5mm 5mm 5mm 5mm,clip]{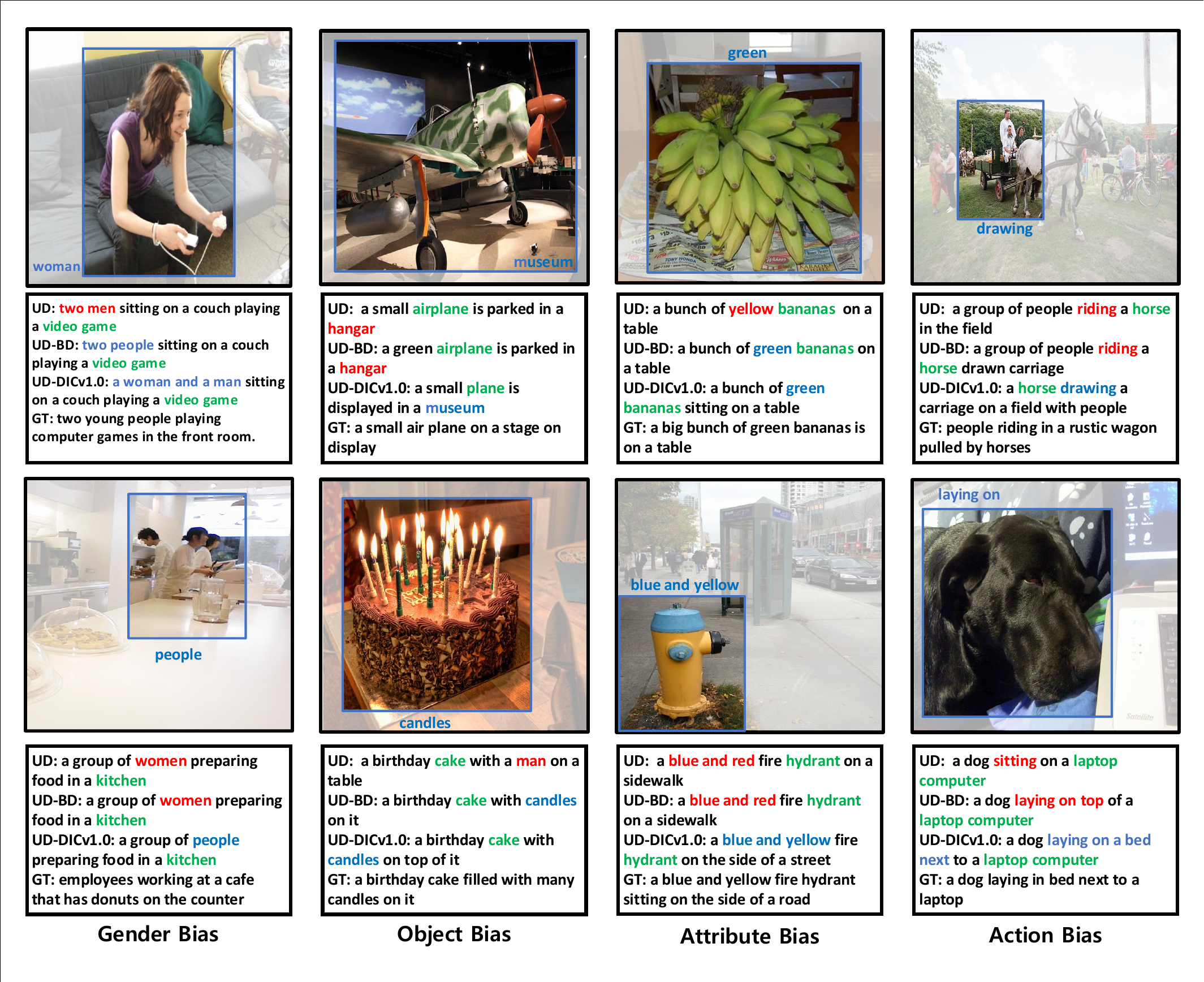}
  \caption{Eight examples about gender, object, attribute, and action biases in captions generated by different models. ``GT'' denotes the ground truth caption. The visual concepts which may cause biases are colored by green. The red and blue words represent the inconsistent and consistent words, respectively. For certain accurate words generated by UD-DICv1.0, we annotate them near the corresponding attended regions.} It can be found that our UD-DICv1.0 can generate the most consistent captions compared with UD and UD-BD. 
\label{fig:fig_example}
\vspace{-0.2in}
\end{figure*}

\noindent\textbf{Analyses of the Biases.}
To measure the bias degree of the generated captions, we not only applied CHs\&CHi~\cite{rohrbach2018object} to measure the object bias but also analyzed four more specific biases: gender bias, action bias, attribute bias, and quantifier bias. We evaluated these biases by calculating the accuracy of these words when a specific visual concept appears in the ground-truth captions. Take the gender bias as an example, we calculated whether the gender word (\eg, ``she'', ``he'', ``girl'', or ``boy'') is consistent between the ground-truth caption and the generated caption when a visual concept, \eg, skateboard, appears. We averaged the accuracy of the gender word over 80 visual concepts and report the mean accuracy in Table~\ref{table:tab_kap_baseline}. It is noteworthy that we considered the balance of the words that we separately calculated the accuracy of each word and averaged the results to obtain the mean accuracy.

From this table, we can observe that when better deconfounding techniques are used, CHs\&CHi decreases and the accuracy of four kinds of words increases, \eg, UD-BD performs better than UD and UD-FD outperforms UD-FD/Cor.
Compared with UD, UD-BD is better since it approximates the backdoor adjustment (Eq.~\eqref{equ:equ_do}) to alleviate the dataset bias. We split the confounder $\mathcal{D}$ into 80 visual concepts since they are the most frequently appeared objects which are more likely to cause biases. Therefore, UD-BD can reduce the biases induced by these concepts.
One interesting observation is that although the similarity performances (\eg, CIDEr-D) of UD-BD/CC is comparable to UD-FD, the bias metrics (\eg, CHs/CHi) of UD-BD/CC are still weaker than UD-FD. Such observation suggests that when the confounder is very complex, the front-door adjustment is still a better deconfounding strategy compared with using more additional knowledge in the backdoor adjustment. We can also find that UD-DICv1.0 achieves the lowest CHs\&CHi and the highest accuracy, indicating that UD-DICv1.0 generates the least biased captions. 

The radar charts in Fig.~\ref{fig:fig_radar} show the accuracy when some specific visual concepts appear. In the radar chart of gender words, we can find that the visual concept $ID_{\text{Bed}}$ induces strong gender biases where the accuracy of UD is smaller than 0.6, while our UD-DICv1.0 can alleviate such bias by improving the accuracy to almost 0.9. Also, we can find that some biases are hardly alleviated, \eg, the action bias induced by $ID_{\text{Horse}}$ where the accuracies of three models are lower than 0.4. There are at least two reasons for this. The first one is about language bias that many verbs have a similar meaning while we do not consider this situation when counting consistent words for the accuracy, \eg, ``horse-pull-carriage'' and ``horse-draw-carriage''. The second one is about the selection bias that humans may use different action words to describe an image, \eg, when one human is near a horse, we may annotate it as ``human-stand near-horse'', ``human-walk with-horse'', or even ``human-have-horse''. Such an example demonstrates the complexities of the bias constructions and the difficulties of the bias evaluations, which requires further investigations to disentangle different biases and develop more advanced metrics for evaluating them.

\noindent\textbf{Human evaluation.} Considering a better way of evaluating biases is human evaluation, we thus asked 20 humans to sort the captions generated by three models UD/UD-BD/UD-DICv1.0 according to the consistencies of these captions with the images. We chose 50 images to generate the captions and we specifically chose the images which are more likely to induce biases, \eg, the images containing the visual concepts like $ID_{\text{Skateboard}}$ and $ID_{\text{Horse}}$. When these judges considered that two captions are similar, they sorted them with the same rank. After sorting, we pairwise compared the results and show the results in TABLE.~\ref{table:tab_human_evaluation}. We can see that the judges consider our UD-DICv1.0's captions are more consistent than UD (70\% vs. 12 \%) and UD-BD (61\% vs. 14 \%), which suggests that it generates less biased captions. Fig.~\ref{fig:fig_example} visualizes some qualitative examples used in the human evaluation.  

\re{
\noindent\textbf{Effects on Attention Mechanism.}
Another interesting question is that whether the deconfounding technique will benefit the attention mechanism. To answer it, we carried an additional human evaluation where we asked 20 humans to compare the attention qualities of UD and UD-DICv1.0. Specifically, we used the same 50 images in the previous human evaluation. For each image-caption pair, we showed the attention regions of the words which are more easily affected by the dataset bias, \eg, the genders, actions, and nouns. We showed two attended regions with the top-2 attention weights and asked the subjects to judge whether the attended region is aligned with the generated word. We report the results in the bottom part of TABLE~\ref{table:tab_human_evaluation}, where we can find that UD-DICv1.0 achieves better word-region alignments than UD, \ie, 68\% vs. 17\%, indicating the better attention quality of UD-DICv1.0. Another implicit way to evaluate the attention mechanism is to calculate the accuracy of quantifiers since a system cannot correctly generate the quantifiers without attending to all the queried objects~\cite{zhang2018learning}. Thus, higher A@Quan of UD-DICv1.0 than UD (0.57 vs. 0.46) in Table~\ref{table:tab_kap_baseline} also suggests that our method has a better attention mechanism. Fig.~\ref{fig:fig_att_response} shows one attention comparison between UD-DICv1.0 and UD. We can find that UD-DICv1.0 generates better words by attending to more suitable regions, \eg, UD-DICv1.0 generates the right action ``draw'' by attending to the carriage and the horse while UD generates an incorrect action ``ride'' by attending to the people and the horse.
}

\subsubsection{Network Size}
\label{subsubsec:ablation_network_size}
Compared with the baseline UD, our UD-DICv1.0 has a larger network, which contains three additional parallel LSTM layers and three \textsc{expt} modules that introduce some additional dictionaries. To check the effects of these additional LSTM layers and dictionaries, we carried the following ablation studies. We set the number of LSTM layers to different values in UD and UD-DICV1.0 to measure their effects. Also, we changed the sizes of the dictionaries $\bm{X}$ and $\bm{Z}$ in UD-DICv1.0. The performances of these baselines are list in Table~\ref{table:tab_netsize_baseline} where the column ``LSTM'' means the numbers of LSTM layers before the embedding layer, the columns ``$\bm{X}$'' and ``$\bm{Z}$'' denote the numbers of the elements in them. For convenience, we report the CIDEr-D, CHs, and CHi scores to measure the similarities and bias degrees.

\noindent\textbf{Results and Analysis.}
\begin{table}[t]
\begin{center}
\caption{The performances of various ablation studies about the network size on MS COCO Karpathy split. For the convenience of the comparisons, we add an index before each model for better reference.}
\label{table:tab_netsize_baseline}
\scalebox{0.95}{
\begin{tabular}{l c c c | c c c}
		\hline
		   Models   & LSTM & $\bm{X}$ & $\bm{Z}$ &  C$\uparrow$ & CHs$\downarrow$ & CHi$\downarrow$\\ \hline
           (1) UD & 1 & - & - & $125.3$ & $13.7$ & $8.9$\\ 
           (2) UD & 4 & - & - & $125.8$ & $13.5$ & $8.8$\\ 
           (3) UD-DICv1.0 & 1 & 10,000 & 9,590 & $127.9$ & $10.6$ & $6.8$\\ 
           (4) UD-DICv1.0 & 4 & 10,000 & 9,590 & $128.8$ & $10.1$ & $6.5$\\
           (5) UD-DICv1.0 & 4 & 5,000 & 9,590 & $128.4$ & $10.3$ & $6.7$\\
           (6) UD-DICv1.0 & 4 & 15,000 & 9,590 & $\bm{129.0}$ & $\bm{10.0}$ & $\bm{6.4}$\\
           (7) UD-DICv1.0 & 4 & 10,000 & 5,000 & $128.3$ & $10.4$ & $6.8$\\
           (8) UD-DICv1.0 & 4 & 10,000 & 15,000 & $128.9$ & $10.1$ & $6.6$\\
           \hline
        
\end{tabular}
}
\end{center}
\vspace{-0.2in}
\end{table}
By comparing (1) with (2) in Table~\ref{table:tab_netsize_baseline}, we can find that although adding more LSTM layers in UD can improve the CIDEr-D score while the improvement is not as large as our UD-DICv1.0, whose CIDEr-D score is 128.8, and more importantly, simply adding LSTM layers will not alleviate the bias since CHs and CHi are not obviously decreased. Also, as shown in (3) and (4), if we only use one LSTM layer to generate the context vector, the performances will not be degraded significantly. These comparisons suggest that the improvements of our UD-DICv1.0, especially its ability to alleviate the biases, come more from the approximations of the deconfounding techniques than the additional LSTM layers. From (4) to (8) in Table~\ref{table:tab_netsize_baseline}, we change the size of $\bm{X}$ and $\bm{Z}$. By comparing their performances, we can find that the changes are marginal, suggesting that the size of the dictionaries does not largely affect the performances.

\subsection{Comparisons with State of the Art}
\begin{table}[t]
\begin{center}
\caption{The performances of the state of the art ICs on Karpathy split. The top and bottom parts report the performances trained by CIDEr-D computed from 5 captions and the whole training set, respectively. ``Group'' specifies each IC's category according to Fig.~\ref{fig:fig_retro}. The symbol ${^\dagger}$ means the re-implemented model.}
\label{table:tab_kap}
\scalebox{0.95}{
\begin{tabular}{l c c c c c c}
           \hline 
		   Models   & Group & B@4& M & R &  C & S\\ \hline
           Up-Down~\cite{anderson2018bottom}   & b & $36.3$ & $27.7$ & $56.9$ & $120.1$ & $21.4$ \\ 
           Up-Down${^\dagger}$~\cite{anderson2018bottom}   & b & $37.2$& $27.5$ & $57.3$ & $125.3$ & $21.1$ \\
           RFNet~\cite{jiang2018recurrent}   & c & $37.9$ & $28.3$ & $58.3$ & $125.7$ & $21.7$ \\ 
           CAVP~\cite{zha2019context}   & c & $38.6$ & $28.3$ & $58.5$ & $126.3$ & $21.6$ \\
           LBPF~\cite{qin2019look}   & c  & $38.3$ & $28.5$ & $58.4$ & $127.6$ & $22.0$ \\
           CNM~\cite{yang2019learning}  & d & $38.7$ & $28.4$ & $58.7$ & $127.4$ & $21.8$\\
           GCN-LSTM~\cite{yao2018exploring}   & e  & $38.2$ & $28.5$ & $58.3$ & $127.6$ & $22.0$ \\
           SGAE~\cite{yang2019auto}  & e & $38.4$ & $28.4$ & $58.6$ & $127.8$ & $\bm{22.1}$ \\ 
           UD-DICv1.0 & f & $\bm{39.0}$ & $\bm{28.8}$ & $\bm{58.8}$ & $\bm{128.8}$ & $22.0$ \\ 
           \hline 
           Up-Down${^\dagger}$~\cite{anderson2018bottom}   & b & $37.7$& $28.2$ & $58.1$ & $126.4$ & $21.8$\\
           UD-HIP~\cite{yao2019hierarchy}  & b & $38.2$ & $28.4$ & $58.3$ & $127.2$ & $21.9$ \\
           VLP~\cite{zhou2019unified}  & b & $\bm{39.5}$ & $-$ & $-$ & $129.3$ & $\bm{23.2}$ \\ 
           AoANet~\cite{huang2019attention}  & c & $38.9$ & $29.2$ & $58.8$ & $129.8$ & $22.4$\\ 
           AoANet${^\dagger}$~\cite{huang2019attention}  & c & $38.9$ & $28.9$ & $58.4$ & $128.7$ & $22.4$ \\
           UD-DICv1.0  & f & $38.3$ & $28.5$ & $58.5$ & $129.5$ & $22.0$ \\
           AoA-DICv1.0 & f & $\bm{39.5}$ & $\bm{29.6}$ & $\bm{59.0}$ & $\bm{131.1}$ & $22.6$ \\
           \hline 
\end{tabular}
}
\end{center}
\vspace{-0.2in}
\end{table}
\begin{table}[t]
\begin{center}
\caption{\re{The comparisons between GroupCap and our models on Karpathy split.}}
\label{table:tab_group}
\scalebox{0.95}{
\begin{tabular}{l c c c c c}
		\hline
		   \re{Models}  & \re{B@1}& \re{B@2} & \re{B@3} &  \re{B@4} & \re{M}\\ \hline
		   \re{GroupCap~\cite{chen2018groupcap}} & \re{74.4} & \re{58.1} & \re{44.3} & \re{33.8} & \re{26.2} \\
		   \re{UD-BD} & \re{75.3} & \re{59.2} & \re{45.6} & \re{35.1} & \re{27.0} \\
		   \re{UD-DICv1.0} & $\re{\bm{77.2}}$ & \re{$\bm{61.5}$} & $\re{\bm{47.8}}$ & $\re{\bm{37.2}}$ & $\re{\bm{28.1}}$ \\ 
           \hline 
\end{tabular}
}
\end{center}
\vspace{-0.2in}
\end{table}

\noindent\textbf{Comparing Methods.} We start our comparisons from \textbf{Up-Down}~\cite{anderson2018bottom}, whose visual features are most frequently used by the subsequent captioners and so do we. We follow the causal retrospect in Fig.~\ref{fig:fig_retro} to group the compared state of the art captioners into four groups: \textbf{large-scale pre-training:} \textbf{Up-Down}~\cite{anderson2018bottom}, \textbf{UD-HIP}~\cite{yao2019hierarchy}, and \textbf{VLP}~\cite{zhou2019unified}; \textbf{attention mechanisms:} \textbf{CAVP}~\cite{liu2018context}, \textbf{RFNet}~\cite{jiang2018recurrent}, \textbf{LBPF}~\cite{qin2019look}, and \textbf{AoANet}~\cite{huang2019attention}; \textbf{sentence patterns:} \textbf{CNM}~\cite{yang2019learning}; and \textbf{structured attention:} \textbf{GCN-LSTM}~\cite{yao2018exploring} and \textbf{SGAE}~\cite{yang2019auto}. 
Note that two different CIDEr-D were used as the self-critical rewards in the previous captioners. The first one computes Inverse Document Frequency (IDF) from each image's five ground-truth captions and the second one computes IDF from the captions of the whole training set. For fair comparisons, we also used two different CIDEr-D based self-critical rewards to train our DICs and compare them with the captioners trained by the same reward. The results are reported in Table~\ref{table:tab_kap}, where the top and bottom parts list the results of the captioners trained by the first and second CIDEr-D based rewards, respectively.
\re{For \textbf{GroupCap}~\cite{chen2018groupcap}, it also carried an experiment on the standard MS COCO one-line captioning, and thus we can compare our model with it. For fair comparisons, we follow GroupCap to use ResNet~\cite{he2016deep} trained by image classification as the visual encoder, LSTM as the language decoder (UD-BD and UD-DICv1.0), and cross-entropy loss (Eq.~\eqref{equ:equ_celoss}) as the training target.
The results are reported in TABLE~\ref{table:tab_group}.
}

\noindent\textbf{Results and Analysis.}
From TABLE~\ref{table:tab_kap}, we can find that our single-model UD-DICv1.0 and AoA-DICv1.0 achieve the best CIDEr-D scores: 128.8 and 131.1 when different training rewards are used. Compared with UD-HIP which pre-trained their captioner by object detection and segmentation, our UD-DICv1.0, which was only pre-trained by object detection, has better performances. Interestingly, compared with VLP which exploited 30 times more samples (3 million) than ours (0.1 million) to pre-train their model, our UD-DICv1.0 and AoA-DICv1.0 still achieve competitive results. Both comparisons suggest that our DIC framework is more cost-effective and efficient than these large-scale pre-training captioners in boosting performances. Our DIC also outperforms the captioners with advanced attention mechanisms, \eg, UD-DICv1.0 is better than RFNet, CAVP, and LBPF; and AoA-DICv1.0 is better than AoANet though we did not use more advanced training strategy as AoANet~\cite{huang2019attention}. Finally, comparing our UD-DICv1.0 with the approximations of the front-door frameworks: sentence patterns and structured attention, we find that our UD-DICv1.0 is still the best, although we did not use multi-step reasoning as CNM and did not deploy complex graph convolution as GCN-LSTM and SGAE. \re{From TABLE~\ref{table:tab_group}, we can also see that UD-BD and UD-DICv1.0 achieve higher BLEUs and METEOR than GroupCap, which suggests that the backdoor and front-door adjustments are also good strategies of using the other image samples for captioning, even without explicitly exploiting the relevance and diversity knowledge.}

We also compare our single-model UD-DICv1.0 and AoA-DICv1.0 with the other captioners on MS COCO online test set, where the results are reported in Table~\ref{table:tab_online}. It can be observed that our UD-DICv1.0 and AoA-DICv1.0 achieve the highest CIDEr-D c5 and c40 scores. All of these comparisons confirm the superiority of the proposed DIC framework over the captioners with weaker deconfounding approximations.

\begin{table}[t]
\begin{center}
\caption{The performances of single methods on the online MS COCO test server. The top and bottom parts report the score trained by CIDEr-D computed from 5 captions and the whole training set, respectively.}
\label{table:tab_online}
\scalebox{0.80}{
\begin{tabular}{l c c c c c c c c c c c}
		\hline
		 Model  & \multicolumn{2}{c}{B@4} &\multicolumn{2}{c}{M} &\multicolumn{2}{c}{R-L} & \multicolumn{2}{c}{C-D}\\ \hline 
		 Metric  &   c5 & c40 & c5 & c40 & c5 & c40 & c5 & c40 \\ \hline
           Up-Down~\cite{anderson2018bottom}    & $36.9$ & $68.5$ & $27.6$ & $36.7$ & $57.1$ & $72.4$ & $117.9$& $120.5$  \\ 
           CAVP~\cite{liu2018context}           & $37.9$ & $69.0$ & $28.1$ & $37.0$ & $58.2$ & $73.1$ & $121.6$& $123.8$    \\
           RFNet~\cite{jiang2018recurrent}      & $\bm{38.0}$ & $\bm{69.2}$ & $28.2$ & $37.2$ & $58.2$ & $73.1$ & $122.9$& $125.1$    \\
           SGAE~\cite{yang2019auto}    &$37.8$ & $68.7$& $28.1$ & $37.0$ & $58.2$ & $73.1$ & $122.7$ & $125.5$\\ 
           CNM~\cite{yang2019learning}    &$37.9$ & $68.4$& $28.1$ & $36.9$ & $\bm{58.3}$ & $72.9$ & $123.0$ & $125.3$\\ 
           UD-DICv1.0 & $37.9$ & $\bm{69.2}$ & $\bm{28.7}$ & $\bm{37.7}$ & $\bm{58.3}$ & $\bm{73.3}$ & $\bm{124.1}$ & $\bm{126.7}$ \\ \hline
           AoANet~\cite{huang2019attention} & $37.3$ & $68.1$ & $28.3$ & $37.2$ & $57.9$ & $72.8$ & $124.0$ & $126.2$ \\ 
           AoA-DICv1.0 & $\bm{38.8}$ & $\bm{70.5}$ & $\bm{28.8}$ & $\bm{38.2}$ & $\bm{58.6}$ & $\bm{73.9}$ & $\bm{126.2}$ & $\bm{128.4}$ \\ 
           \hline
\end{tabular}
}
\end{center}
\vspace{-0.2in}
\end{table}

\section{Conclusions}
\label{sec:epilogue}
We used the causal perspective to offer an in-depth analysis about why modern captioners easily learn the spurious correlation. We discovered that some frequently appeared concepts act as a typical confounder that if we neglect such confounder and train the captioners by the observational correlation $P(\mathcal{L}|\mathcal{X})$, the learned captioners collapse into the dataset bias. Then we introduced two techniques: the backdoor and front-door adjustments which can alleviate such problem by calculating the interventional probability $P(\mathcal{L}|do(\mathcal{X}))$ as the training target. We also retrospected the major progress in IC and found that many advanced captioners are the approximations of the backdoor and front-door adjustments. After that, we derived an effective method called DICv1.0 based on the causal retrospect and validated its effectiveness by incorporating it into two prevailing models: Up-Down and AoANet. 

From the causal retrospect in Fig.~\ref{fig:fig_retro}, we see a promising future: DICv1.0 is just a start! For example, it is derived by assuming that $\mathcal{S}$ can be split into single keywords for deconfounding, while this may also not be the real situation. As the red parts we draw in
Fig.~\ref{fig:fig_retro}(g), if we cannot split $\mathcal{S}$, we can re-assign $\mathcal{S}$ to $\mathcal{D}$ and $\mathcal{Z}$ to $\mathcal{X}$, then we jump back into “where we start” and it is possible to recursively deploy all the previous techniques, including this paper, which will be the “previous IC” in the recursive future.

\section*{Acknowledgments}
This work is supported by Singapore MOE AcRF Tier 2.

\bibliographystyle{IEEEtran}
\bibliography{egbib}

% Generated by IEEEtran.bst, version: 1.14 (2015/08/26)
\begin{thebibliography}{10}
\providecommand{\url}[1]{#1}
\csname url@samestyle\endcsname
\providecommand{\newblock}{\relax}
\providecommand{\bibinfo}[2]{#2}
\providecommand{\BIBentrySTDinterwordspacing}{\spaceskip=0pt\relax}
\providecommand{\BIBentryALTinterwordstretchfactor}{4}
\providecommand{\BIBentryALTinterwordspacing}{\spaceskip=\fontdimen2\font plus
\BIBentryALTinterwordstretchfactor\fontdimen3\font minus
  \fontdimen4\font\relax}
\providecommand{\BIBforeignlanguage}[2]{{%
\expandafter\ifx\csname l@#1\endcsname\relax
\typeout{** WARNING: IEEEtran.bst: No hyphenation pattern has been}%
\typeout{** loaded for the language `#1'. Using the pattern for}%
\typeout{** the default language instead.}%
\else
\language=\csname l@#1\endcsname
\fi
#2}}
\providecommand{\BIBdecl}{\relax}
\BIBdecl

\bibitem{hendricks2018women}
L.~A. Hendricks, K.~Burns, K.~Saenko, T.~Darrell, and A.~Rohrbach, ``Women also
  snowboard: Overcoming bias in captioning models,'' in \emph{European
  Conference on Computer Vision}.\hskip 1em plus 0.5em minus 0.4em\relax
  Springer, 2018, pp. 793--811.

\bibitem{rohrbach2018object}
A.~Rohrbach, L.~A. Hendricks, K.~Burns, T.~Darrell, and K.~Saenko, ``Object
  hallucination in image captioning,'' \emph{Proceedings of the 2018 Conference
  on Empirical Methods in Natural Language Processing}, 2018.

\bibitem{goyal2017making}
Y.~Goyal, T.~Khot, D.~Summers-Stay, D.~Batra, and D.~Parikh, ``Making the v in
  vqa matter: Elevating the role of image understanding in visual question
  answering,'' in \emph{Proceedings of the IEEE Conference on Computer Vision
  and Pattern Recognition}, 2017, pp. 6904--6913.

\bibitem{reed2001pareto}
W.~J. Reed, ``The pareto, zipf and other power laws,'' \emph{Economics
  letters}, vol.~74, no.~1, pp. 15--19, 2001.

\bibitem{liu2019large}
Z.~Liu, Z.~Miao, X.~Zhan, J.~Wang, B.~Gong, and S.~X. Yu, ``Large-scale
  long-tailed recognition in an open world,'' in \emph{Proceedings of the IEEE
  Conference on Computer Vision and Pattern Recognition}, 2019, pp. 2537--2546.

\bibitem{gordon2013reporting}
J.~Gordon and B.~Van~Durme, ``Reporting bias and knowledge extraction,'' 2013.

\bibitem{misra2016seeing}
I.~Misra, C.~Lawrence~Zitnick, M.~Mitchell, and R.~Girshick, ``Seeing through
  the human reporting bias: Visual classifiers from noisy human-centric
  labels,'' in \emph{Proceedings of the IEEE Conference on Computer Vision and
  Pattern Recognition}, 2016, pp. 2930--2939.

\bibitem{bolukbasi2016man}
T.~Bolukbasi, K.-W. Chang, J.~Y. Zou, V.~Saligrama, and A.~T. Kalai, ``Man is
  to computer programmer as woman is to homemaker? debiasing word embeddings,''
  in \emph{Advances in neural information processing systems}, 2016, pp.
  4349--4357.

\bibitem{caliskan2017semantics}
A.~Caliskan, J.~J. Bryson, and A.~Narayanan, ``Semantics derived automatically
  from language corpora contain human-like biases,'' \emph{Science}, vol. 356,
  no. 6334, pp. 183--186, 2017.

\bibitem{johnson2017clevr}
J.~Johnson, B.~Hariharan, L.~van~der Maaten, L.~Fei-Fei, C.~L. Zitnick, and
  R.~Girshick, ``Clevr: A diagnostic dataset for compositional language and
  elementary visual reasoning,'' in \emph{Computer Vision and Pattern
  Recognition (CVPR), 2017 IEEE Conference on}.\hskip 1em plus 0.5em minus
  0.4em\relax IEEE, 2017, pp. 1988--1997.

\bibitem{agrawal2018don}
A.~Agrawal, D.~Batra, D.~Parikh, and A.~Kembhavi, ``Don't just assume; look and
  answer: Overcoming priors for visual question answering,'' in
  \emph{Proceedings of the IEEE Conference on Computer Vision and Pattern
  Recognition}, 2018, pp. 4971--4980.

\bibitem{sharma2018conceptual}
P.~Sharma, N.~Ding, S.~Goodman, and R.~Soricut, ``Conceptual captions: A
  cleaned, hypernymed, image alt-text dataset for automatic image captioning,''
  in \emph{Proceedings of the 56th Annual Meeting of the Association for
  Computational Linguistics (Volume 1: Long Papers)}, 2018, pp. 2556--2565.

\bibitem{cadene2019rubi}
R.~Cadene, C.~Dancette, M.~Cord, D.~Parikh \emph{et~al.}, ``Rubi: Reducing
  unimodal biases for visual question answering,'' \emph{Advances in neural
  information processing systems}, vol.~32, pp. 841--852, 2019.

\bibitem{pearl2000causality}
J.~Pearl, \emph{Causality: models, reasoning and inference}.\hskip 1em plus
  0.5em minus 0.4em\relax Springer, 2000, vol.~29.

\bibitem{pearl2016causal}
J.~Pearl, M.~Glymour, and N.~P. Jewell, \emph{Causal inference in statistics: A
  primer}.\hskip 1em plus 0.5em minus 0.4em\relax John Wiley \& Sons, 2016.

\bibitem{PearlMackenzie18}
J.~Pearl and D.~Mackenzie, \emph{\BIBforeignlanguage{american}{The Book of
  Why}}.\hskip 1em plus 0.5em minus 0.4em\relax New York: Basic Books, 2018.

\bibitem{yao2010i2t}
B.~Z. Yao, X.~Yang, L.~Lin, M.~W. Lee, and S.-C. Zhu, ``I2t: Image parsing to
  text description,'' \emph{Proceedings of the IEEE}, vol.~98, no.~8, pp.
  1485--1508, 2010.

\bibitem{antol2015vqa}
S.~Antol, A.~Agrawal, J.~Lu, M.~Mitchell, D.~Batra, C.~Lawrence~Zitnick, and
  D.~Parikh, ``Vqa: Visual question answering,'' in \emph{Proceedings of the
  IEEE international conference on computer vision}, 2015, pp. 2425--2433.

\bibitem{kazemzadeh2014referitgame}
S.~Kazemzadeh, V.~Ordonez, M.~Matten, and T.~Berg, ``Referitgame: Referring to
  objects in photographs of natural scenes,'' in \emph{Proceedings of the 2014
  conference on empirical methods in natural language processing (EMNLP)},
  2014, pp. 787--798.

\bibitem{mao2016generation}
J.~Mao, J.~Huang, A.~Toshev, O.~Camburu, A.~L. Yuille, and K.~Murphy,
  ``Generation and comprehension of unambiguous object descriptions,'' in
  \emph{Proceedings of the IEEE conference on computer vision and pattern
  recognition}, 2016, pp. 11--20.

\bibitem{vinyals2015show}
O.~Vinyals, A.~Toshev, S.~Bengio, and D.~Erhan, ``Show and tell: A neural image
  caption generator,'' in \emph{CVPR}, 2015.

\bibitem{das2017visual}
A.~Das, S.~Kottur, K.~Gupta, A.~Singh, D.~Yadav, J.~M. Moura, D.~Parikh, and
  D.~Batra, ``Visual dialog,'' in \emph{Proceedings of the IEEE Conference on
  Computer Vision and Pattern Recognition}, 2017, pp. 326--335.

\bibitem{hudson2019gqa}
D.~A. Hudson and C.~D. Manning, ``Gqa: A new dataset for real-world visual
  reasoning and compositional question answering,'' in \emph{Proceedings of the
  IEEE Conference on Computer Vision and Pattern Recognition}, 2019, pp.
  6700--6709.

\bibitem{anderson2018bottom}
P.~Anderson, X.~He, C.~Buehler, D.~Teney, M.~Johnson, S.~Gould, and L.~Zhang,
  ``Bottom-up and top-down attention for image captioning and visual question
  answering,'' in \emph{CVPR}, 2018.

\bibitem{huang2019attention}
L.~Huang, W.~Wang, J.~Chen, and X.-Y. Wei, ``Attention on attention for image
  captioning,'' in \emph{IEEE/CVF International Conference on Computer Vision},
  2019.

\bibitem{geirhos2020shortcut}
R.~Geirhos, J.-H. Jacobsen, C.~Michaelis, R.~Zemel, W.~Brendel, M.~Bethge, and
  F.~A. Wichmann, ``Shortcut learning in deep neural networks,'' \emph{Nature
  Machine Intelligence}, vol.~2, no.~11, pp. 665--673, 2020.

\bibitem{ren2015faster}
S.~Ren, K.~He, R.~Girshick, and J.~Sun, ``Faster r-cnn: Towards real-time
  object detection with region proposal networks,'' in \emph{Advances in neural
  information processing systems}, 2015, pp. 91--99.

\bibitem{lin2014microsoft}
T.-Y. Lin, M.~Maire, S.~Belongie, J.~Hays, P.~Perona, D.~Ramanan,
  P.~Doll{\'a}r, and C.~L. Zitnick, ``Microsoft coco: Common objects in
  context,'' in \emph{European conference on computer vision}.\hskip 1em plus
  0.5em minus 0.4em\relax Springer, 2014, pp. 740--755.

\bibitem{chen2015microsoft}
X.~Chen, H.~Fang, T.-Y. Lin, R.~Vedantam, S.~Gupta, P.~Doll{\'a}r, and C.~L.
  Zitnick, ``Microsoft coco captions: Data collection and evaluation server,''
  \emph{arXiv preprint arXiv:1504.00325}, 2015.

\bibitem{xu2015show}
K.~Xu, J.~Ba, R.~Kiros, K.~Cho, A.~Courville, R.~Salakhudinov, R.~Zemel, and
  Y.~Bengio, ``Show, attend and tell: Neural image caption generation with
  visual attention,'' in \emph{International conference on machine learning},
  2015, pp. 2048--2057.

\bibitem{kulkarni2013babytalk}
G.~Kulkarni, V.~Premraj, V.~Ordonez, S.~Dhar, S.~Li, Y.~Choi, A.~C. Berg, and
  T.~L. Berg, ``Babytalk: Understanding and generating simple image
  descriptions,'' in \emph{CVPR}, 2011.

\bibitem{krishna2017visual}
R.~Krishna, Y.~Zhu, O.~Groth, J.~Johnson, K.~Hata, J.~Kravitz, S.~Chen,
  Y.~Kalantidis, L.-J. Li, D.~A. Shamma \emph{et~al.}, ``Visual genome:
  Connecting language and vision using crowdsourced dense image annotations,''
  \emph{International Journal of Computer Vision}, vol. 123, no.~1, pp. 32--73,
  2017.

\bibitem{yao2019hierarchy}
T.~Yao, Y.~Pan, Y.~Li, and T.~Mei, ``Hierarchy parsing for image captioning,''
  in \emph{Proceedings of the IEEE International Conference on Computer
  Vision}, 2019, pp. 2621--2629.

\bibitem{lu2019vilbert}
J.~Lu, D.~Batra, D.~Parikh, and S.~Lee, ``Vilbert: Pretraining task-agnostic
  visiolinguistic representations for vision-and-language tasks,''
  \emph{Advances in Neural Information Processing Systems 32: Annual Conference
  on Neural Information Processing Systems 2019}, 2019.

\bibitem{zhou2019unified}
L.~Zhou, H.~Palangi, L.~Zhang, H.~Hu, J.~J. Corso, and J.~Gao, ``Unified
  vision-language pre-training for image captioning and vqa,'' \emph{The
  Thirty-Fourth AAAI Conference on Artificial Intelligence}, 2020.

\bibitem{you2016image}
Q.~You, H.~Jin, Z.~Wang, C.~Fang, and J.~Luo, ``Image captioning with semantic
  attention,'' in \emph{Proceedings of the IEEE conference on computer vision
  and pattern recognition}, 2016, pp. 4651--4659.

\bibitem{lu2017knowing}
J.~Lu, C.~Xiong, D.~Parikh, and R.~Socher, ``Knowing when to look: Adaptive
  attention via a visual sentinel for image captioning,'' in \emph{Proceedings
  of the IEEE Conference on Computer Vision and Pattern Recognition (CVPR)},
  vol.~6, 2017, p.~2.

\bibitem{yao2017boosting}
T.~Yao, Y.~Pan, Y.~Li, Z.~Qiu, and T.~Mei, ``Boosting image captioning with
  attributes,'' in \emph{IEEE International Conference on Computer Vision,
  ICCV}, 2017, pp. 22--29.

\bibitem{zha2019context}
Z.-J. Zha, D.~Liu, H.~Zhang, Y.~Zhang, and F.~Wu, ``Context-aware visual policy
  network for fine-grained image captioning,'' \emph{IEEE transactions on
  pattern analysis and machine intelligence}, 2019.

\bibitem{cornia2020meshed}
M.~Cornia, M.~Stefanini, L.~Baraldi, and R.~Cucchiara, ``Meshed-memory
  transformer for image captioning,'' in \emph{Proceedings of the IEEE/CVF
  Conference on Computer Vision and Pattern Recognition}, 2020, pp.
  10\,578--10\,587.

\bibitem{zhou2020more}
Y.~Zhou, M.~Wang, D.~Liu, Z.~Hu, and H.~Zhang, ``More grounded image captioning
  by distilling image-text matching model,'' in \emph{Proceedings of the
  IEEE/CVF Conference on Computer Vision and Pattern Recognition}, 2020, pp.
  4777--4786.

\bibitem{lu2018neural}
J.~Lu, J.~Yang, D.~Batra, and D.~Parikh, ``Neural baby talk,'' in
  \emph{Proceedings of the IEEE Conference on Computer Vision and Pattern
  Recognition}, 2018.

\bibitem{yang2019learning}
X.~Yang, H.~Zhang, and J.~Cai, ``Learning to collocate neural modules for image
  captioning,'' \emph{IEEE/CVF International Conference on Computer Vision},
  2019.

\bibitem{yao2018exploring}
T.~Yao, Y.~Pan, Y.~Li, and T.~Mei, ``Exploring visual relationship for image
  captioning,'' in \emph{Computer Vision--ECCV 2018}.\hskip 1em plus 0.5em
  minus 0.4em\relax Springer, 2018, pp. 711--727.

\bibitem{yang2019auto}
X.~Yang, K.~Tang, H.~Zhang, and J.~Cai, ``Auto-encoding scene graphs for image
  captioning,'' in \emph{Proceedings of the IEEE Conference on Computer Vision
  and Pattern Recognition}, 2019, pp. 10\,685--10\,694.

\bibitem{chen2020say}
S.~Chen, Q.~Jin, P.~Wang, and Q.~Wu, ``Say as you wish: Fine-grained control of
  image caption generation with abstract scene graphs,'' in \emph{Proceedings
  of the IEEE/CVF Conference on Computer Vision and Pattern Recognition}, 2020,
  pp. 9962--9971.

\bibitem{yang2020auto}
X.~Yang, H.~Zhang, and J.~Cai, ``Auto-encoding and distilling scene graphs for
  image captioning,'' \emph{IEEE Transactions on Pattern Analysis and Machine
  Intelligence}, 2020.

\bibitem{micheli2009neural}
A.~Micheli, ``Neural network for graphs: A contextual constructive approach,''
  \emph{IEEE Transactions on Neural Networks}, vol.~20, no.~3, pp. 498--511,
  2009.

\bibitem{atwood2016diffusion}
J.~Atwood and D.~Towsley, ``Diffusion-convolutional neural networks,'' in
  \emph{Advances in neural information processing systems}, 2016, pp.
  1993--2001.

\bibitem{niepert2016learning}
M.~Niepert, M.~Ahmed, and K.~Kutzkov, ``Learning convolutional neural networks
  for graphs,'' in \emph{International conference on machine learning}, 2016,
  pp. 2014--2023.

\bibitem{teney2017graph}
D.~Teney, L.~Liu, and A.~van~den Hengel, ``Graph-structured representations for
  visual question answering,'' in \emph{2017 IEEE Conference on Computer Vision
  and Pattern Recognition (CVPR)}.\hskip 1em plus 0.5em minus 0.4em\relax IEEE,
  2017, pp. 3233--3241.

\bibitem{li2019relation}
L.~Li, Z.~Gan, Y.~Cheng, and J.~Liu, ``Relation-aware graph attention network
  for visual question answering,'' in \emph{Proceedings of the IEEE
  International Conference on Computer Vision}, 2019, pp. 10\,313--10\,322.

\bibitem{jing2020visual}
C.~Jing, Y.~Wu, M.~Pei, Y.~Hu, Y.~Jia, and Q.~Wu, ``Visual-semantic graph
  matching for visual grounding,'' in \emph{Proceedings of the 28th ACM
  International Conference on Multimedia}, 2020, pp. 4041--4050.

\bibitem{liu2019learning}
D.~Liu, H.~Zhang, F.~Wu, and Z.-J. Zha, ``Learning to assemble neural module
  tree networks for visual grounding,'' in \emph{Proceedings of the IEEE
  International Conference on Computer Vision}, 2019, pp. 4673--4682.

\bibitem{yu2020learning}
J.~Yu, X.~Jiang, Z.~Qin, W.~Zhang, Y.~Hu, and Q.~Wu, ``Learning dual encoding
  model for adaptive visual understanding in visual dialogue,'' \emph{IEEE
  Transactions on Image Processing}, vol.~30, pp. 220--233, 2020.

\bibitem{lu2016visual}
C.~Lu, R.~Krishna, M.~Bernstein, and L.~Fei-Fei, ``Visual relationship
  detection with language priors,'' in \emph{European conference on computer
  vision}.\hskip 1em plus 0.5em minus 0.4em\relax Springer, 2016, pp. 852--869.

\bibitem{zellers2018neural}
R.~Zellers, M.~Yatskar, S.~Thomson, and Y.~Choi, ``Neural motifs: Scene graph
  parsing with global context,'' in \emph{Proceedings of the IEEE Conference on
  Computer Vision and Pattern Recognition}, 2018, pp. 5831--5840.

\bibitem{rubin2005causal}
D.~B. Rubin, ``Causal inference using potential outcomes: Design, modeling,
  decisions,'' \emph{Journal of the American Statistical Association}, vol.
  100, no. 469, pp. 322--331, 2005.

\bibitem{suter2019robustly}
R.~Suter, D.~Miladinovic, B.~Sch{\"o}lkopf, and S.~Bauer, ``Robustly
  disentangled causal mechanisms: Validating deep representations for
  interventional robustness,'' in \emph{International Conference on Machine
  Learning}.\hskip 1em plus 0.5em minus 0.4em\relax PMLR, 2019, pp. 6056--6065.

\bibitem{abbasnejad2020counterfactual}
E.~Abbasnejad, D.~Teney, A.~Parvaneh, J.~Shi, and A.~v.~d. Hengel,
  ``Counterfactual vision and language learning,'' in \emph{Proceedings of the
  IEEE/CVF Conference on Computer Vision and Pattern Recognition}, 2020, pp.
  10\,044--10\,054.

\bibitem{kocaoglu2017causalgan}
M.~Kocaoglu, C.~Snyder, A.~G. Dimakis, and S.~Vishwanath, ``Causalgan: Learning
  causal implicit generative models with adversarial training,'' \emph{6th
  International Conference on Learning Representation}, 2018.

\bibitem{veitch2020adapting}
V.~Veitch, D.~Sridhar, and D.~Blei, ``Adapting text embeddings for causal
  inference,'' in \emph{Conference on Uncertainty in Artificial
  Intelligence}.\hskip 1em plus 0.5em minus 0.4em\relax PMLR, 2020, pp.
  919--928.

\bibitem{tang2020unbiased}
K.~Tang, Y.~Niu, J.~Huang, J.~Shi, and H.~Zhang, ``Unbiased scene graph
  generation from biased training,'' in \emph{Proceedings of the IEEE/CVF
  Conference on Computer Vision and Pattern Recognition}, 2020, pp. 3716--3725.

\bibitem{wang2020visual}
T.~Wang, J.~Huang, H.~Zhang, and Q.~Sun, ``Visual commonsense r-cnn,'' in
  \emph{Proceedings of the IEEE/CVF Conference on Computer Vision and Pattern
  Recognition}, 2020, pp. 10\,760--10\,770.

\bibitem{qi2020two}
J.~Qi, Y.~Niu, J.~Huang, and H.~Zhang, ``Two causal principles for improving
  visual dialog,'' in \emph{Proceedings of the IEEE/CVF Conference on Computer
  Vision and Pattern Recognition}, 2020, pp. 10\,860--10\,869.

\bibitem{niu2021counterfactual}
Y.~Niu, K.~Tang, H.~Zhang, Z.~Lu, X.-S. Hua, and J.-R. Wen, ``Counterfactual
  vqa: A cause-effect look at language bias,'' in \emph{Proceedings of the
  IEEE/CVF Conference on Computer Vision and Pattern Recognition}, 2021, pp.
  12\,700--12\,710.

\bibitem{yang2021causal}
X.~Yang, H.~Zhang, G.~Qi, and J.~Cai, ``Causal attention for vision-language
  tasks,'' in \emph{Proceedings of the IEEE/CVF Conference on Computer Vision
  and Pattern Recognition}, 2021, pp. 9847--9857.

\bibitem{chen2018groupcap}
F.~Chen, R.~Ji, X.~Sun, Y.~Wu, and J.~Su, ``Groupcap: Group-based image
  captioning with structured relevance and diversity constraints,'' in
  \emph{Proceedings of the IEEE conference on computer vision and pattern
  recognition}, 2018, pp. 1345--1353.

\bibitem{li2020context}
Z.~Li, Q.~Tran, L.~Mai, Z.~Lin, and A.~L. Yuille, ``Context-aware group
  captioning via self-attention and contrastive features,'' in
  \emph{Proceedings of the IEEE/CVF Conference on Computer Vision and Pattern
  Recognition}, 2020, pp. 3440--3450.

\bibitem{liu2004conceptnet}
H.~Liu and P.~Singh, ``Conceptnet—a practical commonsense reasoning
  tool-kit,'' \emph{BT technology journal}, vol.~22, no.~4, pp. 211--226, 2004.

\bibitem{srivastava2014dropout}
N.~Srivastava, G.~Hinton, A.~Krizhevsky, I.~Sutskever, and R.~Salakhutdinov,
  ``Dropout: a simple way to prevent neural networks from overfitting,''
  \emph{The journal of machine learning research}, vol.~15, no.~1, pp.
  1929--1958, 2014.

\bibitem{baldi2014dropout}
P.~Baldi and P.~Sadowski, ``The dropout learning algorithm,'' \emph{Artificial
  intelligence}, vol. 210, pp. 78--122, 2014.

\bibitem{mairal2009online}
J.~Mairal, F.~Bach, J.~Ponce, and G.~Sapiro, ``Online dictionary learning for
  sparse coding,'' in \emph{Proceedings of the 26th annual international
  conference on machine learning}.\hskip 1em plus 0.5em minus 0.4em\relax ACM,
  2009, pp. 689--696.

\bibitem{miller2016key}
A.~Miller, A.~Fisch, J.~Dodge, A.-H. Karimi, A.~Bordes, and J.~Weston,
  ``Key-value memory networks for directly reading documents,''
  \emph{Proceedings of the 2016 Conference on Empirical Methods in Natural
  Language Processing}, 2016.

\bibitem{dauphin2017language}
Y.~N. Dauphin, A.~Fan, M.~Auli, and D.~Grangier, ``Language modeling with gated
  convolutional networks,'' in \emph{Proceedings of the 34th International
  Conference on Machine Learning-Volume 70}.\hskip 1em plus 0.5em minus
  0.4em\relax JMLR. org, 2017, pp. 933--941.

\bibitem{rennie2017self}
S.~J. Rennie, E.~Marcheret, Y.~Mroueh, J.~Ross, and V.~Goel, ``Self-critical
  sequence training for image captioning,'' in \emph{CVPR}, vol.~1, 2017, p.~3.

\bibitem{vedantam2015cider}
R.~Vedantam, C.~Lawrence~Zitnick, and D.~Parikh, ``Cider: Consensus-based image
  description evaluation,'' in \emph{Proceedings of the IEEE conference on
  computer vision and pattern recognition}, 2015, pp. 4566--4575.

\bibitem{kingma2014adam}
D.~P. Kingma and J.~Ba, ``Adam: A method for stochastic optimization,''
  \emph{3rd International Conference on Learning Representations}, 2015.

\bibitem{karpathy2015deep}
A.~Karpathy and L.~Fei-Fei, ``Deep visual-semantic alignments for generating
  image descriptions,'' in \emph{Proceedings of the IEEE conference on computer
  vision and pattern recognition}, 2015, pp. 3128--3137.

\bibitem{papineni2002bleu}
K.~Papineni, S.~Roukos, T.~Ward, and W.-J. Zhu, ``Bleu: a method for automatic
  evaluation of machine translation,'' in \emph{Proceedings of the 40th annual
  meeting on association for computational linguistics}.\hskip 1em plus 0.5em
  minus 0.4em\relax Association for Computational Linguistics, 2002, pp.
  311--318.

\bibitem{banerjee2005meteor}
S.~Banerjee and A.~Lavie, ``Meteor: An automatic metric for mt evaluation with
  improved correlation with human judgments,'' in \emph{Proceedings of the acl
  workshop on intrinsic and extrinsic evaluation measures for machine
  translation and/or summarization}, 2005, pp. 65--72.

\bibitem{lin2004rouge}
C.-Y. Lin, ``Rouge: A package for automatic evaluation of summaries,''
  \emph{Text Summarization Branches Out}, 2004.

\bibitem{anderson2016spice}
P.~Anderson, B.~Fernando, M.~Johnson, and S.~Gould, ``Spice: Semantic
  propositional image caption evaluation,'' in \emph{European Conference on
  Computer Vision}.\hskip 1em plus 0.5em minus 0.4em\relax Springer, 2016, pp.
  382--398.

\bibitem{zhang2018learning}
Y.~Zhang, J.~Hare, and A.~Pr{\"u}gel-Bennett, ``Learning to count objects in
  natural images for visual question answering,'' in \emph{ICLR}, 2018.

\bibitem{jiang2018recurrent}
W.~Jiang, L.~Ma, Y.-G. Jiang, W.~Liu, and T.~Zhang, ``Recurrent fusion network
  for image captioning,'' in \emph{Proceedings of the European Conference on
  Computer Vision (ECCV)}, 2018, pp. 499--515.

\bibitem{qin2019look}
Y.~Qin, J.~Du, Y.~Zhang, and H.~Lu, ``Look back and predict forward in image
  captioning,'' in \emph{Proceedings of the IEEE Conference on Computer Vision
  and Pattern Recognition}, 2019, pp. 8367--8375.

\bibitem{liu2018context}
D.~Liu, Z.-J. Zha, H.~Zhang, Y.~Zhang, and F.~Wu, ``Context-aware visual policy
  network for sequence-level image captioning,'' in \emph{2018 ACM Multimedia
  Conference on Multimedia Conference}.\hskip 1em plus 0.5em minus 0.4em\relax
  ACM, 2018, pp. 1416--1424.

\bibitem{he2016deep}
K.~He, X.~Zhang, S.~Ren, and J.~Sun, ``Deep residual learning for image
  recognition,'' in \emph{Proceedings of the IEEE conference on computer vision
  and pattern recognition}, 2016, pp. 770--778.

\end{thebibliography}
% \newpage

\begin{IEEEbiography}[{\includegraphics[width=1in,height=1.25in,clip,keepaspectratio]{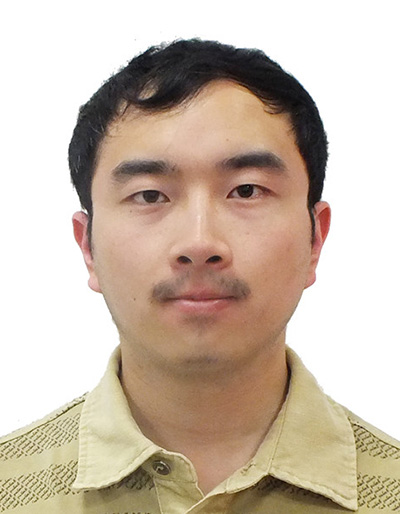}}]{Xu Yang}
	received the B.Eng. degree in Communication Engineering from Nanjing University of Posts and Telecommunications in 2013, the M.Eng. degree in Information Processing from Southeast University in 2016, and the Ph.D. degree in computer science from Nanyang Technological University in 2021. He is currently an Associate Professor at the School of Computer Science and Engineering of Southeast University, China. His research interests mainly include computer vision, machine learning and Image Captioning.
\end{IEEEbiography}
\vskip -2.5\baselineskip plus -1fil
\begin{IEEEbiography}[{\includegraphics[width=1in,height=1.25in,clip,keepaspectratio]{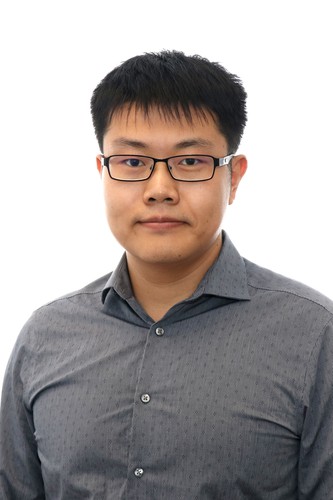}}]
{Hanwang Zhang} is currently an Assistant Professor at Nanyang Technological University, Singapore. He was a research scientist at the Department of Computer Science, Columbia University, USA. He has received the B.Eng (Hons.) degree in computer science from Zhejiang University, Hangzhou, China, in 2009, and the Ph.D. degree in computer science from the National University of Singapore in 2014. His research interest includes computer vision, multimedia, and social media. Dr. Zhang is the recipient of the Best Demo runner-up award in ACM MM 2012, the Best Student Paper award in ACM MM 2013, and the Best Paper Honorable Mention in ACM SIGIR 2016, and TOMM best paper award 2018. He is also the winner of Best Ph.D. Thesis Award of School of Computing, National University of Singapore, 2014.
\end{IEEEbiography}
\vskip -2.5\baselineskip plus -1fil
\begin{IEEEbiography}[{\includegraphics[width=1in,height=1.25in,clip,keepaspectratio]{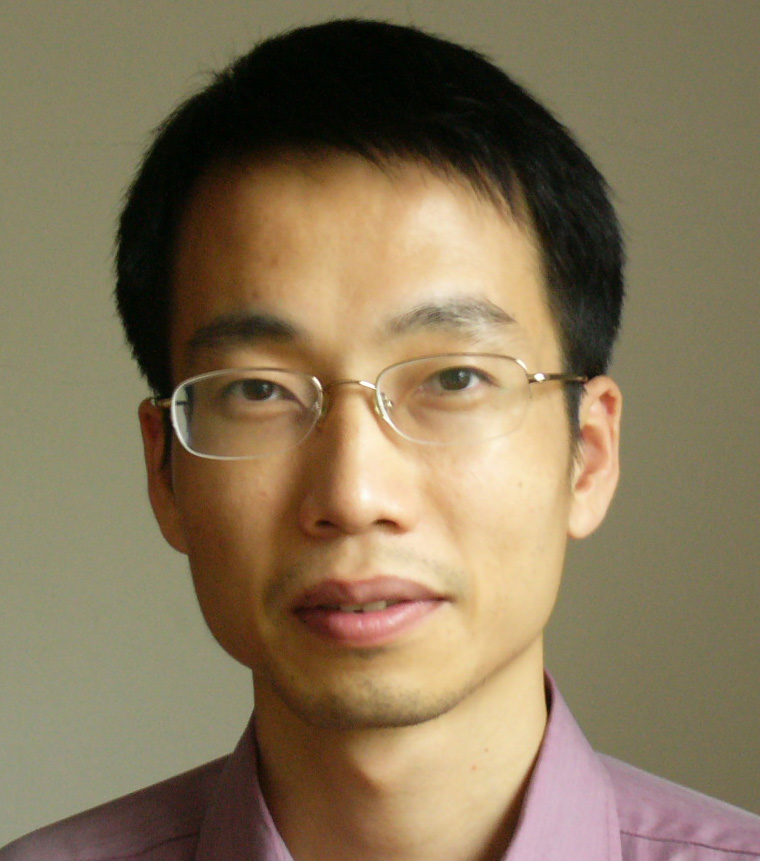}}]{Jianfei Cai}(S'98-M'02-SM'07-F’21) received his PhD degree from the University of Missouri-Columbia. He is currently a Professor and serves as the Head of the Data Science \& AI Department at Faculty of IT, Monash University, Australia. Before that, he had served as a Cluster Deputy Director of Data Science \& AI Research Center (DSAIR), Head of Visual and Interactive Computing Division and Head of Computer Communications Division in Nanyang Technological University (NTU). His major research interests include visual computing, computer vision and multimedia. He has published over 200 technical papers in international journals and conferences. He has served as an Associate Editor for IEEET-IP, T-MM, T-CSVT and Visual Computer as well as serving as Area Chair for ICCV, ECCV, ACM Multimedia, ICME and ICIP. He was the Chair of IEEE CAS VSPC-TC during 2016-2018. He had also served as the leading TPC Chair for IEEE ICME 2012. He is a Fellow of IEEE.
\end{IEEEbiography}

\end{document}

\title{Supplementary Material of Deconfounded Image Captioning: A Causal Retrospect}

\author{Xu~Yang,%~\IEEEmembership{Member,~IEEE,}
        ~Hanwang~Zhang,%~\IEEEmembership{Fellow,~OSA,}
        ~and~Jianfei~Cai%~\IEEEmembership{Life~Fellow,~IEEE}% <-this % stops a space
\IEEEcompsocitemizethanks{\IEEEcompsocthanksitem Xu Yang is currently an Associate Professor at School of Computer Science and Engineering of Southeast University, China.\protect\\
E-mail: xuyangseu@ieee.org

\IEEEcompsocthanksitem Hanwang Zhang is currently an Assistant Professor at Nanyang Technological University, Singapore.\protect\\
E-mail: hanwangzhang@ntu.edu.sg

\IEEEcompsocthanksitem Jianfei Cai is currently a Professor at the Data Science \& AI Department at the Faculty of IT, Monash University, Australia.\protect\\ 
E-mail: Jianfei.Cai@monash.edu

}}% <-this % stops an unwanted space

\maketitle

This supplementary document details the following aspects of the submitted manuscript: A. Causal Preliminaries, B. Formula Derivations, C. Network Architecture D. More Results, E. The Experiments on Video Captioning.
\section{Causal Preliminaries}
\subsection{Causal Graph}
\begin{figure}[t]
\centering
\includegraphics[width=1\linewidth,trim = 5mm 5mm 5mm 5mm,clip]{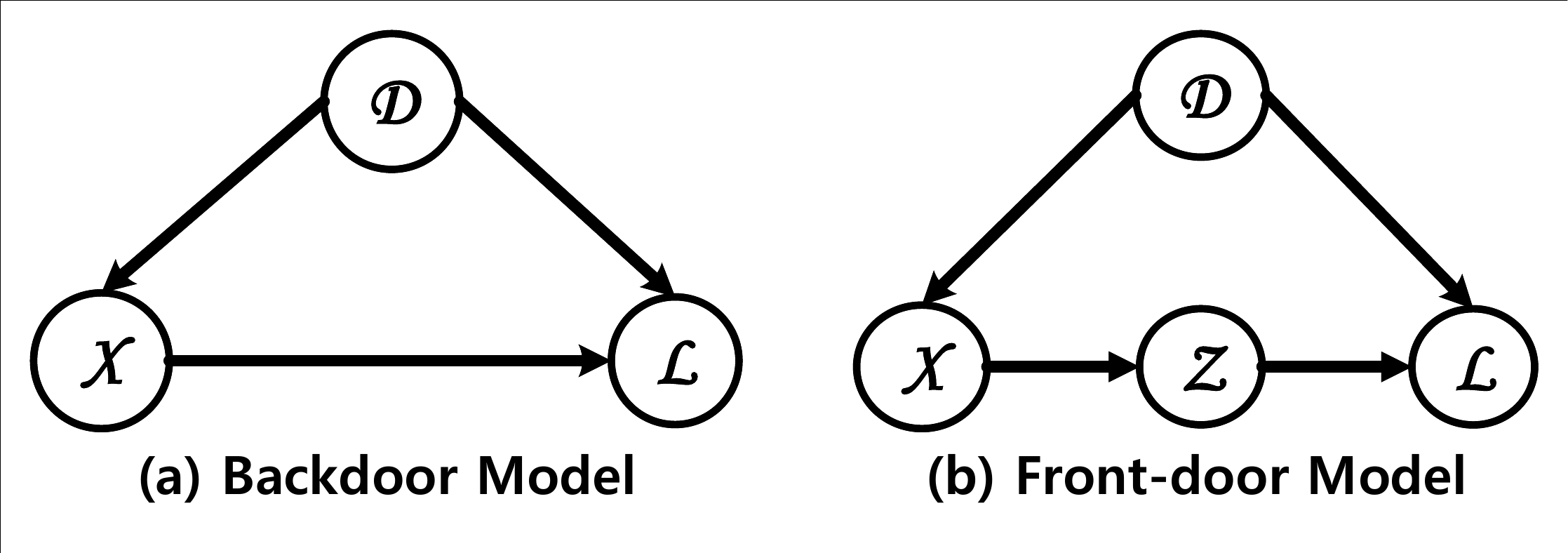}
   \caption{Two causal graphs which are (a) a backdoor model and (b) a front-door model.
   }
\label{fig:fig_supp_cg}
\end{figure}
Before detailing the derivations of the deconfounding techniques, it is beneficial to discuss more details about the causal graph. In the framework of causal inference~\cite{pearl2016causal,PearlMackenzie18}, a causal graph is represented by a directed graph where the direction of the arrow means whether this link is causal or anticausal, \eg, $\mathcal{X}\rightarrow\mathcal{L}$ is a causal link and $\mathcal{X}\leftarrow\mathcal{D}$ is an anticausal link. Note that the information can be passed in both directions: causal or anticausal. For example, as shown in Fig.~\ref{fig:fig_supp_cg}(a), there are two paths between $\mathcal{X}$ and $\mathcal{L}$: a causal path $\mathcal{X}\rightarrow\mathcal{L}$ and a backdoor path $\mathcal{X} \leftarrow \mathcal{D} \rightarrow \mathcal{L}$ which induces the spurious correlation between $\mathcal{X}$ and $\mathcal{L}$ to disturb the learning of the causal effect of $\mathcal{X}$ on $\mathcal{L}$.

Formally, a backdoor path between $\mathcal{X}$ and $\mathcal{L}$ is defined as: \textbf{any path from $\mathcal{X}$ to $\mathcal{L}$ that starts with an arrow pointing into $\mathcal{X}$}. Here we show another two examples for helping understand this concept. In Fig.~\ref{fig:fig_supp_cg}(b), $\mathcal{X} \rightarrow \mathcal{Z}$ is a causal path between $\mathcal{X}$ and $\mathcal{Z}$, while the path $\mathcal{X} \leftarrow \mathcal{D} \rightarrow \mathcal{L} \leftarrow \mathcal{Z}$ is a backdoor path between $\mathcal{X}$ and $\mathcal{Z}$ since $\mathcal{D}$ pointing into $\mathcal{X}$. Another example is the backdoor path $\mathcal{Z} \leftarrow \mathcal{X} \leftarrow \mathcal{D} \rightarrow \mathcal{L}$ between $\mathcal{Z}$ and $\mathcal{L}$ since $\mathcal{X}$ pointing into $\mathcal{Z}$.

In a causal graph, if we want to \textbf{deconfound two variables $\mathcal{X}$ and $\mathcal{L}$ to calculate the causal effect of $\mathcal{X}$ on $\mathcal{L}$: $P(\mathcal{L}|do(\mathcal{X}))$, what we need to do is to block every backdoor path between $\mathcal{X}$ and $\mathcal{L}$}~\cite{PearlMackenzie18}. For example, if we want to get the causal effect of $\mathcal{X}$ on $\mathcal{L}$ in Fig.~\ref{fig:fig_supp_cg}(a), we only need to block the backdoor path $\mathcal{X} \leftarrow \mathcal{D} \rightarrow \mathcal{L}$.

\subsection{Blocking Path}
Here we introduce three rules about how to block a path to stop the information flow between two variables. In a causal graph, there are three different elemental ``junctions'' which construct the whole graph. Correspondingly, there are three rules for blocking information flows in these three junctions. Three junctions are given as follows:

1. $\mathcal{A}\rightarrow\mathcal{B} \rightarrow \mathcal{C}$. This is called the \textbf{chain junction}, where $\mathcal{B}$ is a mediator transmitting information from $\mathcal{A}$ to $\mathcal{C}$. In this junction, once we know the value of the mediator $\mathcal{B}$, learning about $\mathcal{A}$ will not give us any information to raise or lower our belief about $\mathcal{C}$. Therefore, if we directly control $\mathcal{B}$ to certain values, the information flow from $\mathcal{A}$ to $\mathcal{C}$ is blocked. For example, we know that hard-working causes a high-quality paper and finally affects the acceptance of this paper: $\texttt{Hard-Working} \rightarrow \texttt{High-Quality} \rightarrow \texttt{Accept}$, where \texttt{High-Quality} is a mediator. Once we control \texttt{High-Quality} to be true, we know the paper is likely to be accepted and we do not need to know any information about \texttt{Hard-Working} to lower or raise this belief.

2. $\mathcal{A}\leftarrow\mathcal{B} \rightarrow \mathcal{C}$. This is called the \textbf{confounding junction} where $\mathcal{B}$ is a confounder of $\mathcal{A}$ and $\mathcal{C}$. In this junction, once we know what the value of confounder $\mathcal{B}$ is, there is no spurious correlation between $\mathcal{A}$ and $\mathcal{C}$. Therefore, as in the chain junction, if we directly control $\mathcal{B}$ to certain values, the information flow from $\mathcal{A}$ to $\mathcal{C}$ is blocked. We have already met this junction in Introduction of the submitted manuscript where we use digital classification as the example.

%$\texttt{Colorful Paper} \leftarrow \texttt{High-Quality} \rightarrow \texttt{Accept}$ as the example. \jf{There is no such example in the paper. It is also not a good example. `colorful paper' should be the confounder instead of `high quality'. No need to give an example if there is no good example.} In this example, once we control a paper to have \texttt{High-Quality}, the spurious correlation between \texttt{Colorful Paper} and \texttt{Accept} is eliminated, which means we deconfound \texttt{Colorful Paper} and \texttt{Accept}.

3. $\mathcal{A}\rightarrow\mathcal{B} \leftarrow \mathcal{C}$. This is called the \textbf{collider} which works in an exactly opposite way from the above chain and confounding junctions. In a collider, if we do not know what the value of $\mathcal{B}$ is, $\mathcal{A}$ and $\mathcal{C}$ are independent. However, once we know the value of $\mathcal{B}$, $\mathcal{A}$ and $\mathcal{C}$ are correlated! We still use paper acceptance as the example, suppose both \texttt{High-Quality} and \texttt{Luck} affect \texttt{Accept}, though \texttt{High-Quality} and \texttt{Luck} are unrelated. Under this situation, we have $\texttt{Luck} \rightarrow \texttt{Accept} \leftarrow \texttt{High-Quality}$. If we do not know whether a paper is accepted, \texttt{Luck} and \texttt{High-Quality} are independent. Once we know a paper is accepted, there is a negative correlation between \texttt{Luck} and \texttt{High-Quality}: finding out an accepted paper with \texttt{High-Quality} will lower our belief that this paper is accepted due to researchers' \texttt{Luck}. Therefore, this path is naturally blocked if we do not control $\mathcal{B}$.

To sum up, if we want to block the information flow between $\mathcal{A}$ and $\mathcal{C}$, we can directly control $\mathcal{B}$ to certain values in both the chain and confounding junctions, and we must not control $\mathcal{B}$ in a collider.

\section{Formula Derivations}
\subsection{Derivations of Eq.~(17) and Eq.~(21)}
\label{sec:supp_dic}
\begin{figure}[t]
\centering
\includegraphics[width=1\linewidth,trim = 5mm 5mm 5mm 5mm,clip]{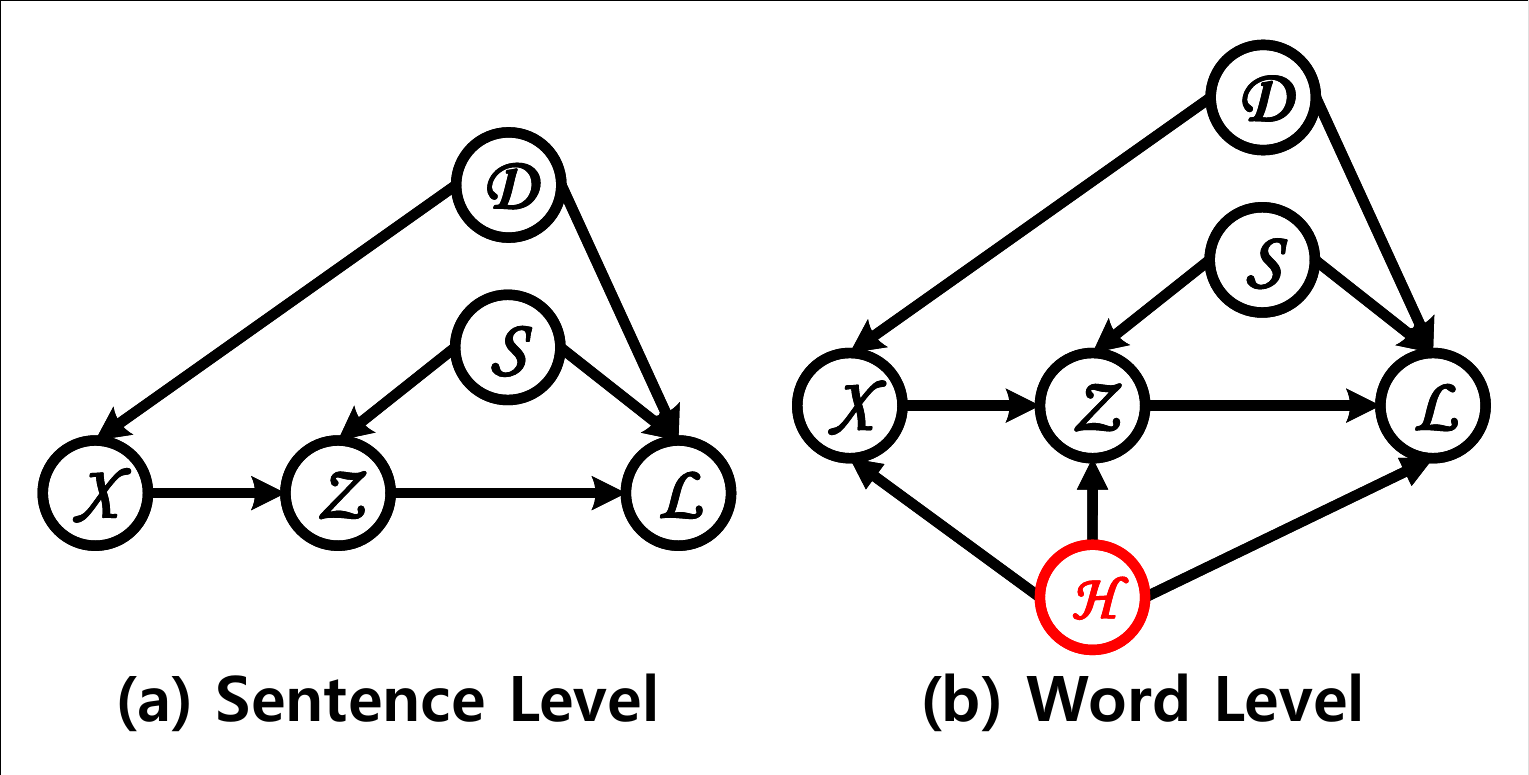}
   \caption{The causal graphs of our DICv1.0 model, (a) and (b) mean two different perspectives of our model: the sentence level and the word level, respectively. In (b), $\mathcal{H}$ denotes the accumulated context knowledge of the partially generated caption, we color $\mathcal{H}$ as red to signify that its value is already known: $\mathcal{H}=\bm{h}$.}
\label{fig:fig_supp_cg2}
\end{figure}
The causal graphs of our DICv1.0 model are shown in Fig.~\ref{fig:fig_supp_cg2}. We will derive Eq.~(17) and Eq.~(21) of the manuscript based on these two causal graphs. 
We first show how to calculate the sentence level $P(\mathcal{L}|do(\mathcal{X}))$ (Eq.~(17)) from the causal graph in Fig.~\ref{fig:fig_supp_cg2}(a). To achieve this, we follow the procedure used in calculating the front-door adjustment (Section 2.2): we first calculate $P(\mathcal{Z}|do(\mathcal{X}))$ and $P(\mathcal{L}|do(\mathcal{Z}))$, then we chain them together to get the final $P(\mathcal{L}|do(\mathcal{X}))$. To get $P(\mathcal{Z}|do(\mathcal{X}))$, we find
\begin{equation}
\label{equ:supp_dic1}
    P(\mathcal{Z}=\bm{z}|do(\mathcal{X})) = P(\mathcal{Z}=\bm{z}|\mathcal{X}).
\end{equation}
since both two backdoor paths $\mathcal{X} \leftarrow \mathcal{D} \rightarrow \mathcal{L} \leftarrow \mathcal{Z}$ and $\mathcal{X} \leftarrow \mathcal{D} \rightarrow \mathcal{L} \leftarrow \mathcal{S} \rightarrow \mathcal{Z}$ between $\mathcal{X}$ and $\mathcal{Z}$ are naturally blocked due to the colliders $\mathcal{D} \rightarrow \mathcal{L} \leftarrow \mathcal{Z}$ and $\mathcal{D} \rightarrow \mathcal{L} \leftarrow \mathcal{S}$, respectively.

To get $P(\mathcal{L}|do(\mathcal{Z}))$, we need to block both two backdoor paths between $\mathcal{Z}$ and $\mathcal{L}$, which are $\mathcal{Z} \leftarrow \mathcal{S} \rightarrow \mathcal{L}$ and $\mathcal{Z} \leftarrow \mathcal{X} \leftarrow \mathcal{D} \rightarrow \mathcal{L}$. To block the former path, we have to control $\mathcal{S}$, and to block the latter path, we have to control $\mathcal{X}$ since $\mathcal{D}$ is not accessible. Therefore, we should control two variables $\mathcal{S}$ and $\mathcal{X}$ to block both the two backdoor paths. Therefore, we have 
\begin{equation}
\begin{aligned}
\label{equ:supp_dic2}
    &P(\mathcal{L}|do(\mathcal{Z}=\bm{z})) \\
    = &\sum\nolimits_{\bm{s}}P(\bm{s})\sum\nolimits_{\bm{x}}P(\bm{x})[P(\mathcal{L}|\bm{s},\bm{x},\bm{z})]. \\
\end{aligned}
\end{equation}
To sum up, after chaining two partial effects together, we have the sentence level $P(\mathcal{L}|do(\mathcal{X}))$ as:
\begin{equation}
\begin{aligned}
\label{equ:supp_dic3}
    &P(\mathcal{L}|do(\mathcal{X})) \\
    = &\sum\nolimits_{\bm{s}}P(\bm{s})\sum\nolimits_{\bm{x}}P(\bm{x})\sum\nolimits_{\bm{z}}P(\bm{z}|\mathcal{X})[P(\mathcal{L}|\bm{s},\bm{x},\bm{z})] \\
    = &\mathbb{E}_{\bm{s}}\mathbb{E}_{\bm{x}}\mathbb{E}_{[\bm{z}|\mathcal{X}]}[P(\mathcal{L}|\bm{s},\bm{x},\bm{z})],
\end{aligned}
\end{equation}
which is Eq.~(17) in the submitted manuscript.

When we calculate $P(\mathcal{L}|do(\mathcal{X}))$ at the word level, as shown in Fig.~\ref{fig:fig_supp_cg2}(b), it is also conditioned on the variable $\mathcal{H}$ which is the accumulated context knowledge of the partially generated caption. At each time step, the value of $\mathcal{H}$ can be pre-computed and we use $\bm{h}$ to denote this value, which means $\mathcal{H}$ is already controlled to be $\bm{h}$. Note that $\mathcal{H}$ only appears in the confounding junctions: $\mathcal{X} \leftarrow \mathcal{H} \rightarrow \mathcal{Z}$, $\mathcal{X} \leftarrow \mathcal{H} \rightarrow \mathcal{L}$, $\mathcal{Z} \leftarrow \mathcal{H} \rightarrow \mathcal{L}$. Therefore, these paths which go through $\mathcal{H}$ are blocked, \eg, $\mathcal{X} \leftarrow \mathcal{H} \rightarrow \mathcal{Z}$ and $\mathcal{X} \leftarrow \mathcal{D} \rightarrow \mathcal{L} \leftarrow \mathcal{H} \rightarrow \mathcal{Z}$ are already blocked. As a result, we can modify Eq.~\eqref{equ:supp_dic1} as:
\begin{equation}
\label{equ:supp_dic4}
    P(\mathcal{Z}=\bm{z}|do(\mathcal{X}),\bm{h}) = P(\mathcal{Z}=\bm{z}|\mathcal{X},\bm{h}),
\end{equation}
Eq.~\eqref{equ:supp_dic2} as:
\begin{equation}
\begin{aligned}
\label{equ:supp_dic5}
    &P(\mathcal{L}|do(\mathcal{Z}=\bm{z}),\bm{h}) \\
    = &\sum\nolimits_{\bm{s}}P(\bm{s}|\bm{h})\sum\nolimits_{\bm{x}}P(\bm{x}|\bm{h})[P(\mathcal{L}|\bm{s},\bm{x},\bm{z},\bm{h})], \\
\end{aligned}
\end{equation}
and Eq.~\eqref{equ:supp_dic3} as:
\begin{equation}
\begin{aligned}
\label{equ:supp_dic6}
     &P(\mathcal{L}|do(\mathcal{X}),\bm{h}) \\
    = &\sum\nolimits_{\bm{s}}P(\bm{s}|\bm{h})\sum\nolimits_{\bm{x}}P(\bm{x}|\bm{h})\sum\nolimits_{\bm{z}}P(\bm{z}|\mathcal{X},\bm{h})[P(\mathcal{L}|\bm{s},\bm{x},\bm{z},\bm{h})] \\
    = &\mathbb{E}_{[\bm{s}|\bm{h}]}\mathbb{E}_{[\bm{x}|\bm{h}]}\mathbb{E}_{[\bm{z}|\mathcal{X},\bm{h}]}[P(\mathcal{L}|\bm{s},\bm{x},\bm{z},\bm{h})],
\end{aligned}
\end{equation}
which is Eq.~(21) in the submitted manuscript. Note that $P(\bm{s}|\bm{h})$ should be $P(\bm{s})$ since there is no direct link from $\mathcal{H}$ to $\mathcal{S}$. While in the experiment, we still set $\mathcal{S}$ to be conditioned on $\bm{h}$ to increase the representation power of the whole model. And if not, by normalized weighted geometric mean approximation (see Section~\ref{sec:supp_nwgm}), the expectation of $\mathcal{S}$ will degrade to a fixed vector.

\subsection{Derivations of Eq.~(19) and Eq.~(20)}
\label{sec:supp_nwgm}
Here we show how to use Normalized Weighted Geometric Mean (NWGM) approximation~\cite{xu2015show,srivastava2014dropout,baldi2014dropout} to absorb the expectations into the network for deriving Eq.~(19) in the submitted manuscript. Before introducing NWGM, we first revisit the calculation of a function $f(\mathcal{X})$'s expectation according to the distribution $P(\mathcal{X})$:
\begin{equation}
\label{equ:supp_nwgm12}
\mathbb{E}_{\bm{x}}[f(\bm{x})]=\sum\nolimits_{\bm{x}}{f(\bm{x})P(\bm{x})},
\end{equation}
which is the weighted arithmetic mean of $f(\bm{x})$ with $P(\bm{x})$ as the weights. Correspondingly, the weighted geometric mean (WGM) of $f(\bm{x})$ with $P(\bm{x})$ as the weights is:
\begin{equation}
\label{equ:supp_nwgm1}
\text{WGM}(f(\bm{x}))=\prod\nolimits_{\bm{x}}f(\bm{x})^{P(\bm{x})},
\end{equation}
where the weights $P(\bm{x})$ are put into the exponential terms. If $f(\bm{x})$ is an exponential function that $f(\bm{x})=\text{exp}[g(\bm{x})]$, we have:
\begin{equation}
\begin{aligned}
\label{equ:supp_nwgm2}
     &\text{WGM}(f(\bm{x}))=\prod\nolimits_{\bm{x}}f(\bm{x})^{P(\bm{x})}\\
     &=\prod\nolimits_{\bm{x}}\text{exp}[g(\bm{x})]^{P(\bm{x})} 
    =\prod\nolimits_{\bm{x}}\text{exp}[g(\bm{x})P(\bm{x})] \\ &=\text{exp}[\sum\nolimits_{\bm{x}}g(\bm{x})P(\bm{x})]=\text{exp}\{\mathbb{E}_{\bm{x}}[g(\bm{x})]\},
\end{aligned}
\end{equation}
where the expectation $\mathbb{E}_{\bm{x}}$ is absorbed into the exponential term. Based on this observation, researchers approximate the expectation of a function by the WGM of this function in the deep network whose last layer is a Softmax layer~\cite{xu2015show,srivastava2014dropout,baldi2014dropout}:
\begin{equation}
\label{equ:supp_nwgm3}
     \mathbb{E}_{\bm{x}}[f(\bm{x})] \approx \text{WGM}(f(\bm{x}))=\text{exp}\{\mathbb{E}_{\bm{x}}[g(\bm{x})]\},
\end{equation}
where $f(\bm{x})=\text{exp}[g(\bm{x})]$.

In our case, we parameterize $p(\mathcal{L}|\bm{s},\bm{x},\bm{z})$ in Eq.~\eqref{equ:supp_dic3} as a network with a Softmax layer as the last layer:
\begin{equation}
\begin{aligned}
\label{equ:supp_nwgm4}
     P(\mathcal{L}|\bm{s},\bm{x},\bm{z})=\text{Softmax}[g(\bm{s},\bm{x},\bm{z})] \propto \exp[g(\bm{s},\bm{x},\bm{z})].
\end{aligned}
\end{equation}

We follow Eq.~\eqref{equ:supp_dic3} and~\eqref{equ:supp_nwgm3} to get:
\begin{equation}
\begin{aligned}
\label{equ:supp_nwgm5}
    &P(\mathcal{L}|do(\mathcal{X})) =  \mathbb{E}_{\bm{s}}\mathbb{E}_{\bm{x}}\mathbb{E}_{[\bm{z}|\mathcal{X}]}[P(\mathcal{L}|\bm{s},\bm{x},\bm{z})] \\
    &\approx \text{WGM}(P(\mathcal{L}|\bm{s},\bm{x},\bm{z}))
    \approx \exp\{\mathbb{E}_{\bm{s}}\mathbb{E}_{\bm{x}}\mathbb{E}_{[\bm{z}|\mathcal{X}]}[g(\bm{s},\bm{x},\bm{z})]\}.
\end{aligned}
\end{equation}
Note that, as in Eq.~\eqref{equ:supp_nwgm4}, $P(\mathcal{L}|\bm{s},\bm{x},\bm{z})$ is only proportional to $\exp[g(\bm{s},\bm{x},\bm{z})]$ instead of strictly equalling to, we only have $\text{WGM}(P(\mathcal{L}|\bm{s},\bm{x},\bm{z})) \approx \exp\{\mathbb{E}_{\bm{s}}\mathbb{E}_{\bm{x}}$ $\mathbb{E}_{[\bm{z}|\mathcal{X}]}[g(\bm{s},\bm{x},\bm{z})]\}$ in Eq.~\eqref{equ:supp_nwgm5} instead of equalling to. Furthermore, to guarantee the sum of $P(\mathcal{L}|do(\mathcal{X}))$ to be 1, we use a Softmax layer to normalize these exponential units:
\begin{equation}
\begin{aligned}
\label{equ:supp_nwgm6}
    P(\mathcal{L}|do(\mathcal{X})) \approx \text{Softmax}\{\mathbb{E}_{\bm{s}}\mathbb{E}_{\bm{x}}\mathbb{E}_{[\bm{z}|\mathcal{X}]}[g(\bm{s},\bm{x},\bm{z})]\},
\end{aligned}
\end{equation}
which is Eq.~(19) in the submitted manuscript. Since the Softmax layer normalizes these exponential terms, this is called the normalized weighted geometric mean (NWGM) approximation.
In addition, if $g(\cdot)$ is a fully connected layer, we have:
\begin{equation}
\label{equ:supp_nwgm7}
P(\mathcal{L}|do(\mathcal{X})) \approx \text{Softmax}\{ g(\mathbb{E}_{\bm{s}}[\bm{s}],\mathbb{E}_{\bm{x}}[\bm{x}],\mathbb{E}_{[\bm{z}|\mathcal{X}]}[\bm{z}])\}.
\end{equation}
In the same vein, we can use NWGM to approximate the word level distribution Eq.~\eqref{equ:supp_dic6} as:
\begin{equation}
\begin{aligned}
\label{equ:supp_nwgm8}
    &P(\mathcal{L}|do(\mathcal{X}),\bm{h}) \\
    &\approx \text{Softmax}\{ g(\mathbb{E}_{[\bm{s}|\bm{h}]}[\bm{s}],\mathbb{E}_{[\bm{x}|\bm{h}]}[\bm{x}],\mathbb{E}_{[\bm{z}|\mathcal{X},\bm{h}]}[\bm{z}])\}.
\end{aligned}
\end{equation}
Then we can use \textsc{expt} modules introduced in Section 4.2 (Eq.~(25)) of the submitted manuscript to estimate these expectations. As discussed in the end of Section~\ref{sec:supp_dic}, we set $\mathcal{S}$ to be conditioned on $\bm{h}$ to increase the representation power. If not, we find that $\mathbb{E}_{[\bm{s}]}[\bm{s}]$ will be the same value all the time during the caption generation.

\section{Network Architecture}
In this section, we will detail the network architectures of UD-DICv1.0 and AoA-DICv1.0 proposed in Section 4.4 of the submitted manuscript.
\subsection{EXPT Module}
In Eq.~(25) of the submitted manuscript, we show how \textsc{expt} module works. The detail structure of this module is listed in Table~\ref{table:supp_expt}.

\subsection{ATT Module}
Two \textsc{att} modules are respectively deployed in UD-DICv1.0 and AoA-DICv1.0, which are Top-Down Attention and Attention on Attention. The details of two modules are given in Table~\ref{table:supp_topdown} and Table~\ref{table:supp_aoa}, respectively. Self attention in Table~\ref{table:supp_aoa}(c) is computed as follows:
\begin{equation} \label{equ:rela_mod}
\small
\begin{aligned}
 \textbf{Input:} \quad  &\mathcal{X},\bm{h} \\
 \textbf{Head:} \quad &  \textbf{head}_k= \text{Softmax}( \frac{\bm{h}_1\bm{W}_k^1(\mathcal{X}\bm{W}_k^2)^T}{\sqrt{d_k}} )\mathcal{X}\bm{W}_k^3,  \\
 \textbf{Multihead:}  \quad & \mathcal{M} = \text{Cat}(\textbf{head}_1,...,\textbf{head}_8)\bm{W}_{C}, \\
 \textbf{Output:} \quad  & \hat{\bm{v}}= \text{LeakyReLU}(\text{MLP}(\mathcal{M})),\\
\end{aligned}
\end{equation}
where $\bm{W}_k^1,\bm{W}_k^2,\bm{W}_k^3 \in \mathbb{R}^{1,000*125}$, $\bm{W}_C \in \mathbb{R}^{1,000*1,000}$ are all trainable matrices and $d_k=1000/8=125$.

\subsection{Common Structure of the Decoder}
The common structure of the two decoders of UD-DICv1.0 and AoA-DICv1.0 is given in Table~\ref{table:supp_dec}. When UD-DICv1.0 or AoA-Dicv1.0 is implemented, \textsc{att} module in (14) is Top-Down attention (Table~\ref{table:supp_topdown}) or Attention on Attention (Table~\ref{table:supp_aoa}), and $g(\cdot)$ in (18) is an LSTM layer or a GLU layer.

\begin{table*}[t]
\begin{center}
\caption{The details of \textsc{expt} module.}
\label{table:supp_expt}
\begin{tabular}{|c|c|c|c|c|c|}
		\hline
		   \textbf{Index}&\textbf{Input}&\textbf{Operation}&\textbf{Output}&\textbf{Trainable Parameters}\\ \hline
		   (1)  & context vector &  -  & $\bm{h}$ (1,000) & - \\ \hline
		   (2)  & Dictionary $\bm{X}$ &  -  & $\bm{X}$ (1,000 $\times$ 10,000) & - \\ \hline
		   (3)  & (1),(2) & inner product $\bm{X}^T\bm{h}$ & $\bm{p}$ (10,000) & $\bm{X}$(1,000 $\times$ 10,000) \\ \hline
		   (4)  & (3)          & Softmax & $P$ (10,000) & - \\ \hline
		   (5)  & (4)          & weighted sum $\bm{X}P$ & $\bar{\bm{x}}$(1,000) & -\\ \hline
\end{tabular}
\end{center}
\end{table*}

\begin{table*}[t]
\begin{center}
\caption{The details of Top-Down Attention.}
\label{table:supp_topdown}
\begin{tabular}{|c|c|c|c|c|c|}
		\hline
		   \textbf{Index}&\textbf{Input}&\textbf{Operation}&\textbf{Output}&\textbf{Trainable Parameters}\\ \hline
		   (1)  &    -    &  feature set  & $\mathcal{X}$ ($M \times 1,000$) & - \\ \hline
		   (2)  &    -    &  context vector  & $\bm{h}_1$ ($1,000$) & - \\ \hline
		   (3)  &  (1),(2)&
		    \begin{tabular}{c}
		         attention weights \\
		         $\bm{\omega}_{a}^T\tanh({\bm{W}_x\bm{x}_n+\bm{W}_h\bm{h}_1})$\\
		   \end{tabular}
		     & $\bm{\alpha}$ ($M$) &
		   \begin{tabular}{c}
		         $\bm{\omega}_{a}$ (512), $\bm{W}_x$ ($512\times1,000$) \\
		         $\bm{W}_h$($512\times1,000$)
		   \end{tabular}
		   \\ \hline 
		   (4)  &    (3)     & Softmax & $\bm{\alpha}$ ($M$) & - \\ \hline
		   (5)  & (1),(4) & weighted sum $\mathcal{X}\bm{\alpha}$ & $\hat{\bm{x}}$ ($1,000$) & - \\ \hline 
\end{tabular}
\end{center}
\end{table*}

\begin{table*}[t]
\begin{center}
\caption{The details of Attention on Attention.}
\label{table:supp_aoa}
\scalebox{1}{
\begin{tabular}{|c|c|c|c|c|c|}
		\hline
		   \textbf{Index}&\textbf{Input}&\textbf{Operation}&\textbf{Output}&\textbf{Trainable Parameters}\\ \hline
		   (1)  &    -    &  feature set  & $\mathcal{X}$ ($M \times 1,000$) & - \\ \hline
		   (2)  &    -    &  context vector  & $\bm{h}_1$ ($1,000$) & - \\ \hline
		   (3)  &    (1),(2)    &  self attention   &$\hat{\bm{v}}$ ($1,000$)  & - \\  \hline
		   (4)  &  (1),(2),(3)&
		    \begin{tabular}{c}
		         information vector \\
		         $\bm{W}_h^i\bm{h}_1+\bm{W}_{\mathcal{X}}^i\hat{\bm{v}}+\bm{b}^i$\\
		   \end{tabular}
		     & $\bm{v}^i$ ($1,000$) &
		   \begin{tabular}{c}
		         $\bm{W}_h^i$ ($1,000\times1,000$) \\
		         $\bm{W}_{\mathcal{X}}^i$($1,000\times1,000$),$\bm{b}^i$ (1000)
		   \end{tabular}
		   \\ \hline 
		   (5)  &  (1),(2),(3)&
		    \begin{tabular}{c}
		         attention gate \\
		         $\sigma(\bm{W}_h^g\bm{h}_1+\bm{W}_{\mathcal{X}}^g\hat{\bm{x}}+\bm{b}^g)$\\
		   \end{tabular}
		     & $\bm{v}^g$ ($1,000$) &
		   \begin{tabular}{c}
		         $\bm{W}_h^g$ ($1,000\times1,000$) \\
		         $\bm{W}_{\mathcal{X}}^g$($1,000\times1,000$),$\bm{b}^g$ (1000)
		   \end{tabular}
		   \\ \hline 
		   (6)  &    (4),(5)    &  element-wise multiplication  $\bm{v}^i .*\bm{v}^g$ &$\hat{\bm{v}}$ ($1,000$)  & - \\  \hline
		   
\end{tabular}
}
\end{center}
\end{table*}

\begin{table*}[t]
\begin{center}
\caption{The details of the common structure of the two decoders.}
\label{table:supp_dec}
\scalebox{1}{
\begin{tabular}{|c|c|c|c|c|}
		\hline
		   \textbf{Index}&\textbf{Input}&\textbf{Operation}&\textbf{Output}&\textbf{Trainable Parameters}\\ \hline
		   (1)  & - &  word label     &  $\bm{w}^{t-1}$ (10,369) & - \\ \hline
		   (2)  & - & word generator's output at $t-1$     &  $\bm{o}^{t-1}$ ($1,000$) & - \\ \hline
		   (3)  & - & image feature set $\mathcal{X}$ & $\mathcal{X}$ ($M \times 1,000$) & -  \\ \hline
		   (4)  & - & Dictionary $\bm{Z}$ & $\bm{Z}$ ($1,000 \times 9,590$) & -  \\ \hline
		   (5)  & - & Dictionary $\bm{S}$ & $\bm{S}$ ($1,000 \times 1,342$) & -  \\ \hline
		   (6)  & - & Dictionary $\bm{X}$ & $\bm{X}$ ($1,000 \times 10,000$) & -  \\ \hline
		   (7)  & (1)   &  word embedding  $\bm{W}_{\Sigma}\bm{w}^{t-1}$  &  $\bm{e}^{t-1}$ (1,000) & $\bm{W}_{\Sigma}$ (1,000 $\times$ 10,369) \\ \hline
		   (8)  & (3)    & mean pooling & $\tilde{\bm{x}}$ (1,000) & - \\ \hline 
		   (9)  & (2),(7),(8) & concatenate & $\bm{u}^t$ (3,000) & - \\ \hline 
		   (10)  & (9)  & LSTM$_1$ $(\bm{u}^t;\bm{h}^{t-1}_1)$  & $\bm{h}^{t}_1$ (1,000) &   LSTM$_1$ (3,000 $\rightarrow$ 1,000)\\ \hline 
		   (11)  & (9)  & LSTM$_2$ $(\bm{u}^t;\bm{h}^{t-1}_2)$  & $\bm{h}^{t}_2$ (1,000) &   LSTM$_2$ (3,000 $\rightarrow$ 1,000)\\ \hline 
		   (12)  & (9)  & LSTM$_3$ $(\bm{u}^t;\bm{h}^{t-1}_3)$  & $\bm{h}^{t}_3$ (1,000) &   LSTM$_3$ (3,000 $\rightarrow$ 1,000)\\ \hline 
		   (13)  & (9)  & LSTM$_4$ $(\bm{u}^t;\bm{h}^{t-1}_4)$  & $\bm{h}^{t}_4$ (1,000) &   LSTM$_4$ (3,000 $\rightarrow$ 1,000)\\ \hline 
		   (14)  & (3),(10)  & \textsc{att} $\mathcal{X}$  & $\hat{\bm{x}}$ ($1,000$) & \textsc{att}  \\ \hline
		   (15)  & (4),(11),(14)  & \textsc{expt}$[\bm{z}]$  & $\bar{\bm{z}}$ ($1,000$) & \textsc{expt}  \\ \hline
		   (16)  & (5),(12)  & \textsc{expt}$[\bm{s}]$  & $\bar{\bm{s}}$ ($1,000$) & \textsc{expt}  \\ \hline
		   (17)  & (6),(13)  & \textsc{expt}$[\bm{x}]$  & $\bar{\bm{x}}$ ($1,000$) & \textsc{expt}  \\ \hline
		   (18)  & (14),(15),(16),(17)  & $g(\tilde{\bm{x}},\bar{\bm{z}},\bar{\bm{s}},\bar{\bm{x}})$  & $\bm{o}_t$ (10,369) & $g(\cdot)$  \\ \hline
		   (19) & (18) & Softmax & $P_t$ (10,369) & - \\ \hline
		   
\end{tabular}
}
\end{center}
\end{table*}

\section{More results}
\subsection{Effects of the Dataset Biases.}
\begin{figure}[t]
\centering
\includegraphics[width=1\linewidth,trim = 5mm 5mm 5mm 5mm,clip]{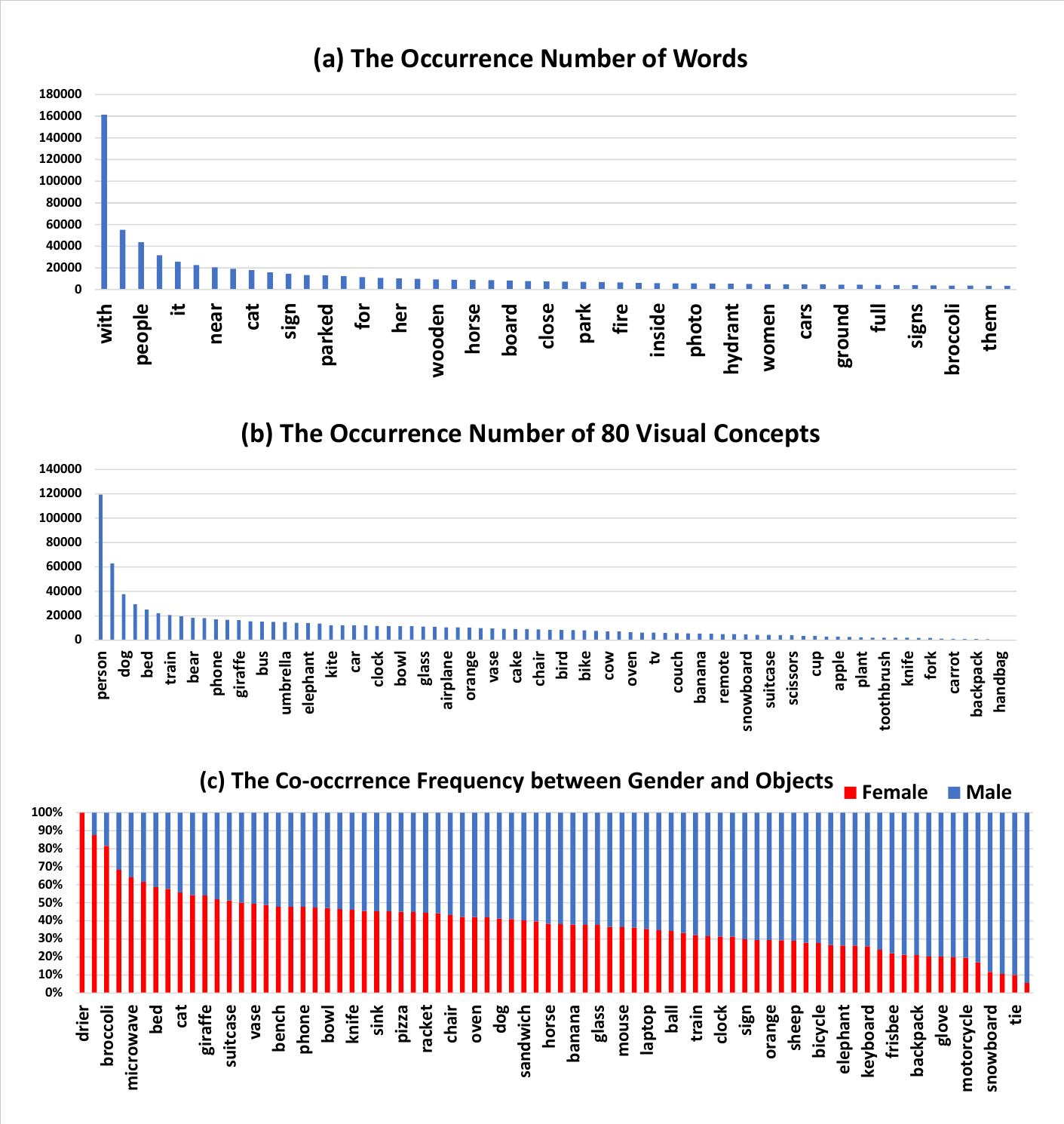}
  \caption{Three different types of dataset biases.
  }
\label{fig:fig_bias_response}
\vspace{-0.2in}
\end{figure}

\begin{table}[t]
\begin{center}
\caption{UD* and UD-DICv1.0* denote that they are trained by the gender balanced dataset and the red color in the subscript denotes the changes compared with the models trained by the integral dataset. UD and UD-DICv1.0 are trained by the integral dataset.}
\label{table:tab_gender_balance}
\scalebox{1}{
\begin{tabular}{l c c c c}
		\hline
		   Models   &  UD* & UD-DICv1.0* & UD & UD-DICv1.0 \\ \hline
             CIDEr-D      & 118.1$_{\color{red}{-7.2}}$ &  123.4  $_{\color{red}{-5.4}}$& 125.3 & 128.8 \\ 
            A@Gen   &0.87$_{\color{red}{+0.6}}$ & 0.94$_{\color{red}{+0.3}}$& 0.81 &  0.91 \\ 
          \hline 
\end{tabular}
}
\end{center}
\vspace{-0.2in}
\end{table}
As discussed in Introduction and Section~2.2 of the main paper, since the composition of the confounder is often complex, an exhaustively quantitative analysis of the dataset bias is almost impossible. However, we can analyze certain types of dataset biases, as shown in Fig.~\ref{fig:fig_bias_response}.

Fig.~\ref{fig:fig_bias_response}~(a) and~(b) show the long-tail bias about the words and 80 visual concepts of COCO, where some words or concepts appear much more frequently than the others. For example, in (a), ``with'' appears more than ``full'', or in (b), ``person'' is more frequently described than ``cup''. Fig.~\ref{fig:fig_bias_response}(c) shows the gender bias that we calculate the co-occurrence frequency between 80 visual concepts with two genders. We can clearly see the gender biases, \eg, most ``drier'' occur with female genders while most ``snowboard'' occur with male genders. As a result, when a captioner recognizes a ``snowboard'', it may directly infer that it is ``man-playing-snowboard'' due to the high co-occurrence frequency. 

Furthermore, biases may appear when certain words co-occur, \eg, when both ``person'' and ``horse'' occur, the action ``ride'' appears more often than ``stand with'', which can cause a captioner to directly infer ``ride'' after recognizing ``person'' and ``horse''. Similarly, the combinations of three or more words can also cause biases, while it would be an arduous task to exhaustively enumerate all these cases.
As a result, it would be impossible to collect a perfect balanced dataset. However, it is feasible to create a ``relatively balanced'' dataset to eliminate some specific biases like the gender-object bias in Fig.~\ref{fig:fig_bias_response}(c). To achieve this, we re-sample the captions in which we keep the co-occurrence frequency between an object with female or male genders to be the same. We delete the captions if the co-occurrence between one gender with an object is larger than another gender. In this way, a total of 34,576 captions are deleted and the rest are used to train and test UD-DICv1.0 and UD.

The results are reported in Table~\ref{table:tab_gender_balance}, where we have three interesting observations. First, the CIDEr-D loss of UD-DICv1.0* is less than UD* when fewer data are available. One possible reason is that UD-DICv1.0 incorporates abundant additional knowledge from ConceptNet, which is beneficial in this case. Second, the improvements of A@Gen in UD* is larger than UD-DICv1.0*, which suggests that the superiority of our DICv1.0 becomes weaker when a more balanced dataset is used. Third, although two models' A@Gens improve when we balance the co-occurrence between the genders and the single visual concepts, the accuracies still do not get 100\%. One possible reason is that there may still exist other factors which cause gender biases, \eg, the non-visual concepts or the combinations of the concepts. Such biases can hardly be eradicated in a captioning dataset, and thus UD-DICv1.0* still performs much better than UD*. 

\re{\subsection{More attention comparisons}}
\begin{figure*}[t]
\centering
\includegraphics[width=1\linewidth,trim = 5mm 5mm 5mm 5mm,clip]{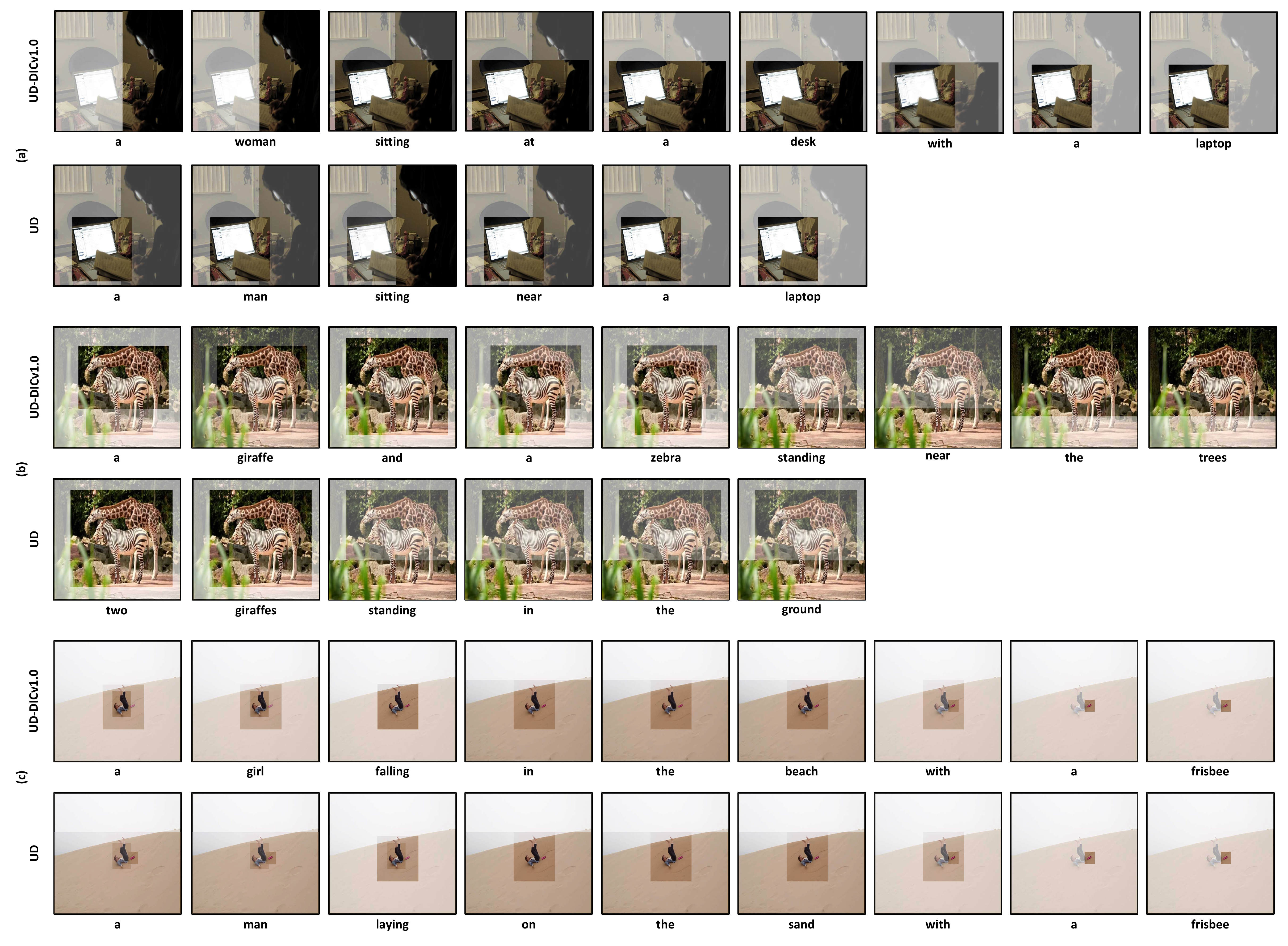}
  \caption{\re{The attention comparisons between UD-DICv1.0 and UD. In each case, the top and bottom ones show the attended regions of UD-DICv1.0 and UD.}
  }
\label{fig:fig_att_response2}
\end{figure*}
\re{In Fig.~\ref{fig:fig_att_response2}, we compare the attention qualities of UD-DICv1.0 and UD. From these cases, we can discvoer that UD-DICv1.0 can improve the attention quality. For example, in Fig.~D(a), when generating the gender, the baseline model UD focuses more on the computer instead of the person and then exploits the co-occurrence between ``computer'' and ``male'' to generate the incorrect gender. However, our UD-DICv1.0 can attend to the correct ``person'' region and exploit the visual clues from this region to generate the correct gender ``woman''. 
}

\re{\section{Video Captioning}
\subsection{Dataset}
\noindent\textbf{VATEX~\cite{wang2019vatex}.}
We also implemented our DIC model on the VATEX dataset. Compared with some previous video captioning datasets like MSR-VTT, VATEX has the largest number of video-caption pairs, \eg, containing over 41,250 videos and 825,000 captions, and it is more semantic rich, \eg, covering 600 human activities. It also provides an official I3D visual feature extractor~\cite{carreira2017quo}, which facilitates us to compare with the other models. More importantly, since this extractor is pre-trained on action classification, it will contain abundant knowledge about human actions and thus may cause the captioner to over-exploit such knowledge to generate biased captions. Therefore, this dataset is a suitable study case for evaluating our DIC in solving the bias problem. We used the official split: 25,991/3000/6000 for training/validation/testing to evaluate our models and compare them with the other models.}

\re{\subsection{Experiments and Results}}
\begin{table}[t]
\begin{center}
\re{
\caption{The performances of the state of the art captioners on VATEX dataset. }
\label{table:tab_vatex}
\scalebox{1}{
\begin{tabular}{l c c c c}
		\hline
		   Model & B@4& M & R &  C\\ \hline
           Up-Down~\cite{anderson2018bottom}  & $28.3$ & $20.1$ & $45.1$ & $47.1$ \\ 
           AoANet~\cite{huang2019attention} & $29.4$ & $22.3$ & $ 47.9$ & $48.6$ \\ 
           VATEX~\cite{wang2019vatex}  & $28.7$& $21.9$ & $47.2$ & $45.6$ \\
           HRN~\cite{wang2019vatex}  & $32.1$& $21.9$ & $48.4$ & $48.5$ \\   
           ORG~\cite{zhang2020object} & $31.5$ & $21.9$ & $48.7$ & $48.8$  \\ 
           ORG+TRL~\cite{zhang2020object} & $32.1$ & $22.2$ & $48.9$ & $49.7$  \\ 
           UD-DICv1.0 & $32.3$ & $22.3$ & $48.7$ & $50.6$ \\ 
           AoA-DICv1.0 & $\bm{33.0}$ & $\bm{22.6}$ & $\bm{49.1}$ & $\bm{51.5}$ \\
           \hline 
\end{tabular}
}
}
\end{center}
\vspace{-0.2in}
\end{table}
\re{
For fair comparisons, we compared our UD-DICv1.0 and AoA-DICv1.0 with the models using the same visual features and the LSTM-based decoder, which are \textbf{UP-Down}~\cite{anderson2018bottom},
\textbf{VATEX}~\cite{wang2019vatex}, \textbf{AoA}~\cite{huang2019attention},
\textbf{HRN}~\cite{lei2021hierarchical},
and \textbf{ORG}~\cite{zhang2020object}. When implementing the DIC on VATEX, we also searched the commonsense structures from ConceptNet. Specifically, we used the nouns, verbs, and adjectives which respectively appear more than 20 times in the training set to search the related commonsense knowledge. We followed the previous models~\cite{wang2019vatex,lei2021hierarchical,zhang2020object} to use the cross-entropy loss~Eq.(19) for fair comparisons. For UD-DICv1.0/AoA-DICv1.0, we trained the models 35 epochs and initialized the learning rates to $5e^{-4}$/$1e^{-4}$ and decayed them 0.8 for every 5 epochs, respectively. The results are reported in TABLE~\ref{table:tab_vatex}. 

For HRN, it builds hierarchical representation and exploits the syntax pattern as the additional supervision, while such supervision still comes from the training dataset, thus the model is not deconfounded. ORG builds a spatial-temporal graph as the mediator and ORG+TRL further uses the pre-trained BERT word embedding to distill the knowledge for alleviating the long tail biases. However, since the graph is got from the training dataset and the pre-trained model may also form a confounder, the whole system is still confounded. From TABLE~\ref{table:tab_vatex}, we can find that our UD-DICv1.0 and AoA-DICv1.0 achieve higher CIDEr-D scores than these models. Such comparisons also confirm the advantage of using deconfounding techniques in video captioning.
}

\re{Discussion About the Other Tasks }
\re{Besides captioning, many vision-language models are also damaged from the dataset bias, \eg, VQA or Visual Dialog systems are discovered to reason by over-relying on the spurious question-answer correlations~\cite{goyal2017making,agrawal2018don} or the history information~\cite{qi2020two}. Meantime, lots of vision-language systems apply various mediators, \eg, graphs~\cite{teney2017graph,jing2020visual} or neural modules~\cite{liu2019learning,hu2018explainable}, to help alleviate the dataset bias problem. However, the used mediators are usually got from the same training set (the confounder) and these systems are usually trained by the cross-entropy loss, which means they are not deconfounded. Therefore, they may still be damaged from the dataset bias. To alleviate the dataset bias, researchers can also follow our method to use the mediators from the third-party dataset, \eg, some knowledge graph datasets, to achieve the front-door adjustment.}

\bibliographystyle{IEEEtran}
\bibliography{egbib}